\pdfoutput=1

\documentclass[pdflatex,sn-mathphys-num]{sn-jnl}%

\usepackage{graphicx}%
\usepackage{multirow}%
\usepackage{amsmath,amssymb,amsfonts}%
\usepackage{amsthm}%
\IfFileExists{mathcomp.sty}{\usepackage{mathcomp}}{\newcommand{\tcdegree}{^{\circ}}}
\usepackage{mathrsfs}%
\usepackage[title]{appendix}%
\usepackage{xcolor}%
\usepackage{textcomp}%
\usepackage{manyfoot}%
\usepackage{booktabs}%
\usepackage{listings}%

\IfFileExists{bxcjkjatype.sty}{\usepackage[whole]{bxcjkjatype}}{}
\usepackage{subcaption}
\newlength{\figtilew}\newlength{\figherow}\newlength{\figcapht}\newlength{\figcolgap}\newlength{\figrowgap}
\newlength{\figclh}%
\newsavebox{\aabcmeasbox}
\usepackage{tikz}
\usetikzlibrary{shapes.geometric, arrows.meta, positioning, calc, fit, backgrounds}
\usepackage[section]{placeins}

\usepackage{etoolbox}
\makeatletter
\patchcmd{\@maketitle}{\newpage\null}{\newpage}{\typeout{TPATCH0 OK}}{\typeout{TPATCH0 FAIL}}
\patchcmd{\@maketitle}{\vskip21pt}{\vskip0pt}{\typeout{TPATCH1 OK}}{\typeout{TPATCH1 FAIL}}
\patchcmd{\@maketitle}{\removelastskip\vskip20pt\nointerlineskip}{\removelastskip\vskip6pt\nointerlineskip}{\typeout{TPATCH2 OK}}{\typeout{TPATCH2 FAIL}}
\patchcmd{\@maketitle}{\vskip20pt}{\vskip8pt}{\typeout{TPATCH3 OK}}{\typeout{TPATCH3 FAIL}}
\patchcmd{\@maketitle}{\removelastskip\vskip24pt}{\removelastskip\vskip10pt}{\typeout{TPATCH4 OK}}{\typeout{TPATCH4 FAIL}}
\patchcmd{\@maketitle}{\removelastskip\vskip24pt}{\removelastskip\vskip10pt}{\typeout{TPATCH5 OK}}{\typeout{TPATCH5 FAIL}}
\patchcmd{\@maketitle}{\removelastskip\vskip36pt\vskip0pt}{\removelastskip\vskip12pt\vskip0pt}{\typeout{TPATCH6 OK}}{\typeout{TPATCH6 FAIL}}
\makeatother

\theoremstyle{thmstyleone}%

\theoremstyle{thmstyletwo}%

\theoremstyle{thmstylethree}%

\usepackage{longtable,array,bm}
\providecommand{\T}[2]{T^{#1}_{#2}}
\providecommand{\SO}{\mathrm{SO}(3)}
\begin{document}

\title[quARtet Marker]{quARtet Marker: A 3D-Printable Multi-Tag Fiducial for Robust Near-Frontal Pose Estimation}

\author*[1]{\fnm{Araki} \sur{Wakiuchi}}\email{araki\_wakiuchi@jsrgr.com}
\author[2]{\fnm{Hikaru} \sur{Sasaki}}\email{sasaki.hikaru@is.naist.jp}
\author[2]{\fnm{Takamitsu} \sur{Matsubara}}\email{takam-m@is.naist.jp}

\affil[1]{\orgname{Materials Informatics Initiative, RD Technology and Digital Transformation Center, JSR Corporation},
\orgaddress{\street{3-103-9 Tonomachi, Kawasaki-ku}, \city{Kawasaki}, \state{Kanagawa}, \country{Japan}, \postcode{210-0821}}}

\affil[2]{\orgdiv{Division of Information Science, Graduate School of Science and Technology}, \orgname{Nara Institute of Science and Technology},
\orgaddress{\street{8916-5 Takayama-cho}, \city{Ikoma}, \state{Nara}, \country{Japan}, \postcode{630-0192}}}

\abstract{
Robotic manipulation of labware is difficult when transparent or reflective objects must be identified and localized. Coded planar fiducials are a practical retrofit: easy to print, they leave the marked face flat and graspable. Yet a single planar tag is least reliable in near-frontal views, where perspective cues fade. Non-planar geometries restore those cues but intrude on the flat face that a parallel-jaw gripper must contact. Our idea is to tilt multiple tags within one compact footprint, so that each tag is seen at a non-frontal angle even when the marker faces the camera. We propose the quARtet marker, a 3D-printable fiducial embodying this idea: all detected corners of its four tilted AprilTags enter one Perspective-n-Point solve, and a shared configuration defines the fabricated geometry and the detector model. Because tilting consumes flat area, its three layouts trade pose-estimation consistency against graspability. In robot-referenced, same-setup fixed-camera experiments, all three layouts reduced the mean frontal orientation error from $2.18\tcdegree$ for a single planar tag to $0.24$--$0.47\tcdegree$ and the root-mean-square position error from $1.50$ to $0.17$--$0.20$~mm. A robot-mounted-camera pose-hold test confirmed this separation under closed-loop visual feedback. In swing-down trials under identical conditions, the two layouts with flat contact strips retained the object, whereas the layout without flat strips slipped about a hundred times more than a single planar tag. For the tested conditions, the results support a rule: the layout without flat strips when pose-estimation consistency dominates, a layout with flat strips when the marked face must remain graspable.
}


\maketitle

\section{Introduction}\label{sec:intro}
\begin{figure}[!t]
    \centering
    \includegraphics[width=\linewidth]{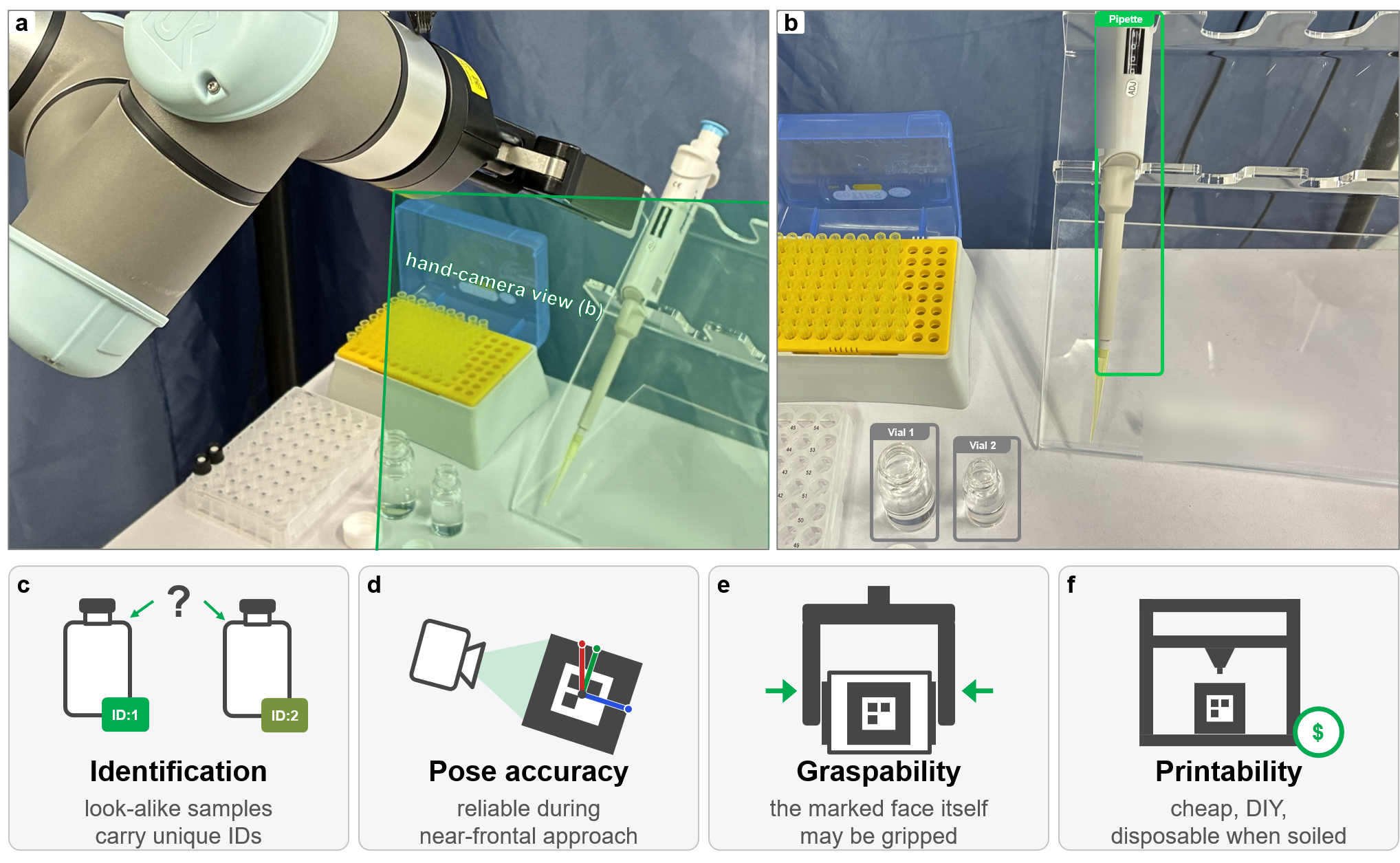}
    \caption{Target context and the four practical demands on a laboratory fiducial. (a)~A robot operates among weakly textured, transparent, or specular labware within the hand-camera field of view. (b)~The hand-camera view, annotated for illustration: the camera must distinguish the pipette to be picked (green box) from two visually similar transparent vials (gray boxes). (c)--(f)~The marker must provide identification, reliable pose estimation during a near-frontal approach, access for grasping, and low-cost printability.}
    \label{fig:research:overview}
\end{figure}

Laboratory automation can raise experimental throughput and keep experiments running for days with little human intervention~\cite{Burger2020,Szymanski2023}, and it can spare people contact with hazardous materials~\cite{Burger2020}, but it also places robots in visually difficult workspaces: transparent vessels challenge appearance and depth sensing~\cite{Sajjan2020,Fang2022}, reflective or weakly textured tools provide few stable visual features, and visually identical containers may hold different materials. To manipulate such labware, a robot must know both \emph{which} object is present and its six-degree-of-freedom (6-DOF) pose: its position and orientation. Manually taught positions and pre-calibrated placements can sidestep part of this perception problem, but the setup and re-teaching effort they demand works against the plug-and-play integration sought when existing laboratories are automated~\cite{Wolf2023}. Printed coded fiducials, reference markers that a camera can both identify and localize, are therefore an attractive retrofit. Printed code tags already label chemical containers in laboratories~\cite{Shukran2017}, marker tracking has been used to automate sensor registration in robot workspaces~\cite{Gong2019}, and an ordinary camera decodes object identity and estimates pose from the same observation~\cite{Romero-Ramirez2018,Wan2016,Rekimoto2000}. The markers are also inexpensive to print and to replace after contamination. Figure~\ref{fig:research:overview} illustrates this application context and the four practical demands considered in this work (the marker proposed to meet them is introduced in Section~\ref{sec:system}, Fig.~\ref{fig:flowchart}). It is not a depiction of the validation experiments.

The remaining problem is the physical and geometric design of the marker itself. A single square tag estimates orientation from the perspective distortion of its outline. Near a frontal view that distortion becomes weak: a tag tilted slightly toward the camera projects to nearly the same image as one tilted equally away, so pixel-level image errors translate into large orientation changes~\cite{Tanaka2012,Tanaka2014}, and the reported pose can flip between these two near-mirror solutions~\cite{Schweighofer2006,Tanaka2014}. This near-frontal viewpoint matters in practice because a robot-mounted camera often approaches a target approximately along the marker normal. It is also the view in which a planar tag is easiest to detect and decode, because its image is largest and least foreshortened, so the pose is least reliable exactly where detection is most reliable. Adding more coplanar tags does not remove the ambiguity, because the enlarged array is still a single plane seen head-on, and enlarging a planar board is unattractive for small labware. Non-coplanar, multi-face marker geometries are an effective remedy~\cite{Mateos2020,Garca-Ruiz2023}, as discussed further in Section~\ref{sec:related}. However, they raise structure above the marked face, and that face may be the only one available to a parallel-jaw gripper. Improving pose estimation through geometry can therefore cost exactly the flat surface that grasping requires.

A laboratory fiducial must therefore do more than estimate pose accurately: it must remain compatible with manipulation. When the marked face doubles as the gripper's contact face, pose-estimation performance and graspability have to be designed together, alongside coded identification and low-cost fabrication paired with a detector model that matches the printed geometry. These are the four linked demands of Fig.~\ref{fig:research:overview}\mbox{(c)--(f)}. The problem addressed in this paper is thus to find a marker geometry that improves near-frontal pose estimation while preserving flat regions that a gripper can contact.

Our idea is to tilt multiple tags within one compact footprint. Even when the marker as a whole is viewed frontally, each tag is then observed at a non-frontal angle and supplies the perspective cues that a single planar tag loses near that view. At the same time, how the tags are tilted determines which flat regions of the face survive, so the tilt pattern itself becomes a design parameter that balances pose estimation against graspability.

Based on this insight, we propose the quARtet marker, a family of compact 3D-printable multi-tag fiducials. Four AprilTags are tilted within one square footprint, and all reliably detected corners are combined in a single Perspective-n-Point (PnP) solve, which recovers the marker pose from the correspondences between known 3D points and their 2D image projections~\cite{Lepetit2009} and is computed with OpenCV~\cite{Bradski2000}. Three layouts use the same detection principle but expose different flat regions for gripper contact, and a shared configuration drives both marker fabrication and reconstruction of the detector-side 3D corner coordinates, avoiding manual measurement of inter-tag geometry. The experiments compare the three layouts and a single-tag control in near-frontal pose estimation and in a physical grasp-retention test, making the pose-estimation--graspability trade-off of the layouts explicit.

Throughout this paper, \emph{pose estimation} is the umbrella term for estimating both orientation and position. The experiments report these two error components separately. The terminology does not denote two different estimation tasks.

The main contributions are:
\begin{enumerate}
    \item \textbf{A compact, configuration-driven multi-tag marker system.} Four tilted tags retain non-coplanar perspective cues within the footprint of one planar tag and are solved jointly. One parameter file defines both the editable computer-aided design (CAD) model and the detector geometry, and the toolchain and data are publicly available.
    \item \textbf{A controlled evaluation of the consistency--graspability trade-off.} Fixed-camera measurements under a common robot-referenced protocol and an independently re-grasped swing-down test separate the benefits and costs of the three layouts (Section~\ref{sec:system}) relative to the single-tag control. The layout whose four tags lean outward from a raised center is the most consistent across the tested viewpoints, whereas the two layouts whose tags tilt about their base edges keep flat contact strips for the gripper.
    \item \textbf{A closed-loop characterization under visual feedback.} A robot-mounted-camera pose-hold test with five runs per marker type shows that the near-frontal separation persists under visual feedback: every quARtet layout holds the target with a significantly smaller orientation error and step-to-step reorientation than the single-tag control.
\end{enumerate}

Section~\ref{sec:related} positions the work relative to geometric and detection-side alternatives. Section~\ref{sec:system} describes the marker and shared-parameter pipeline. Section~\ref{sec:eval} evaluates pose estimation and graspability against two research questions. Sections~\ref{sec:discussion} and~\ref{sec:conclusion} interpret the trade-off, limitations, and practical layout choice.

\section{Related Work}\label{sec:related}
Square fiducial markers such as ARTag, ArUco, and AprilTag are widely used as practical visual references for robot perception, localization, and pose estimation~\cite{Fiala2005,Garrido-Jurado2016,Romero-Ramirez2018,Olson2011,Wang2016,Krogius2019}. As noted in Section~\ref{sec:intro}, it is their reliance on the perspective distortion of a single square outline that leaves them least reliable exactly under near-frontal views~\cite{Schweighofer2006,Tanaka2012,Tanaka2014}. Prior efforts to overcome this limitation can broadly be grouped into two directions: enriching the \emph{marker geometry} itself with multi-tag or three-dimensional configurations, and enriching the \emph{detection process} with stronger algorithms or specialized acquisition hardware. We review each direction in turn and clarify how the quARtet marker positions itself with respect to them.

\subsection{Multi-Tag and 3D Marker Geometries}
To overcome the limited geometric information of a single planar tag, several studies have enriched marker geometry with multi-tag and three-dimensional configurations. Multi-tag targets and marker maps have been used to obtain stronger geometric constraints for camera calibration~\cite{Richardson2013} and for localization and visual simultaneous localization and mapping (SLAM)~\cite{Wang2023}, while flexible tag layouts extend tag shape and data density beyond the standard square~\cite{Krogius2019}. ChArUco boards combine marker coding with checkerboard geometry to support accurate calibration and surface estimation~\cite{Hu2019,Puggelli2020,An2018,Lin2022}. However, the tags on any one such board remain coplanar, so from a single view the near-frontal ambiguity persists at the board level and can be mitigated only by enlarging the footprint, which small labware cannot afford. The quARtet marker instead obtains non-coplanar corner geometry within a single compact ($35$~mm) footprint.

Beyond planar arrays, multi-face and object-specific fiducials extend this idea to three-dimensional surfaces or custom-shaped substrates, including dynamic three-dimensional tag arrangements for swarm robotics in highly reflective environments~\cite{Mateos2020}, pen-shaped 6-DOF marker tools~\cite{Le2022}, cylinder-specific designs based on projective invariants~\cite{Wang2024}, and other application-tailored fiducial objects~\cite{Garca-Ruiz2023}. These works show that spreading tags in space improves pose estimation across viewpoints.

Although effective, prior 3D and multi-tag designs typically pay a price in fabrication complexity, physical bulk, or application specificity. Volumetric and object-tailored markers often demand dedicated assembly or per-object redesign, which limits their reuse across heterogeneous laboratory targets and replaceable fixtures. The quARtet marker is intended to retain the geometric benefits of multi-tag/3D fiducials while remaining a compact, low-profile plate printable on standard fused-deposition-modeling (FDM) printers. Tiling four tilted AprilTags on a single compact plate-like base and exposing the geometry through a shared parameter set allows the marker geometry to be regenerated for different sizes and layouts without re-measuring inter-tag relationships, lowering the practical cost of adopting a multi-tag design.

\subsection{Augmented Detection Process}
A complementary line of work strengthens the detection process itself rather than the marker geometry, aiming to recover more reliable poses from challenging observations. On the algorithmic side, successive AprilTag systems have improved detection robustness and efficiency for the standard single planar tag~\cite{Olson2011,Wang2016}, ArUco-based detectors have been accelerated for real-time use~\cite{Romero-Ramirez2018}, and learning-based pipelines such as Deep ChArUco recover marker corners under low light and severe motion blur~\cite{Hu2019}. On the hardware side, optical markers with lenticular (lens-array) or moir\'e (interference-pattern) structures provide increased sensitivity to tilt at the cost of specialized fabrication~\cite{Tanaka2012,Tanaka2014,Tanaka2015}, while small fiducials combined with dynamic camera-parameter control such as zoom and focus can achieve high accuracy under demanding viewing conditions~\cite{Haraguchi2023}.

Effective as they are, these augmented detection schemes typically depend on specialized optical materials, tightly tuned acquisition control, or learned models that must be retrained for new targets. The quARtet system instead augments detection at a purely geometric level: corner correspondences from all reliably visible tags are pooled into a single PnP solve over the shared marker geometry. Because the same parameter set defines the fabricated geometry and the detection-side 3D corner coordinates, this joint solve operates with standard 2D camera input, requires no per-marker measurement of inter-tag geometry and no retraining, and falls back, by design, to the remaining tags when some become occluded or unreliable (Section~\ref{sec:detection}).

\section{quARtet Marker System}\label{sec:system}

\subsection{System Overview}
\begin{figure}[t]
    \centering
    \includegraphics[width=\linewidth]{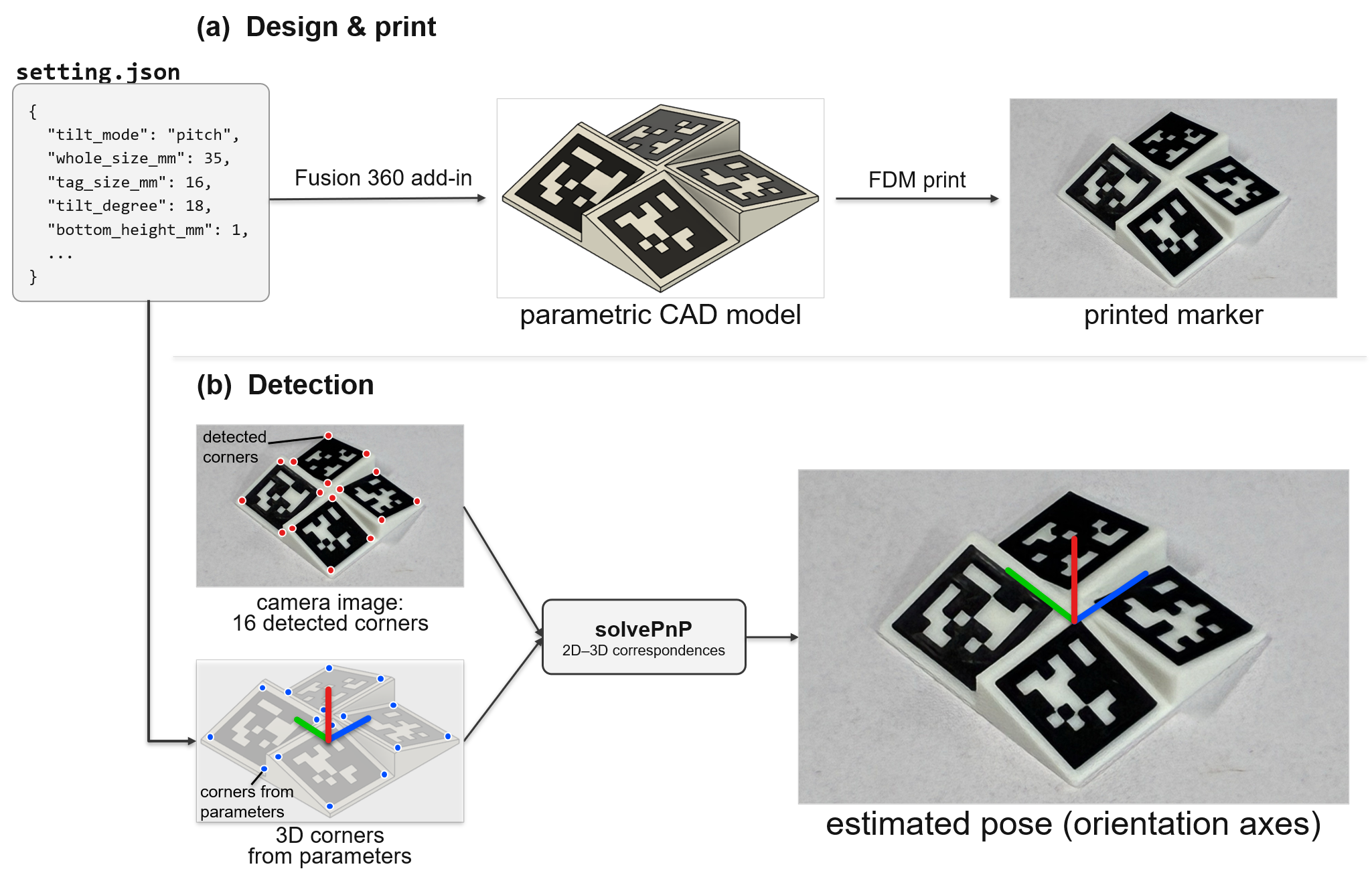}
    \caption{
    Overview of the quARtet system and its shared-parameter workflow, shown for the pitch layout (quARtet-P). (a)~Design and print: a single configuration file (\texttt{setting.json}) is read by an add-in for the CAD software Fusion~360. The add-in constructs the marker as an editable model, and the model is then fabricated by FDM printing. (b)~Detection: the 16 tag corners observed in the camera image are paired with the corresponding 3D corner coordinates reconstructed from the same parameters, and a single \texttt{solvePnP} call over these 16 2D--3D correspondences yields the marker pose, visualized by the orientation axes overlaid at the parameter-derived marker center.
    }
    \label{fig:flowchart}
\end{figure}

The quARtet marker is a 3D-printable multi-tag fiducial designed to improve pose estimation under near-frontal observation. The marker tiles four AprilTags on a single compact plate-like base, with each tag face embedded at a predefined tilt relative to the base plane (Fig.~\ref{fig:flowchart}(a)). The central idea (cf.\ the side views in Fig.~\ref{fig:layouts_overview}) is that even when the overall marker appears nearly frontal to the camera, at least part of the marker is likely to be observed under a more favorable non-frontal angle than would be obtained from a conventional single planar tag, providing the perspective cues that the PnP solver needs to recover orientation.
The proposed system comprises both the physical marker design and the corresponding detection procedure. These two sides are linked by a configuration-driven toolchain released with this paper (Fig.~\ref{fig:flowchart}): a single shared configuration defines the marker geometry for fabrication and reconstructs the same geometry for pose estimation, which keeps the fabricated marker consistent with the detection model and simplifies adapting the marker to different sizes, layouts, or object-specific constraints.
Three layouts were implemented in this study: diagonal, pitch, and reversed-pitch, denoted quARtet-D, quARtet-P, and quARtet-rP in the figures and experiments. Section~\ref{sec:design} details how their tilt patterns differ and which grasp surfaces remain.

\subsection{Marker Design}\label{sec:design}
\begin{figure}[t]
    \centering
    \begin{minipage}[b]{0.40\linewidth}
        \centering
        \includegraphics[width=0.95\linewidth]{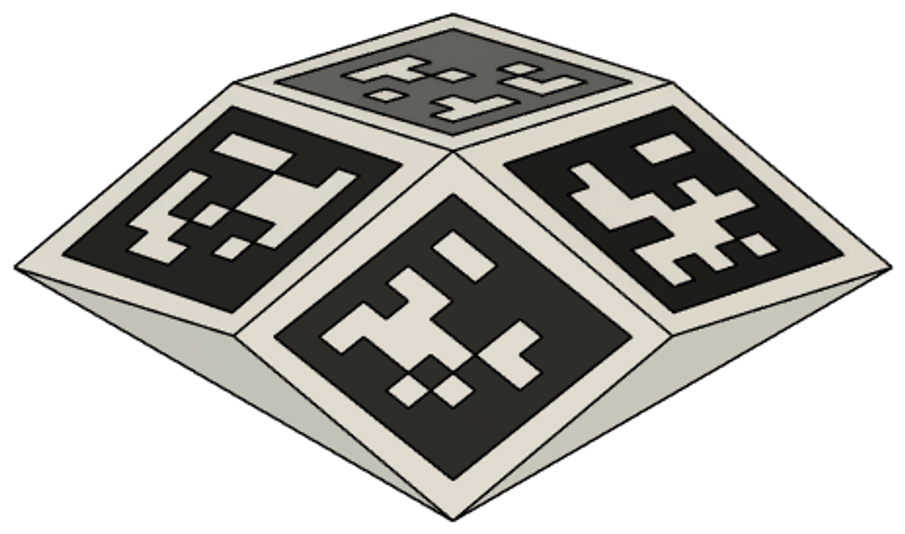}
        \subcaption{quARtet-D: 3D overview}
    \end{minipage}\hfill
    \begin{minipage}[b]{0.25\linewidth}
        \centering
        \includegraphics[width=0.62\linewidth]{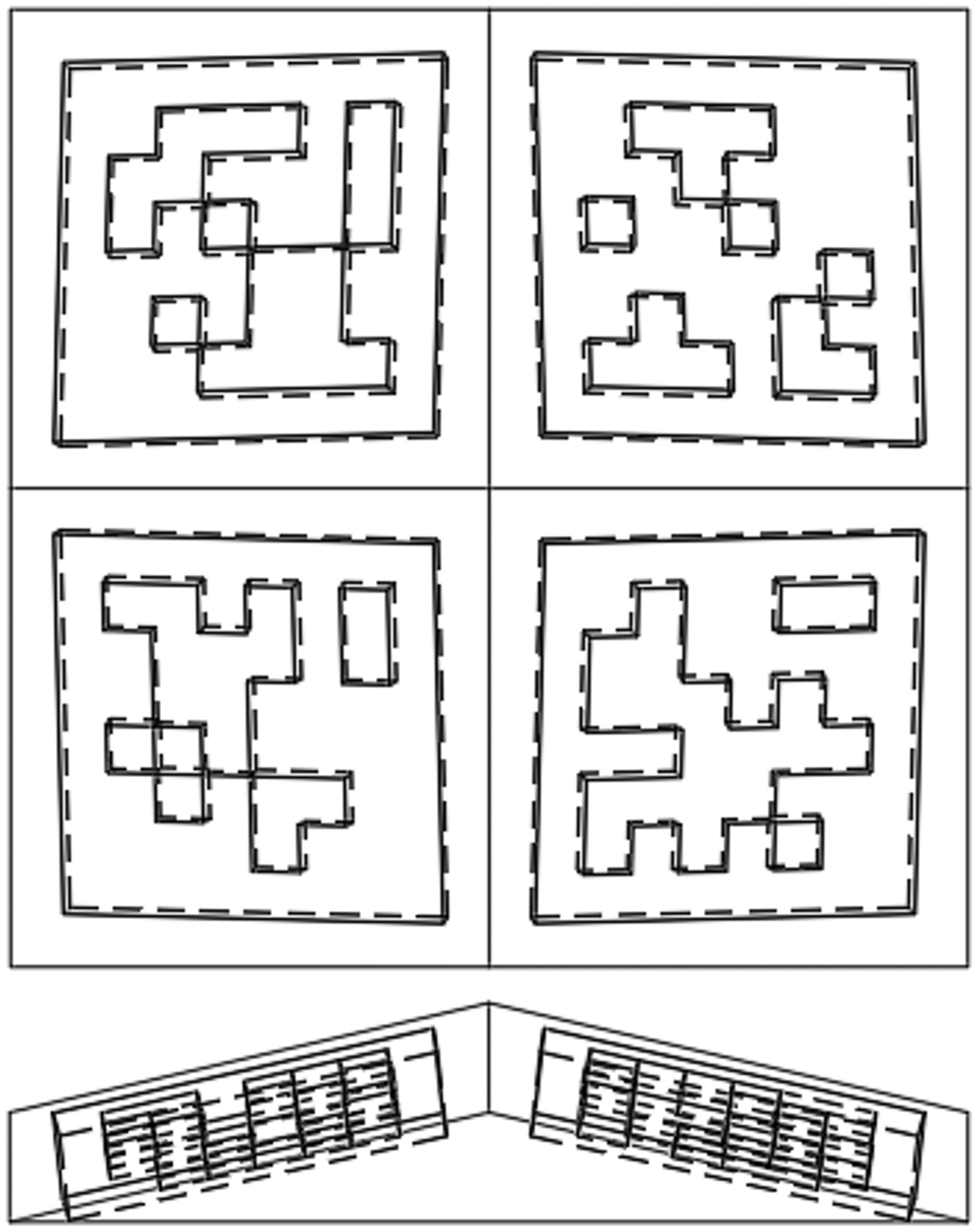}
        \subcaption{Top and side views}
    \end{minipage}\hfill
    \begin{minipage}[b]{0.30\linewidth}
        \centering
        \includegraphics[width=0.95\linewidth]{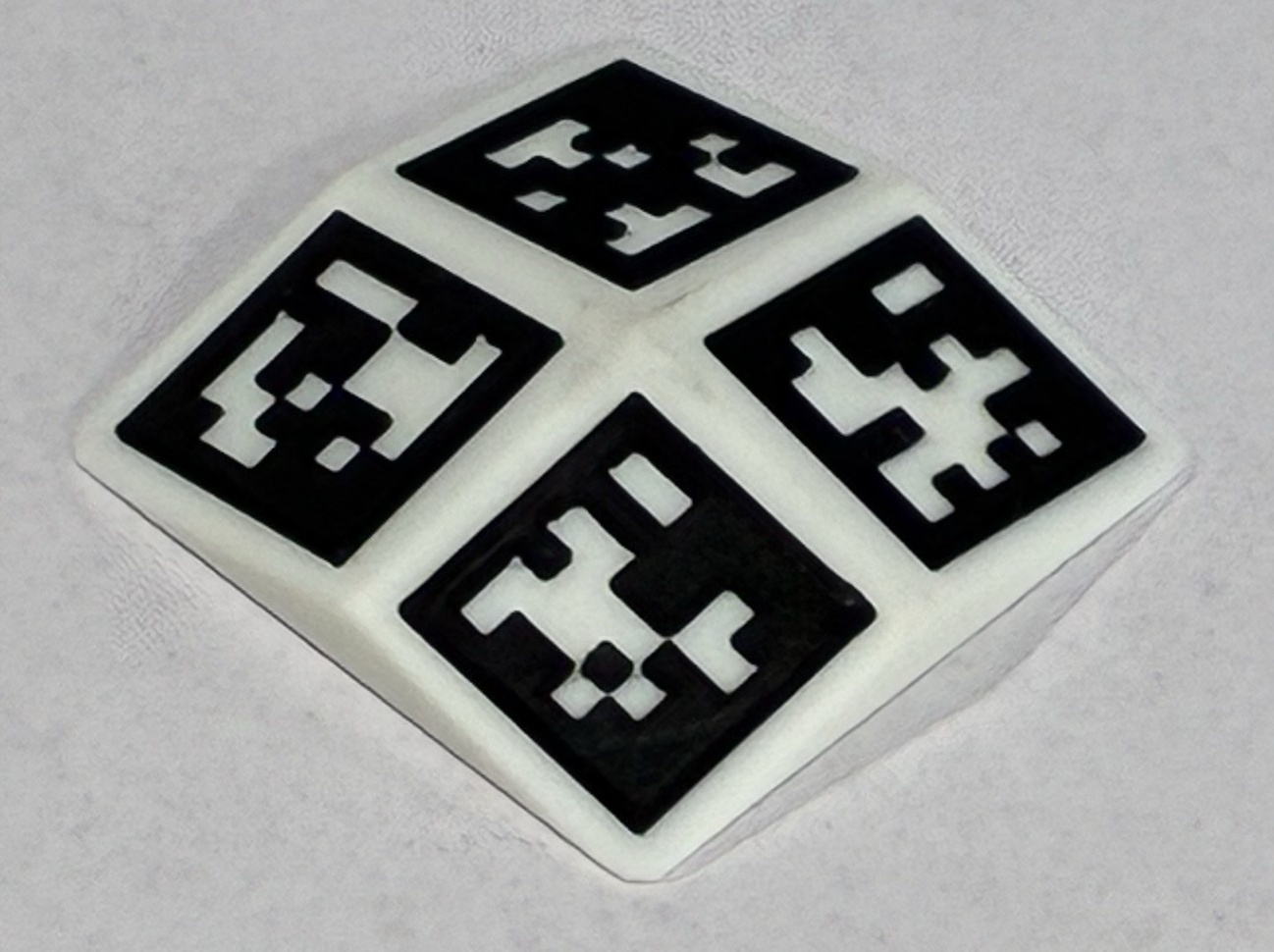}
        \subcaption{Printed marker}
    \end{minipage}
    \vspace{1ex}
    \begin{minipage}[b]{0.40\linewidth}
        \centering
        \includegraphics[width=0.95\linewidth]{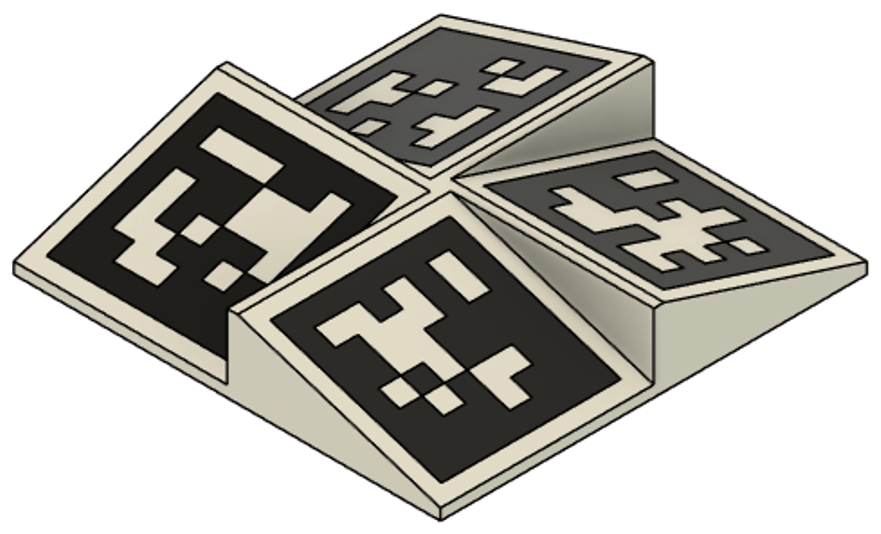}
        \subcaption{quARtet-P: 3D overview}
    \end{minipage}\hfill
    \begin{minipage}[b]{0.25\linewidth}
        \centering
        \includegraphics[width=0.62\linewidth]{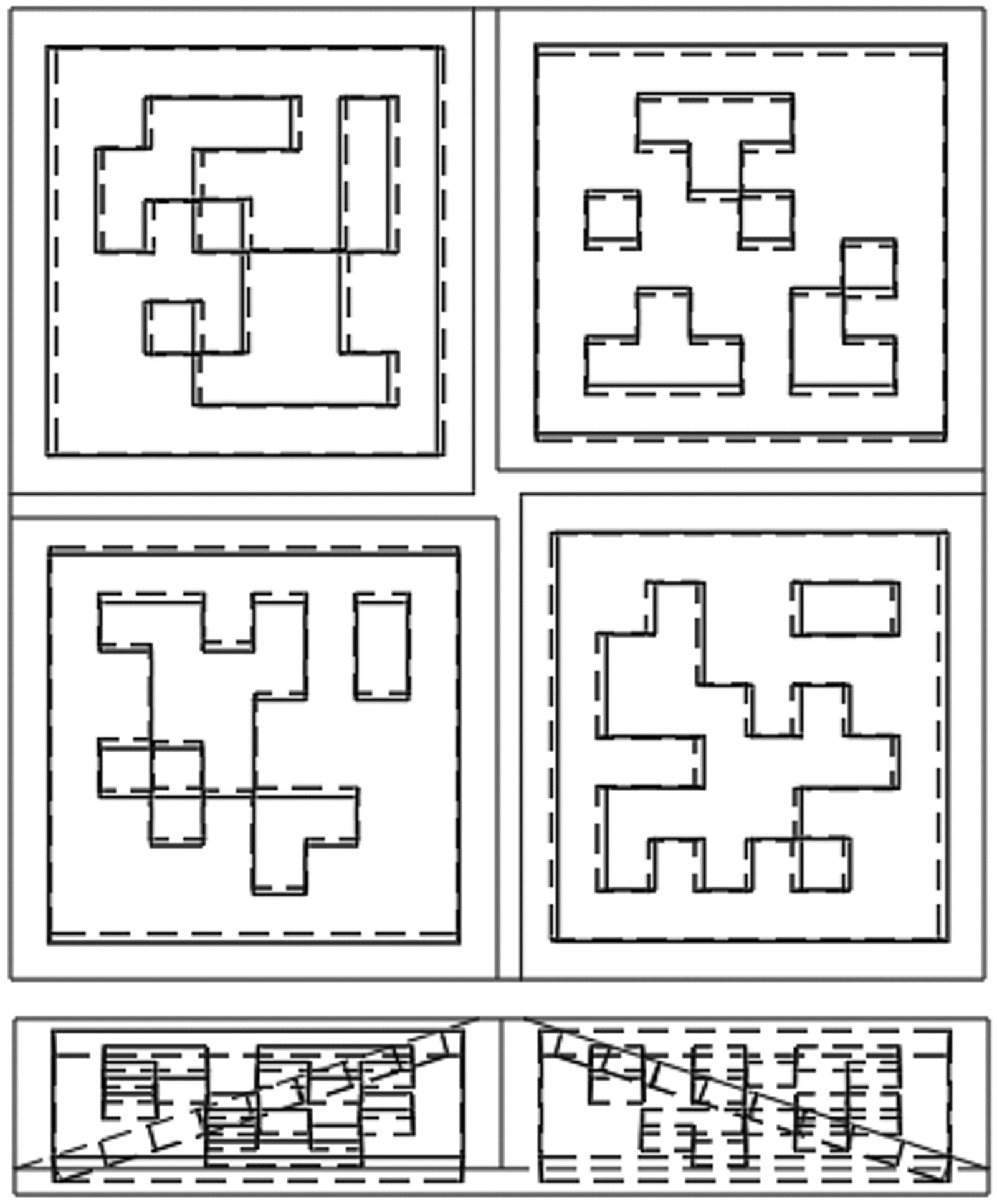}
        \subcaption{Top and side views}
    \end{minipage}\hfill
    \begin{minipage}[b]{0.30\linewidth}
        \centering
        \includegraphics[width=0.95\linewidth]{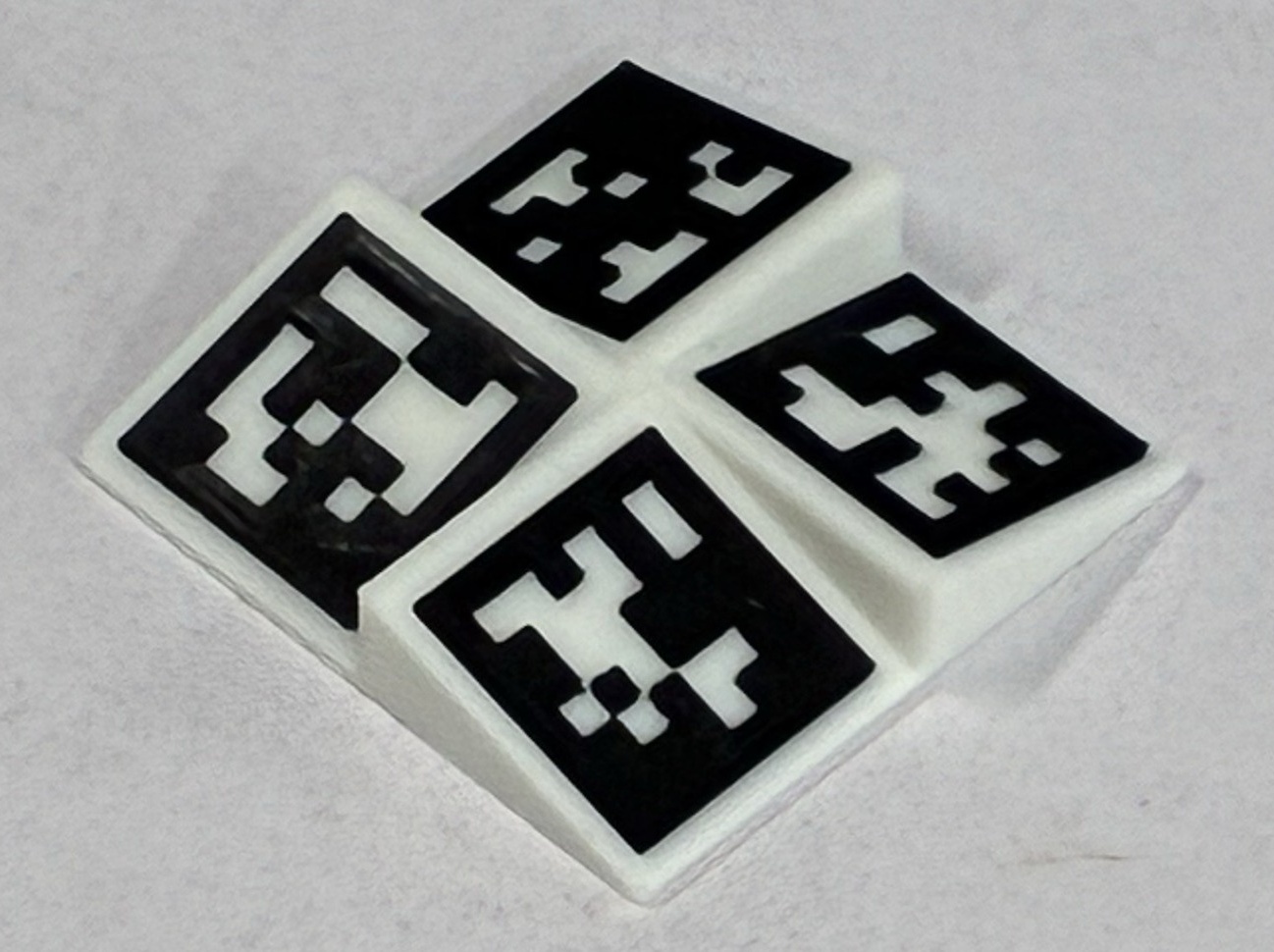}
        \subcaption{Printed marker}
    \end{minipage}
    \vspace{1ex}
    \begin{minipage}[b]{0.40\linewidth}
        \centering
        \includegraphics[width=0.95\linewidth]{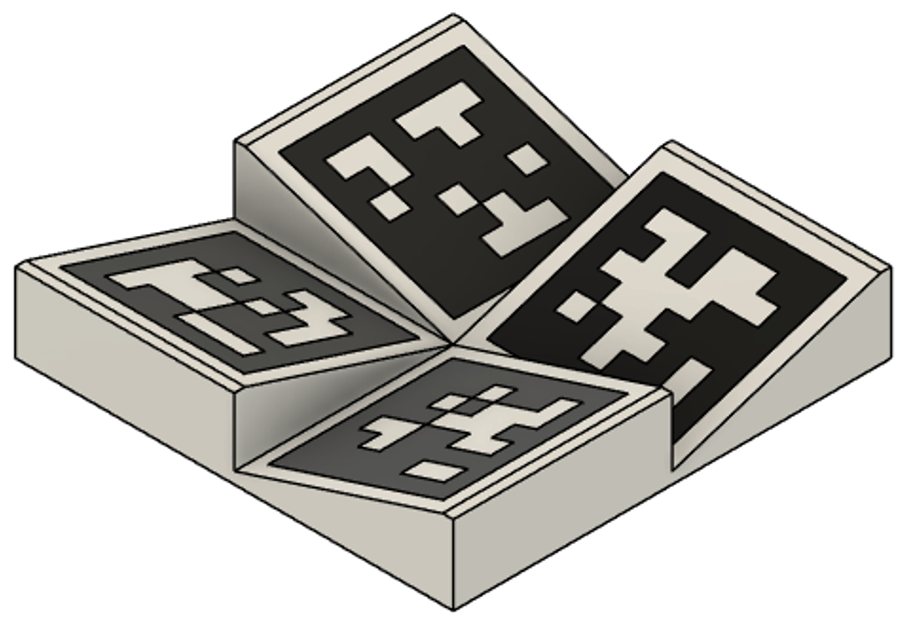}
        \subcaption{quARtet-rP: 3D overview}
    \end{minipage}\hfill
    \begin{minipage}[b]{0.25\linewidth}
        \centering
        \includegraphics[width=0.62\linewidth]{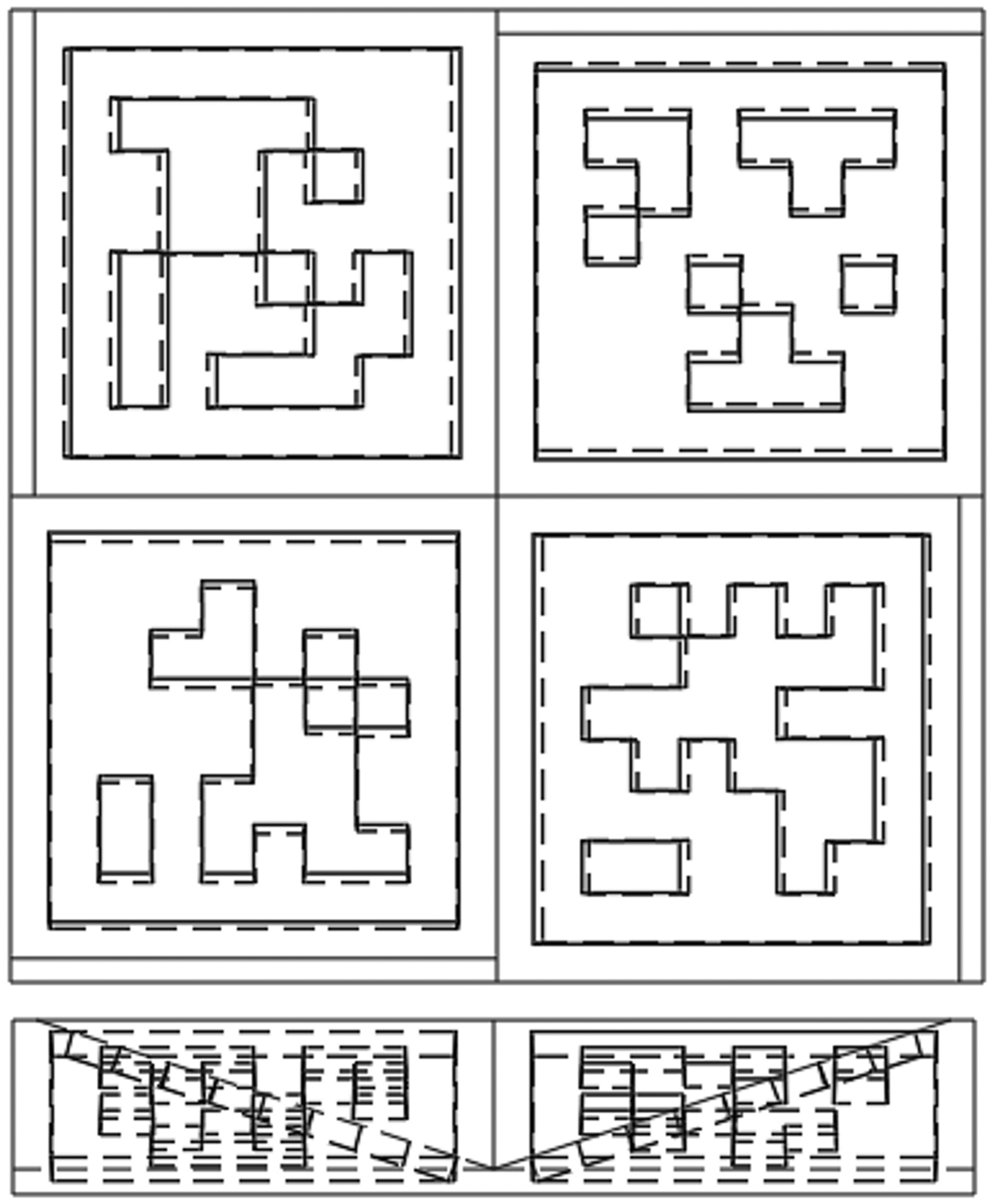}
        \subcaption{Top and side views}
    \end{minipage}\hfill
    \begin{minipage}[b]{0.30\linewidth}
        \centering
        \includegraphics[width=0.95\linewidth]{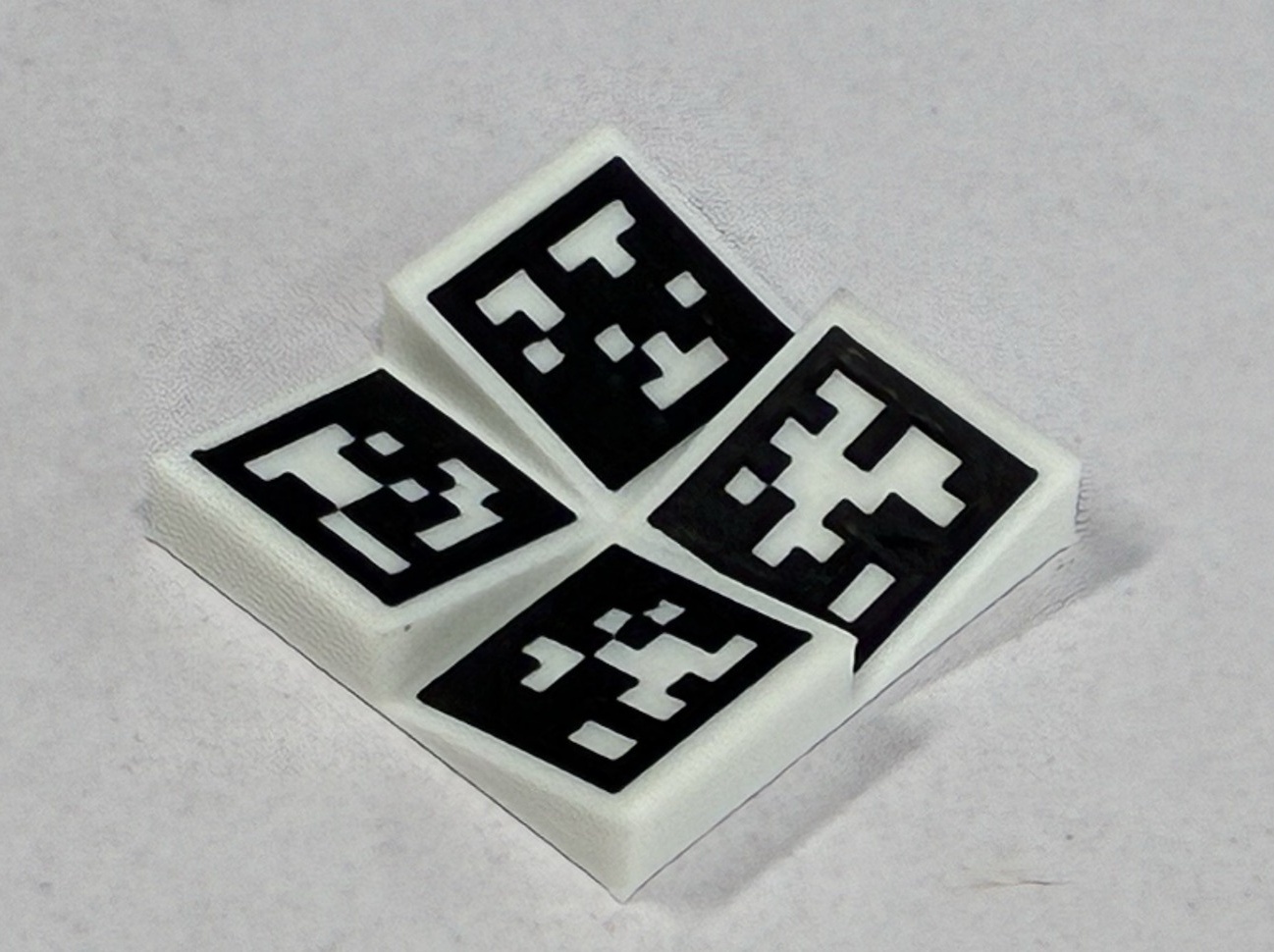}
        \subcaption{Printed marker}
    \end{minipage}
    \caption{
    The three implemented quARtet layouts, one per row: (a)--(c)~quARtet-D (diagonal), (d)--(f)~quARtet-P (pitch), (g)--(i)~quARtet-rP (reversed-pitch). Each layout is shown as a 3D overview rendering (left), orthographic top and side views (center), and a photograph of the FDM-printed marker (right). The top views show the arrangement of the four tags on the shared square footprint, and the side views show their out-of-plane tilt ($18\tcdegree$ in the present designs). The layouts share the four-tag concept and differ in how the tags are tilted and, consequently, in the graspability of the marker face. Geometric parameters are summarized in Table~\ref{tab:params}.
    }
    \label{fig:layouts_overview}
\end{figure}

Figure~\ref{fig:layouts_overview} shows the three implemented layouts (diagonal, pitch, and reversed-pitch) as 3D overview renderings, orthographic top and side views, and photographs of the FDM-printed markers. All layouts tile the four tags on the same square footprint. They differ in how the tags are tilted, which determines both how non-frontal perspective cues are exposed and which flat regions of the marker face remain accessible for grasping. The configurable parameters that define these layouts (overall footprint, tag size, tilt angle, and the related fabrication quantities) are defined in Table~\ref{tab:params}. The values used in this study are given in Section~\ref{sec:hardware}, and the resulting 3D corner coordinates are tabulated in Supplementary Section~S3.4.

In the diagonal layout, each tag is tilted about a diagonal axis so that the marker surface rises toward the center and the four tag faces lean outward in four different directions. This arrangement tends to reduce mutual occlusion and improve visibility from off-axis viewpoints, making it favorable for consistent pose estimation. In the pitch layout, each tag is tilted about one of its base edges, and the four tag blocks (a tag together with its supporting wedge) are arranged at successive $90\tcdegree$ in-plane rotations. Compared with the diagonal layout, this configuration preserves longer straight edge regions that remain more accessible to planar-finger contact. The reversed-pitch layout is constructed from the pitch layout by rotating each tag block by $180\tcdegree$ in the base plane about a vertical axis fixed in its quadrant, which inverts the tilt direction and also shifts each tag position slightly (by $0.78$~mm in the implemented geometry; Supplementary Section~S3.4). The rotation-based construction was chosen because simply reversing each tilt in place would thin the raised edges that serve as grasp surfaces. These pitch-based variants were introduced because, in practical robotic use, the marker geometry may need to balance pose-estimation performance against graspability and mounting convenience.

The base tilt used in the present designs was set to $18\tcdegree$. This value was selected as a practical compromise between geometric diversity and fabrication simplicity: larger tilts can strengthen the non-frontal perspective cues, but they also increase the out-of-plane height of the marker and make the marker less low-profile. In this work, $18\tcdegree$ provided a compact and fabrication-friendly geometry while still introducing sufficient angular diversity for improved pose estimation. A more detailed simulation-based rationale for this choice is provided in Supplementary Section~S1.

\subsection{Detection and Pose Estimation}\label{sec:detection}
During operation, the marker is observed by a calibrated camera and processed using an AprilTag detector~\cite{Olson2011,Wang2016,Krogius2019}. The detector identifies the visible tags in the image and extracts the 2D coordinates of their four corners. A tag is used whenever the detector decodes it with no bit errors (Hamming threshold zero), and no further filtering is applied. The 2D corner observations are then paired with the corresponding 3D corner coordinates reconstructed from the shared marker configuration: the tag in slot $t\in\{0,1,2,3\}$, whose image is given by \texttt{AR\{0,1,2,3\}\_file\_path} in Table~\ref{tab:params} (tag ID $t$ in this study), occupies quadrant $t$ of the footprint at a $90\tcdegree t$ in-plane rotation, corner order follows the AprilTag convention, and the resulting coordinates for the implemented layouts are tabulated in Supplementary Section~S3.4.

Pose estimation is performed by solving a single PnP problem using the combined corner correspondences from all reliably visible tags. Because each AprilTag contributes four corners, the number of correspondences varies according to visibility, but all correspondences are interpreted within the same marker reference frame. This joint solve yields a single pose estimate for the entire marker, rather than independent estimates for individual tags. Because each quARtet instance is defined in the shared configuration by its four tag images, and therefore by their IDs, distinct markers (and hence distinct objects) are distinguished by ID within the same detection pipeline.

A practical consequence of referring every tag to the same marker frame is that the marker degrades gracefully when part of it is lost: if one or more of the four tags are missed, the pose can still be recovered by matching the remaining tags to their predefined 3D corners, without any change to the pipeline. If only one tag remains visible, the solve reduces to that of a single (tilted) planar tag and the near-frontal ambiguity can partially reappear. Tags are lost for ordinary reasons (occlusion by the gripper or by neighboring labware, a specular highlight that washes out part of the plate, or contamination of the printed surface) and, in the pitch and reversed-pitch layouts, also for a reason intrinsic to the geometry: the raised edge of each tilted tag stands beside the face of a neighboring tag, so it can shadow or partly occlude that neighbor when the marker is viewed obliquely from the corresponding direction. The diagonal layout avoids this because its four faces rise jointly toward the center, leaving no raised edge standing in front of a neighboring face.

All predefined 3D corner coordinates, and therefore the pose reported by the detector, are expressed in a common marker-fixed frame. Its origin is the center of the square marker footprint on the base plane of the tag geometry (the plane from which the four tags are tilted), with the $x$- and $y$-axes aligned with the tag grid and the $z$-axis normal to the base plane. The axes overlaid at the marker center in the detection views (Fig.~\ref{fig:flowchart}(b)) visualize this origin. The origin is defined purely by the tag geometry and does not depend on \texttt{bottom\_height\_mm} (Table~\ref{tab:params}): when a bottom substrate is added to stiffen the marker ($\texttt{bottom\_height\_mm} > 0$), the physical marker extends below the base plane, and the origin then lies embedded inside the marker body, one substrate thickness above the mounting face.

The tilted geometry keeps informative non-frontal cues available near a frontal view, and the joint solve suppresses the near-mirror ambiguity of any single tag.

\subsection{Marker Generation and Fabrication}\label{sec:fabrication}

Marker generation is implemented as a Python add-in for the CAD software Fusion~360 (Autodesk): the add-in reads the shared configuration file and automatically constructs the marker as a fully editable model in the Fusion~360 workspace, rather than exporting a fixed stereolithography (STL) mesh directly, so that the generated geometry can be inspected or adjusted before being exported for standard FDM printing. This editable-model design lets a generated marker be merged with other 3D-printed parts, such as object-specific mounting fixtures, into a single printable model before export. The present implementation supports both integrated two-color printing and separate-part fabrication, in which the tags are printed separately and then inserted into the base. The latter is supported through an inner-taper parameter that can be adjusted to printer-dependent fitting tolerances. As a representative example, the 35~mm quARtet markers of Section~\ref{sec:hardware}, printed in polylactic acid (PLA), have masses of 4.4--4.9~g depending on the layout and require several hours of printing on a standard FDM printer.

The shared-parameter toolchain provides two practical conveniences for deployment. First, because both the printable geometry and the detection-side 3D corner coordinates are derived from a single shared configuration file, the fabricated marker and the pose-estimation model are consistent by construction, and markers can be regenerated or resized without a separate manual post-fabrication step to measure or define inter-tag spatial relationships (Fig.~\ref{fig:flowchart}). Second, the pose-estimation code operates on standard 2D camera input, and both it and the marker generator are released together with the design files and data. Table~\ref{tab:params} defines the configurable parameters of this shared framework.

\begin{table}[t]
    \centering
    \caption{
    Configurable parameters of the shared-parameter generation-and-detection framework. A single \texttt{setting.json} containing these parameters is consumed by both the model-generation and detection pipelines.
    }
    \label{tab:params}
    \footnotesize
    \begin{tabular}{@{}p{4.0cm} p{8.2cm}@{}}
    \toprule
    Parameter name & Description \\
    \midrule
    \texttt{AR\{0,1,2,3\}\_file\_path} & File path for the input image of AprilTag~0--3 (one parameter per tag). \\
    \texttt{tag\_family} & Family name of the AprilTags. \\
    \texttt{tilt\_mode} & Marker layout type: ``diagonal'', ``pitch'', or ``pitch\_r'' (reversed-pitch). \\
    \texttt{whole\_size\_mm} & Overall dimension (mm) of the square defined by the outer tag corners. \\
    \texttt{tilt\_degree} & Rotational inclination (degrees) of each tag relative to the base plane. \\
    \texttt{tag\_size\_mm} & Edge length (mm) of an individual AprilTag. \\
    \texttt{bottom\_height\_mm} & Thickness (mm) of the bottom substrate added beneath the tag geometry to stiffen the marker; the marker-frame origin is independent of this value (Section~\ref{sec:detection}). \\
    \texttt{tag\_thickness\_mm} & Physical thickness (mm) of the tag's solid section during fabrication. \\
    \texttt{tag\_inner\_taper\_degree} & Inward taper angle (degrees) of the tag from its surface toward its interior. \\
    \texttt{image\_margin\_frame\_dots\_num} & Number of pixels trimmed from all sides of the source image to isolate the tag. \\
    \bottomrule
    \end{tabular}
\end{table}

\section{Experimental Evaluation}\label{sec:eval}
Of the four demands raised in the Introduction, coded identification and low-cost fabrication are met by construction in Section~\ref{sec:system}. The evaluation therefore targets the remaining two as separate design questions. Section~\ref{sec:rq1} first compares pose-estimation errors under controlled fixed-camera viewpoints and then reports a closed-loop pose-hold characterization with five runs per marker type. Section~\ref{sec:rq2} tests retention and in-grasp slip under a specified parallel-jaw swing-down protocol. Keeping these questions separate makes clear which experiment supports which claim. Raw estimates, alignment and calibration procedures, and complementary analyses are documented in Supplementary Sections~S1--S7 and referenced where used.

Accordingly, we formulate the following two research questions:
\begin{enumerate}
    \item[\textbf{RQ1.}] Under a common robot-referenced measurement protocol, does the quARtet marker estimate pose with smaller errors than a single planar tag, particularly under near-frontal observation? (Section~\ref{sec:rq1})
    \item[\textbf{RQ2.}] Under the specified parallel-jaw swing-down protocol, which quARtet layouts retain the grasped object, and how much in-grasp slip do they incur relative to the single-tag control? (Section~\ref{sec:rq2})
\end{enumerate}

\subsection{Common Hardware Setup}\label{sec:hardware}
Markers and supporting fixtures were designed using Fusion~360 and fabricated using the shared-parameter workflow of Section~\ref{sec:system}. Fabrication used a Raise3D E2 FDM printer with white PLA for the base and black PLA for the tag parts, a $0.1$~mm layer height, and $10\%$ infill. The markers used in the experiments were made by separate-part fabrication (Section~\ref{sec:fabrication}): the black tag parts were printed separately and inserted into the white base, with the inner taper providing the fit. Robot experiments used a Universal Robots UR5e 6-DOF arm. Fixed-camera experiments used a Keyence IV4-G600MA camera with an IV4-G120 controller and IV4-LG6M illumination module. The robot-mounted pose-hold experiment used the same camera without the illumination module. The camera was calibrated in OpenCV using a $16\times22$-intersection checkerboard with $8.12$~mm spacing~\cite{Zhang2000}. Tag detection used the pyapriltags AprilTag detector~\cite{Krogius2019}, and the joint pose was obtained by solving the PnP problem~\cite{Lepetit2009} with OpenCV \texttt{solvePnP} (\texttt{SOLVEPNP\_ITERATIVE})~\cite{Bradski2000}.

Four marker types were compared in the order used throughout the result displays: quARtet-D (diagonal), quARtet-P (pitch), quARtet-rP (reversed-pitch), and Single (single planar tag). All had a $35\times35$~mm footprint. Each quARtet marker used four 16~mm tag36h11 AprilTags (IDs 0--3), an $18\tcdegree$ base tilt, a 1~mm bottom thickness, a 1~mm tag thickness, and a $-1\tcdegree$ inner taper. Single used one 35~mm tag36h11 AprilTag (ID~0) covering the same footprint. The printed markers weighed 4.4~g (quARtet-D), 4.7~g (quARtet-P), 4.9~g (quARtet-rP), and 5.0~g (Single). quARtet-P and quARtet-rP are referred to jointly as the pitch-based layouts.

\FloatBarrier
\subsection{Pose-Estimation and Closed-Loop Evaluation}\label{sec:rq1}
RQ1 is evaluated at two levels: static pose-estimation errors under controlled viewpoints with a fixed camera, and closed-loop pose holding with a robot-mounted camera. Section~\ref{sec:setup_fixed} describes the settings of both, and Section~\ref{sec:results_rq1} reports the results.

\subsubsection{Settings}\label{sec:setup_fixed}
\begin{figure}[t]
    \centering
    \begin{minipage}[b]{0.49\linewidth}
        \centering
        \includegraphics[width=0.95\linewidth]{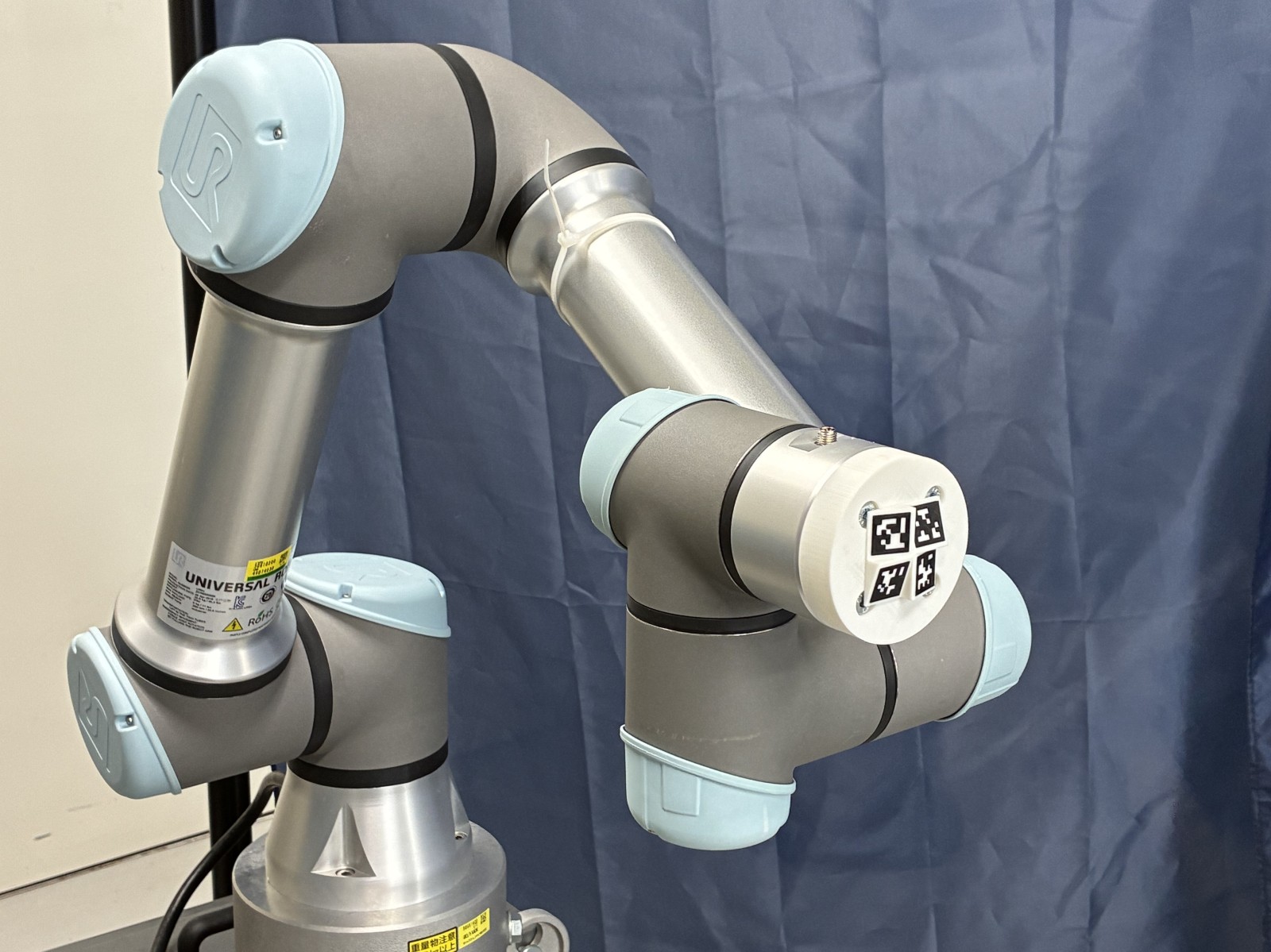}
        \subcaption{quARtet marker mounted on the end-effector}
        \label{fig:setup_fixed_camera:marker}
    \end{minipage}
    \begin{minipage}[b]{0.49\linewidth}
        \centering
        \includegraphics[width=0.95\linewidth]{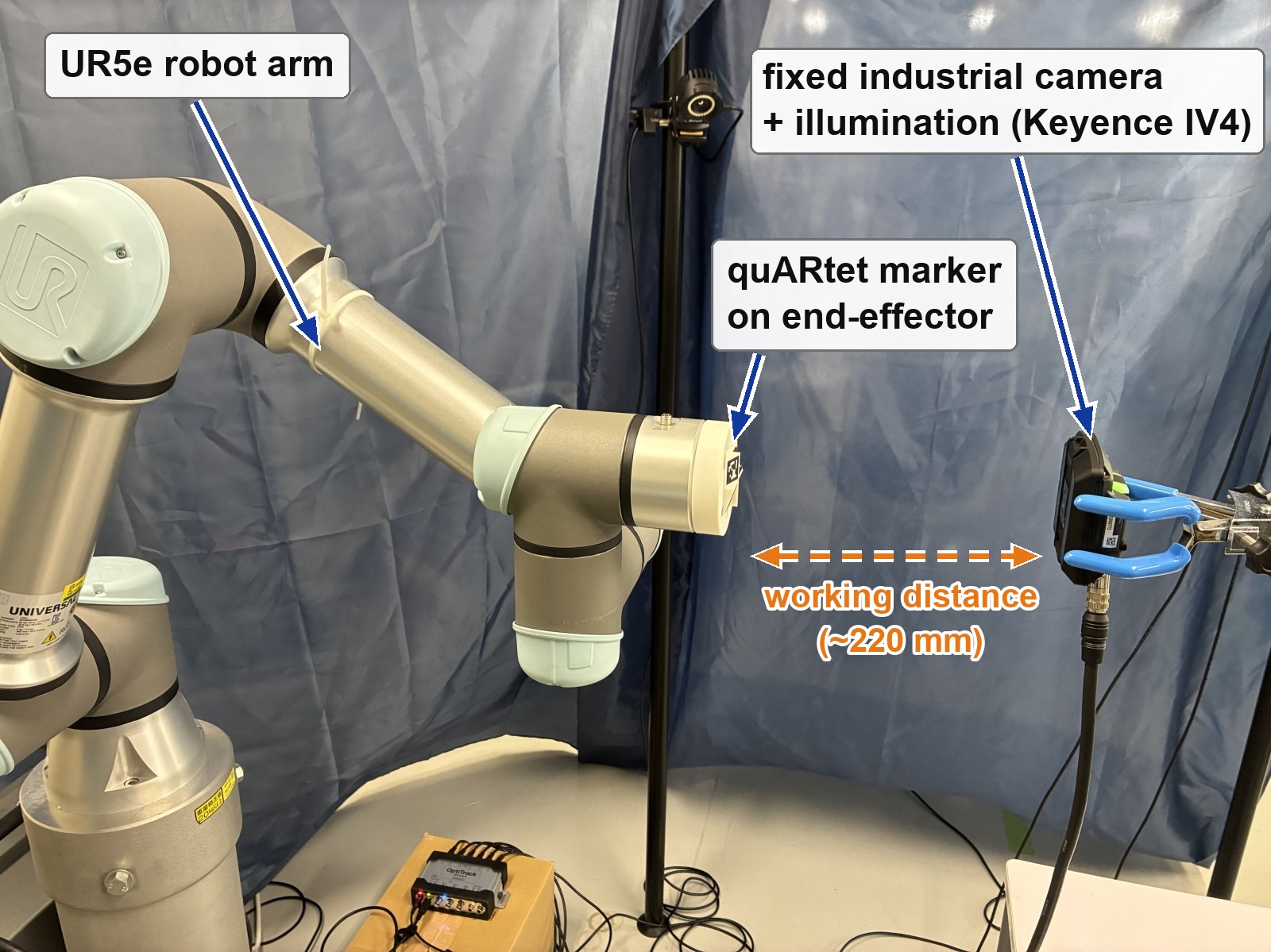}
        \subcaption{Fixed camera observing the marker-mounted end-effector}
        \label{fig:setup_fixed_camera:overview}
    \end{minipage}
    \caption{Fixed-camera setup for the pose-estimation evaluation. The robot provided repeatable changes in orientation and position while the industrial camera remained stationary.}
    \label{fig:setup_fixed_camera}
\end{figure}

\textbf{Fixed-camera setup.}
A fixed camera observed a marker mounted on the robot end-effector (Fig.~\ref{fig:setup_fixed_camera}). The robot changed the marker pose relative to the stationary camera, starting from a near-frontal pose at an approximately 220~mm working distance.

\textbf{Operational reference, alignment, and error definitions.}
Three elements define the evaluation.
\emph{(i)~Operational reference.} Each commanded tilt or translation is expressed relative to the initial frontal pose, and these robot-commanded relative changes serve as the common operational reference for comparing the marker types. They are not an independently calibrated ground truth: the arm's $\pm0.03$~mm specification is a translational repeatability figure, comparable to the smallest position variations observed here at a single commanded pose, and it neither establishes absolute translational accuracy at that scale nor quantifies the orientation-reference uncertainty. Sub-$0.1$~mm values and all orientation values should therefore be read within this robot-referenced protocol. Reference uncertainty is expected to be largely common-mode and is therefore less consequential for the relative comparison, although it remains unquantified.
\emph{(ii)~Constant frame alignment.} Because the camera frame and the command frame are not perfectly aligned, a single constant frame offset is estimated robustly per dataset (once for each marker type in each experiment) from all samples of that same dataset, and removed before errors are computed (Supplementary Section~S3.3). This is a coordinate registration, not an independent calibration or a held-out estimate: it removes the global frame mismatch, and with it any constant marker-specific bias, but its few shared degrees of freedom cannot remove pose-dependent effects across the 25 commanded poses of each experiment (the cells of the grids described below). All reported errors are therefore residuals to the commanded reference after this alignment. For brevity we call them orientation and position errors below.
\emph{(iii)~Image acquisition and error metrics.} At each commanded pose, the robot held the pose statically while ten images were captured consecutively, and all valid detections enter the analysis. Because these frames share one mounting and one held pose, they characterize within-acquisition estimator variation, not independent re-mounting replicates. Orientation error is the geodesic distance on the rotation group $\mathrm{SO}(3)$ between the estimated and the commanded relative orientation, i.e., the smallest rotation angle separating the two. Position error is the Euclidean distance between the estimated and the commanded relative translation. Samples whose orientation residual exceeds $12\tcdegree$ (pose flips) are excluded from the offset fit but retained in all reported errors. Supplementary Section~S3 records the full procedure.

\textbf{Orientation component.}
The commanded rotations about the $x$- and $y$-axes ($\theta_x$ and $\theta_y$, i.e., pitch and roll) were varied from $-30\tcdegree$ to $+30\tcdegree$ in $15\tcdegree$ increments, giving 25 commanded orientations. Ten images were acquired per marker at each orientation. Translation was held fixed. The exactly frontal cell ($\theta_x=\theta_y=0$) is used for the frontal-pose summary in Table~\ref{tab:summary}. A denser $\pm10\tcdegree$ grid in $5\tcdegree$ increments was also collected and is reported in Supplementary Section~S2.

\textbf{Position component.}
The robot was translated horizontally from $-20$ to $+20$~mm in 10~mm increments and in depth from $-40$ to $+40$~mm in 20~mm increments, giving 25 commanded positions. Ten images were acquired per marker at each position while orientation was held at the frontal pose.

\begin{figure}[t]
    \centering
    \setlength{\figclh}{0.45\linewidth}%
    \begin{minipage}[b]{0.38\linewidth}\centering
        \includegraphics[height=\figclh]{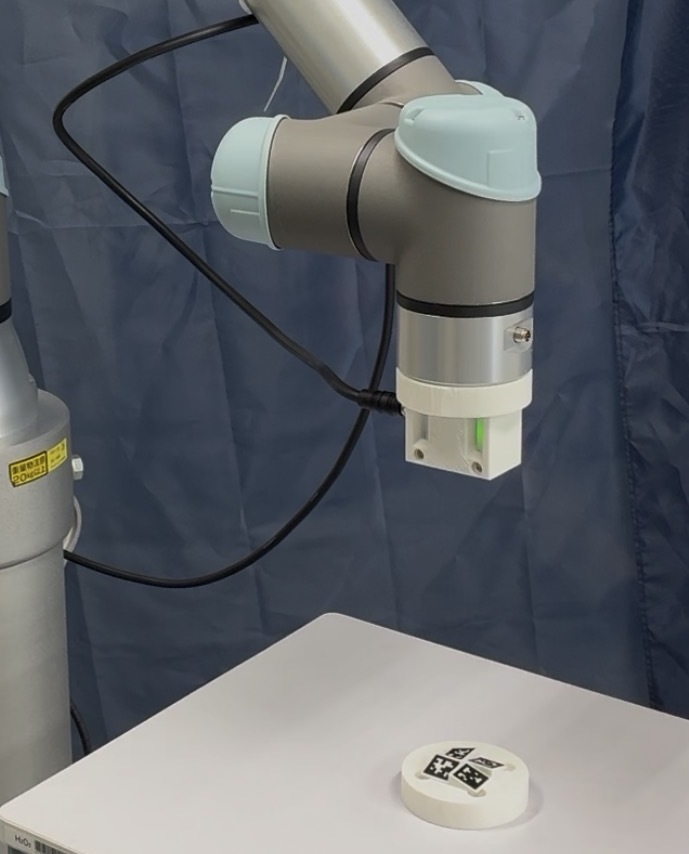}
        \subcaption{Robot-mounted camera facing the fixed marker}
    \end{minipage}\hspace{0.03\linewidth}%
    \begin{minipage}[b]{0.30\linewidth}\centering
        \includegraphics[height=\figclh]{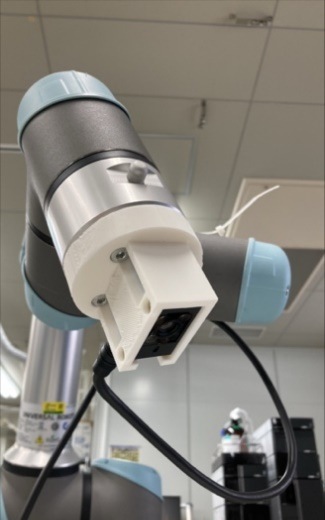}
        \subcaption{Camera fixture on the end-effector}
    \end{minipage}\par\vspace{0.6em}
    \begin{minipage}[b]{\linewidth}\centering
        \includegraphics[width=\linewidth]{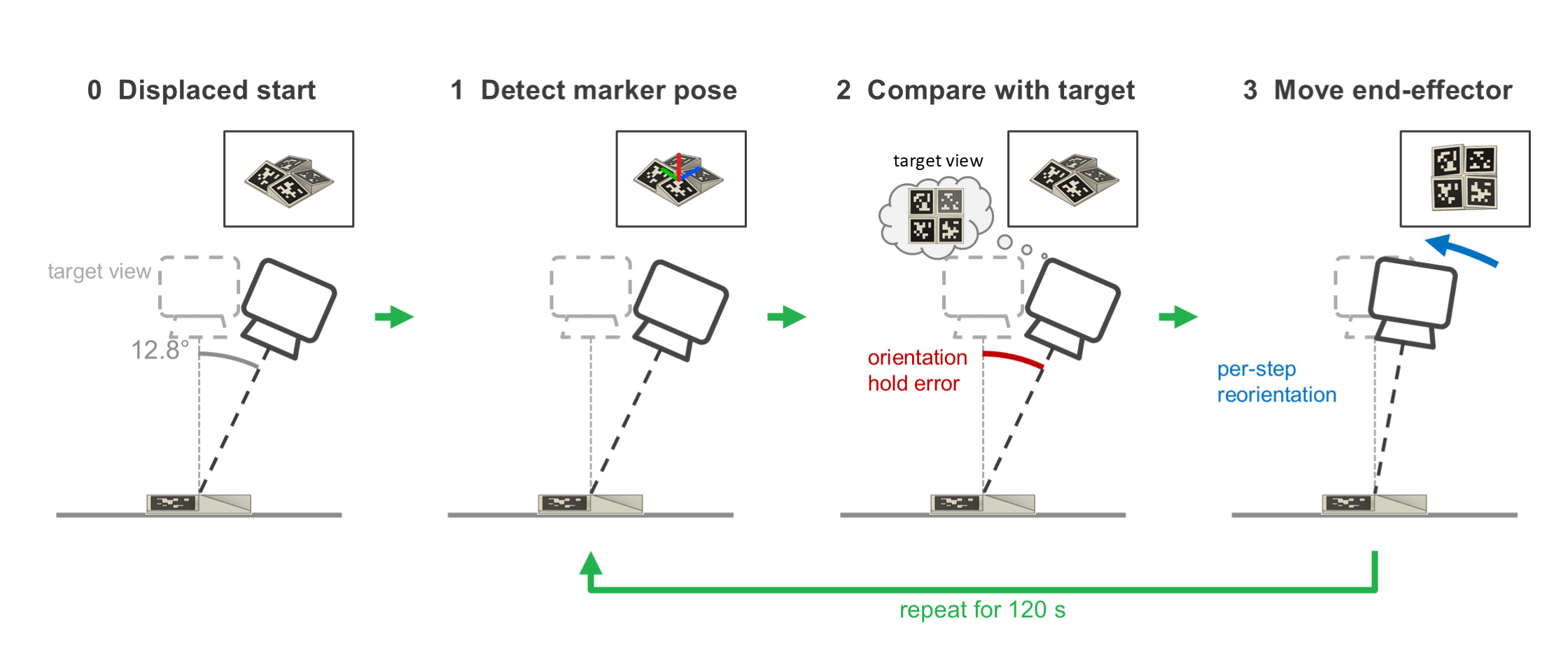}
        \subcaption{One cycle of the closed loop}
    \end{minipage}
    \caption{Closed-loop pose-hold setup and protocol. (a)~The camera is mounted on the end-effector and observes a quARtet marker fixed on the table. (b)~Close-up of the 3D-printed fixture that holds the camera on the end-effector. (c)~One cycle of the closed loop, drawn as a side view of the camera and the fixed marker with the current camera image as an inset. The initial near-frontal view of the marker is locked as the target (dashed camera). The camera is then orbited about the marker to the displaced start, $12.8\tcdegree$ off the marker normal with a $6\tcdegree$ rotation about the viewing axis, so that the marker stays in view (0). At every control step, the camera-to-marker pose is estimated from the current image (1) and compared with the locked target view (2), which gives the orientation hold error and the end-effector pose that restores the target view. The end-effector is moved there (3), and the rotation between consecutive end-effector poses is the per-step reorientation. Steps 1--3 repeat for 120~s.}
    \label{fig:setup_closedloop}
\end{figure}

\textbf{Closed-loop pose-hold setup.}
The closed-loop characterization asks whether the near-frontal separation observed with the fixed camera persists when pose estimates drive repeated robot corrections. The camera was mounted on the end-effector with a 3D-printed fixture and the marker was fixed on a table (Fig.~\ref{fig:setup_closedloop}). An initial near-frontal view at a camera-to-marker distance of about $200$~mm was locked as the target. The end-effector was displaced by a $12.8\tcdegree$ viewing-axis tilt plus a $6\tcdegree$ rotation about the viewing axis, then commanded to restore the target view and hold it for 120~s (Fig.~\ref{fig:setup_closedloop}(c)). At each control step, the camera-to-marker orientation error relative to the target (the orientation hold error) and the geodesic rotation between consecutive end-effector orientations (the per-step end-effector reorientation) were logged. Five runs were recorded for each of quARtet-D, quARtet-P, quARtet-rP, and Single, with the marker re-placed by hand between runs (approximately centered and near-frontal in the camera view) and the target view locked afresh at the start of every run, and the twenty logs are compared over their common ${\sim}117$~s window, the duration of the shortest run. Each run contributes one value per metric (its within-run mean), so the runs, not the control steps, are the replicates for the statistics. Hand--eye estimation and the full protocol are given in Supplementary Sections~S4 and~S5.

\subsubsection{Results}\label{sec:results_rq1}
\begin{figure}[t]
    \centering
    \begin{minipage}[b]{0.48\linewidth}\centering
        \includegraphics[width=0.7\linewidth]{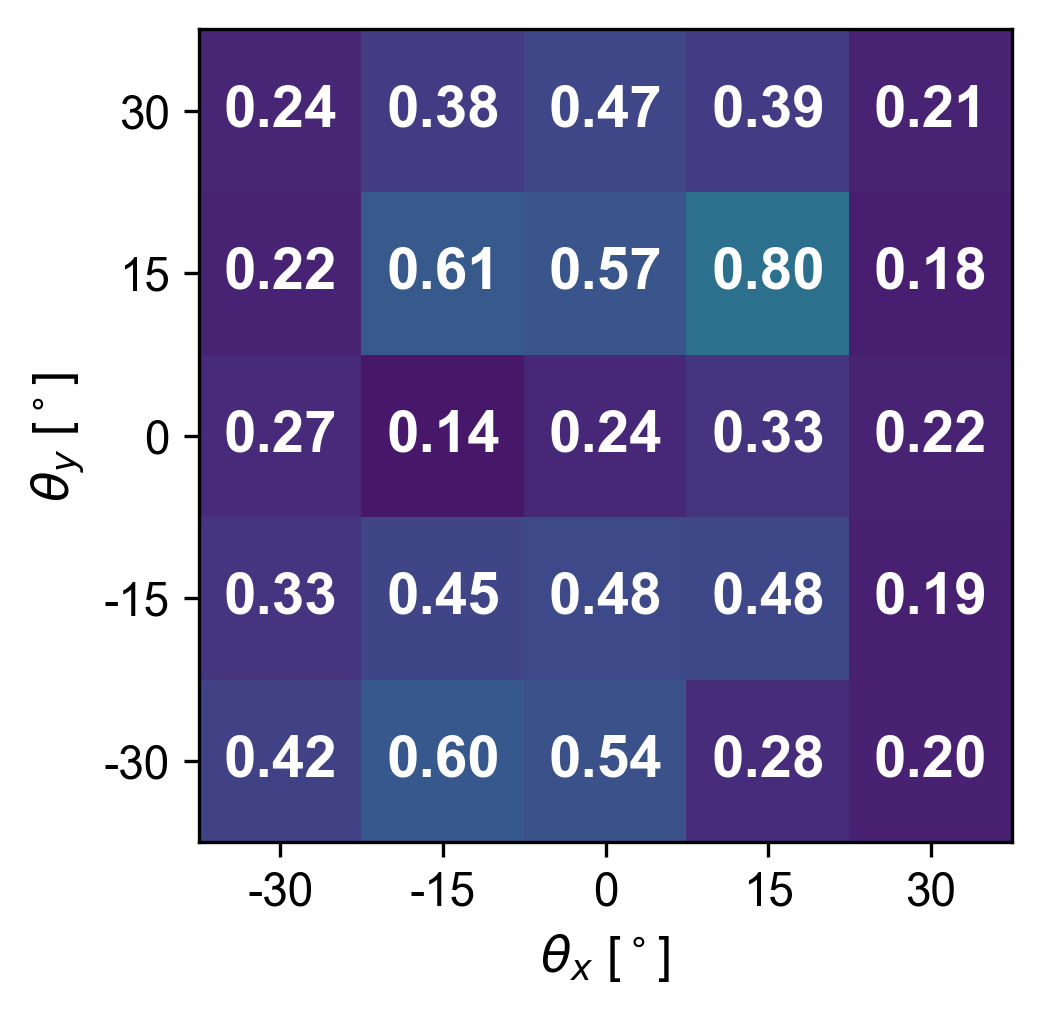}
        \subcaption{quARtet-D}\label{fig:orientation_broad:a}
    \end{minipage}\hfill
    \begin{minipage}[b]{0.48\linewidth}\centering
        \includegraphics[width=0.7\linewidth]{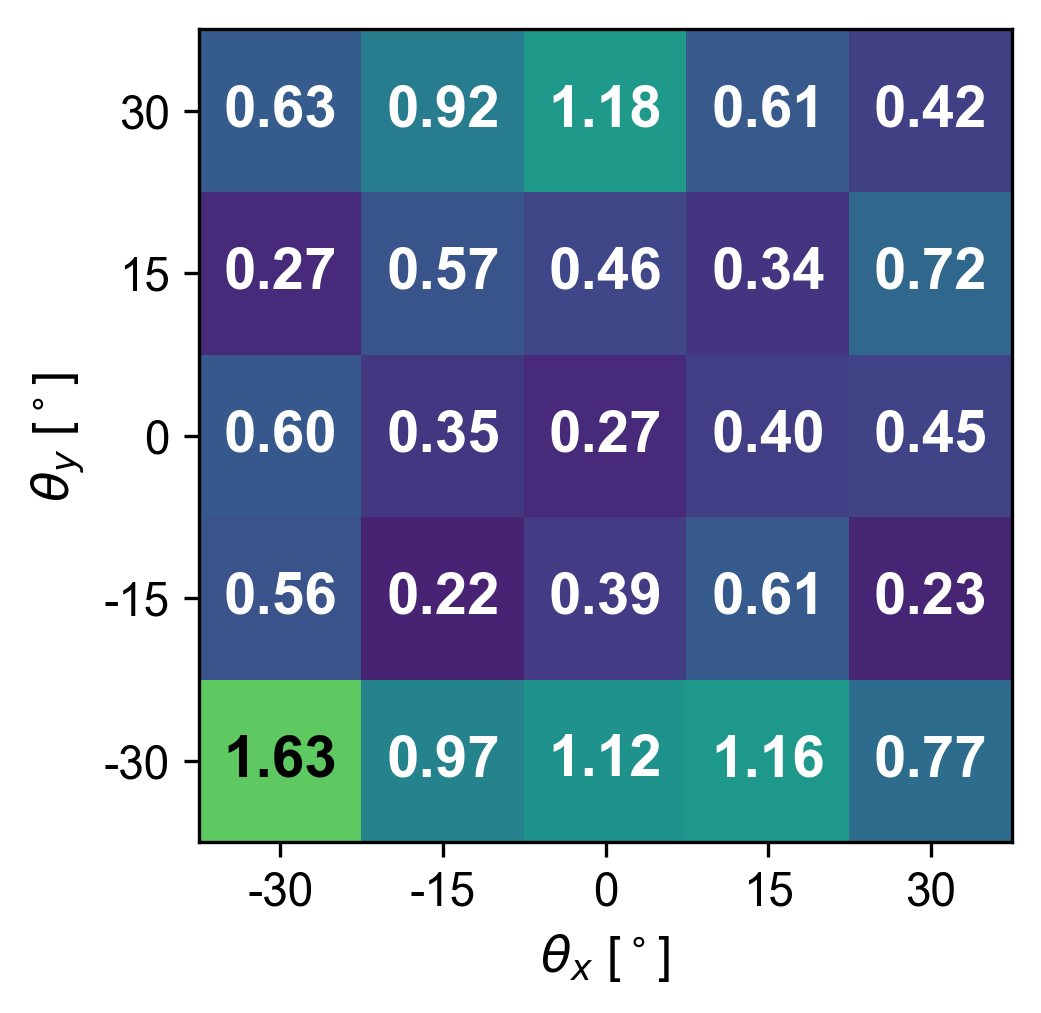}
        \subcaption{quARtet-P}\label{fig:orientation_broad:b}
    \end{minipage}
    \vspace{1ex}
    \begin{minipage}[b]{0.48\linewidth}\centering
        \includegraphics[width=0.7\linewidth]{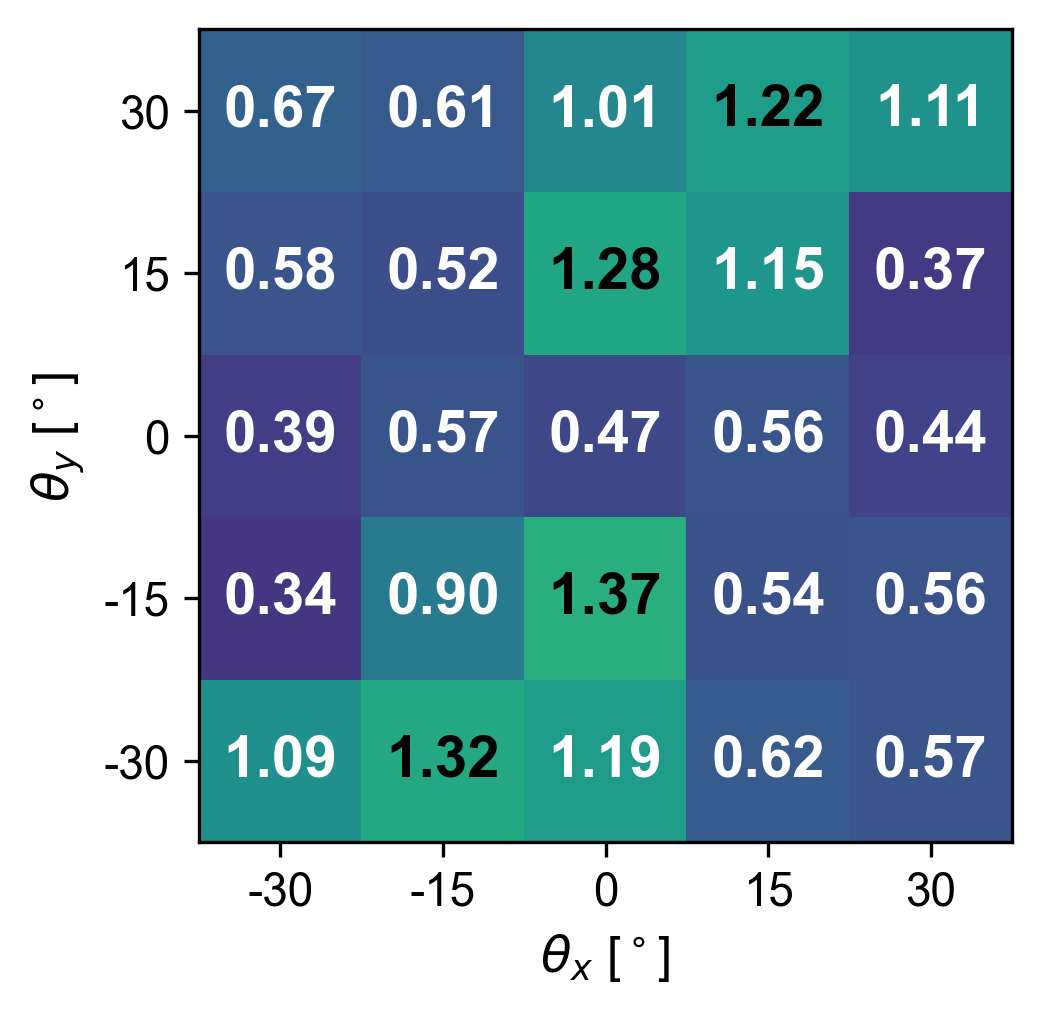}
        \subcaption{quARtet-rP}\label{fig:orientation_broad:c}
    \end{minipage}\hfill
    \begin{minipage}[b]{0.48\linewidth}\centering
        \includegraphics[width=0.7\linewidth]{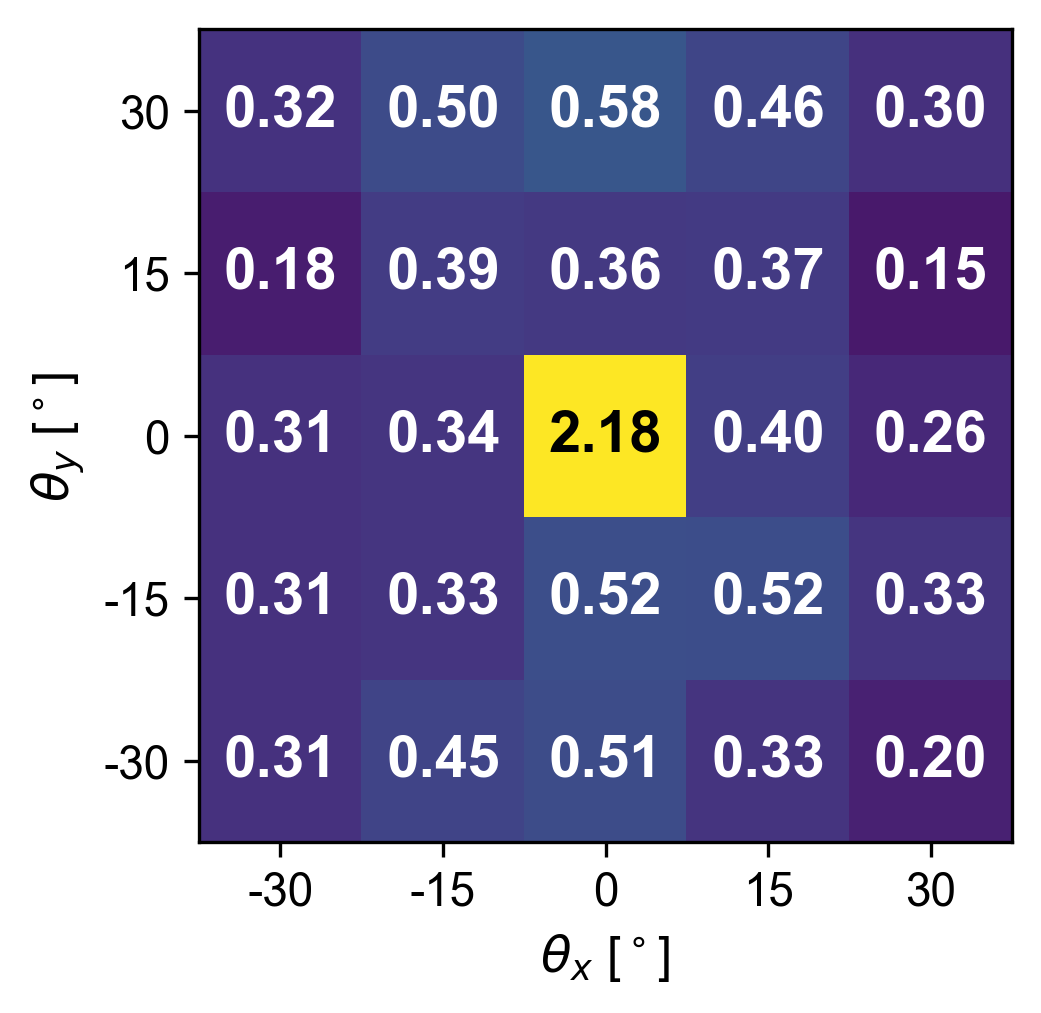}
        \subcaption{Single}\label{fig:orientation_broad:d}
    \end{minipage}

    \vspace{0.5ex}
    \includegraphics[width=0.5\linewidth]{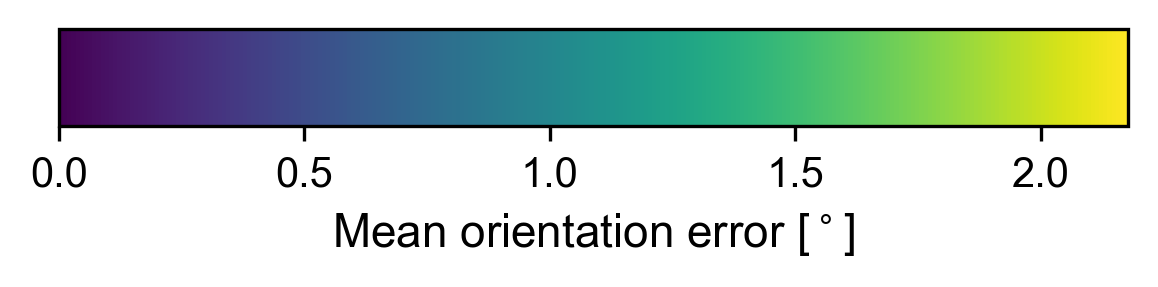}
    \caption{Mean orientation error per commanded cell over the broad tilt grid (up to 10 consecutive frames per cell, Supplementary Section~S3.1), with per-cell values annotated. Errors are relative to the robot-commanded reference after the constant alignment of Section~\ref{sec:setup_fixed}. Panels are ordered quARtet-D, quARtet-P, quARtet-rP, and Single, with one color scale shared by all. Frames within a cell are repeated images of one held pose, not independent re-mountings. Cell-wise summaries are given in Supplementary Section~S2.3.}
    \label{fig:orientation_broad}
\end{figure}

\textbf{Orientation component.}
Figure~\ref{fig:orientation_broad} maps the mean orientation error over the commanded tilt grid. The Single error is concentrated at the frontal view, consistent with the near-mirror ambiguity, and falls off once the marker is tilted by $15\tcdegree$ or more. quARtet-D remains consistent across the tested range, whereas quARtet-P and quARtet-rP show larger errors at some oblique views, where their raised edges can reduce tag visibility. Per-image scatter for both orientation grids is reported in Supplementary Section~S2.

\begin{figure}[t]
    \centering
    \begin{minipage}[b]{0.48\linewidth}\centering
        \includegraphics[width=0.7\linewidth]{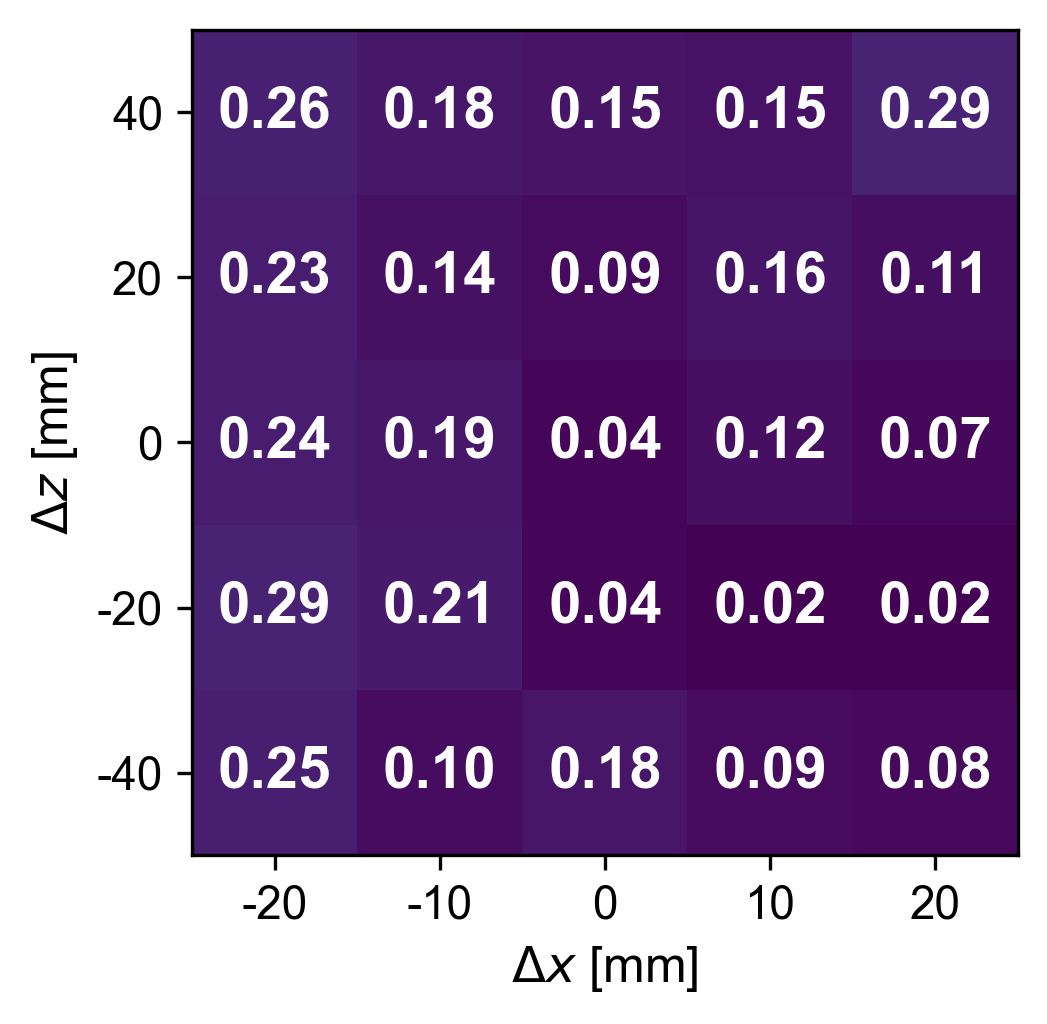}
        \subcaption{quARtet-D}\label{fig:position:a}
    \end{minipage}\hfill
    \begin{minipage}[b]{0.48\linewidth}\centering
        \includegraphics[width=0.7\linewidth]{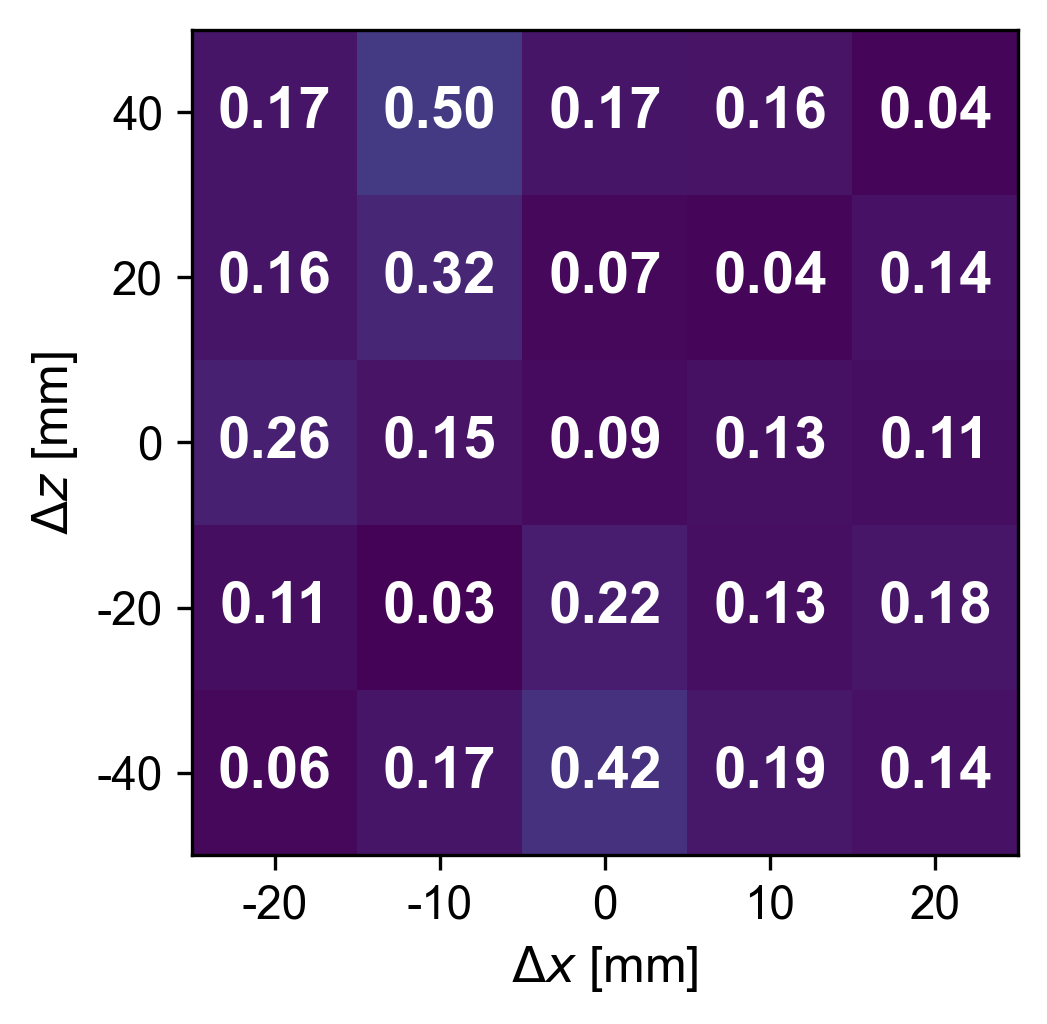}
        \subcaption{quARtet-P}\label{fig:position:b}
    \end{minipage}
    \vspace{1ex}
    \begin{minipage}[b]{0.48\linewidth}\centering
        \includegraphics[width=0.7\linewidth]{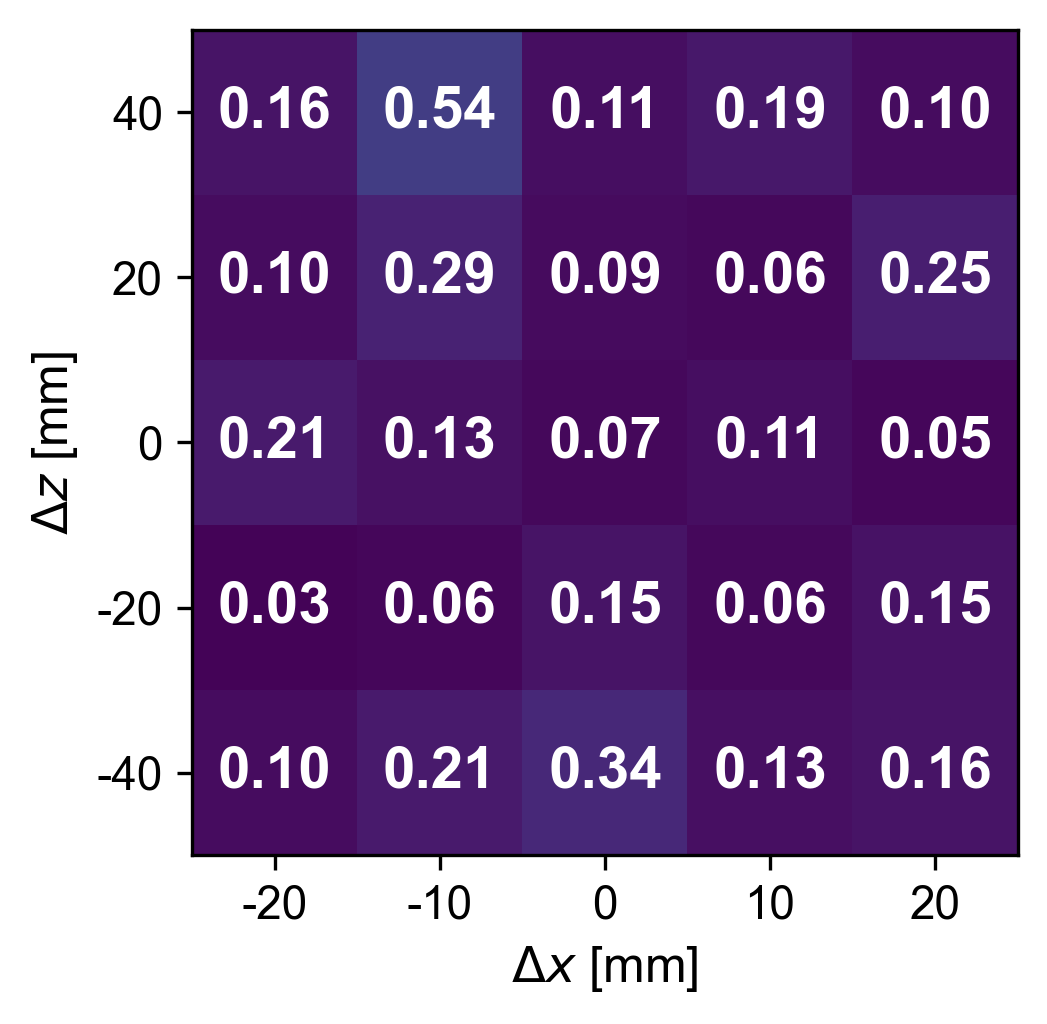}
        \subcaption{quARtet-rP}\label{fig:position:c}
    \end{minipage}\hfill
    \begin{minipage}[b]{0.48\linewidth}\centering
        \includegraphics[width=0.7\linewidth]{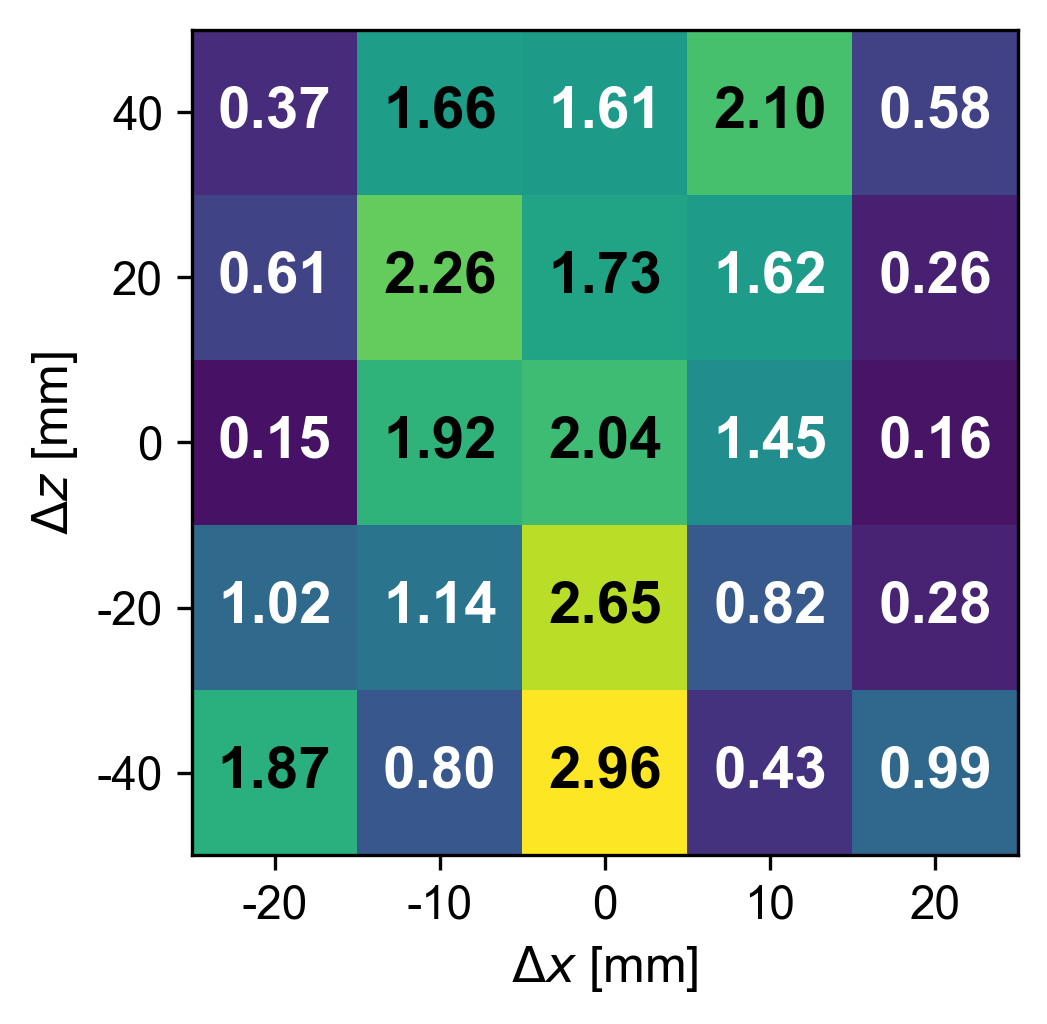}
        \subcaption{Single}\label{fig:position:d}
    \end{minipage}

    \vspace{0.5ex}
    \includegraphics[width=0.5\linewidth]{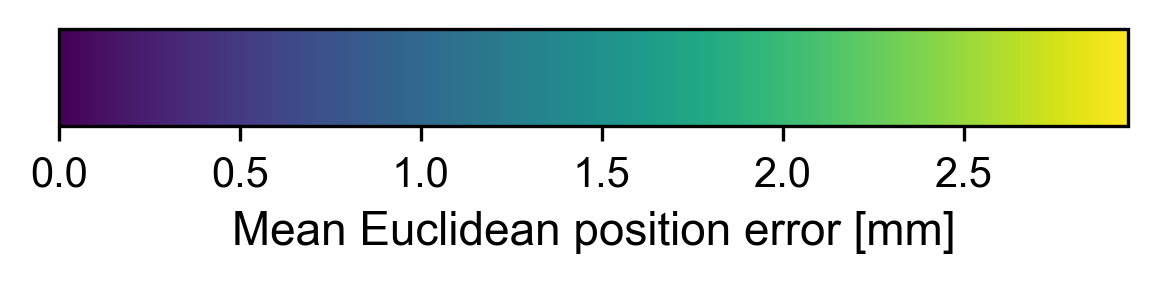}
    \caption{Mean Euclidean position error per commanded cell over the translation grid (up to 10 consecutive frames per cell, Supplementary Section~S3.1), with per-cell values annotated. Panels are ordered quARtet-D, quARtet-P, quARtet-rP, and Single, with one color scale shared by all. Errors are residuals to the robot-commanded relative translation after the same-dataset constant alignment of Section~\ref{sec:setup_fixed}. Cell-wise summaries are given in Supplementary Section~S2.3.}
    \label{fig:position}
\end{figure}

\textbf{Position component.}
Figure~\ref{fig:position} shows lower position errors for all quARtet layouts than for Single over almost the entire translation grid (one cell excepted, Supplementary Section~S2.3). The larger Single residual is concentrated around the frontal column, consistent with orientation ambiguity coupling into translation estimation. The fitted constant frame offsets included an in-plane rotation of approximately $-3.1\tcdegree$ in all four datasets. This agreement is compatible with a common camera-to-robot frame mismatch, but because each offset is estimated from the evaluated dataset itself, it is not an independent validation of that interpretation. Raw per-image estimates are reported in Supplementary Section~S2.

\FloatBarrier
The fixed-camera columns of Table~\ref{tab:summary} summarize both components against the Single control as descriptive statistics: group means, standard deviations (SDs), and the root-mean-square error (RMSE). At the frontal cell, Single produced a mean orientation error of $2.18\tcdegree$ with a large within-cell spread (SD $0.80\tcdegree$) consistent with the near-mirror ambiguity, whereas quARtet-D, quARtet-P, and quARtet-rP produced $0.24$--$0.47\tcdegree$. Over the translation grid, the RMSE was $0.17$--$0.20$~mm for the three quARtet layouts versus $1.50$~mm for Single, with a mean within-cell SD of $0.02$--$0.03$~mm versus $0.15$~mm (Supplementary Section~S3). Cell-wise mean$\pm$SD tables for every commanded condition, together with difference maps that show where each layout's errors are smaller or larger than Single's, are provided in Supplementary Section~S2.3. Away from the frontal view the comparison partly reverses: at oblique cells, where a planar tag already has small errors ($0.15$--$0.6\tcdegree$), quARtet-P and quARtet-rP have cell-mean orientation errors of up to $1.6\tcdegree$, and quARtet-D, though also above Single at ten of the 24 oblique cells (by up to $0.43\tcdegree$), stays at or below $0.8\tcdegree$ over the tested grid. These numerical separations are reported without frame-level significance claims because the ten consecutive images of a cell are technical repeats of one held pose (Section~\ref{sec:setup_fixed}). Independent re-mounting or re-acquisition runs would be required for inference beyond this setup.

The fixed-camera result answers the static part of RQ1 within the robot-referenced, same-setup protocol: all three quARtet layouts show markedly smaller near-frontal orientation and position errors than Single, at the cost of modest oblique-view degradations for the pitch-based layouts, and quARtet-D is the most consistent across the tested viewpoints. This conclusion concerns errors relative to the common operational reference after constant alignment. It does not establish absolute pose accuracy.

\begin{figure}[t]
    \centering
    \begin{minipage}{0.95\linewidth}\centering
        \includegraphics[width=\linewidth]{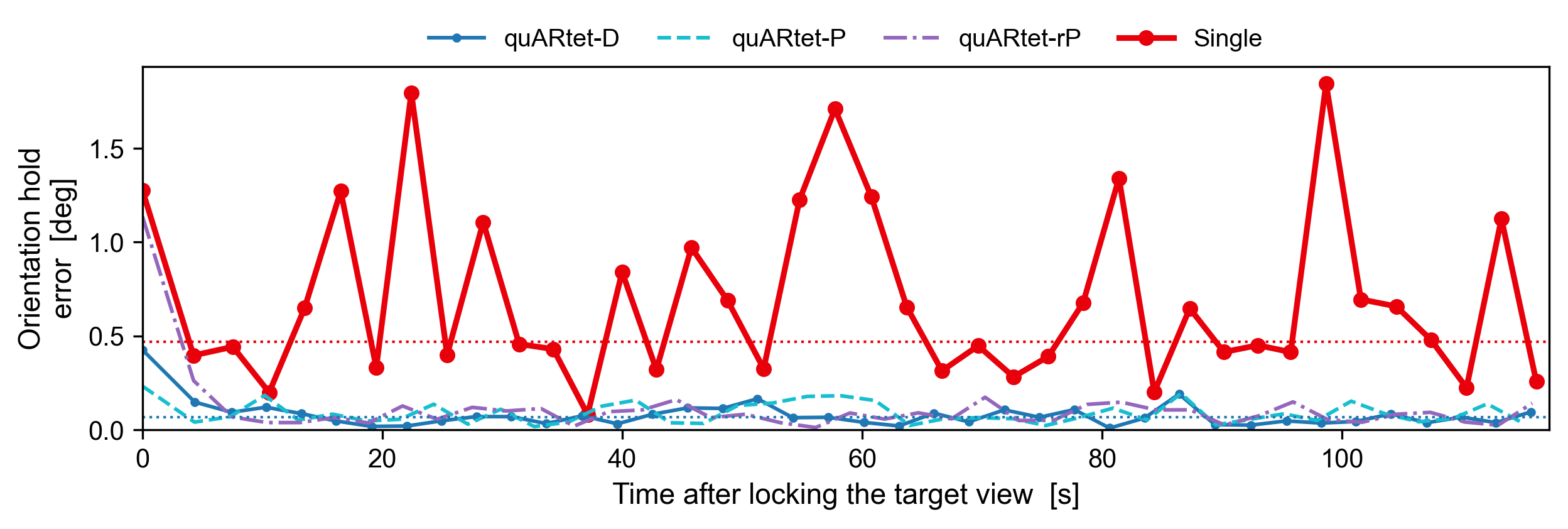}
        \subcaption{Orientation hold error}
    \end{minipage}
    \vspace{1ex}
    \begin{minipage}{0.95\linewidth}\centering
        \includegraphics[width=\linewidth]{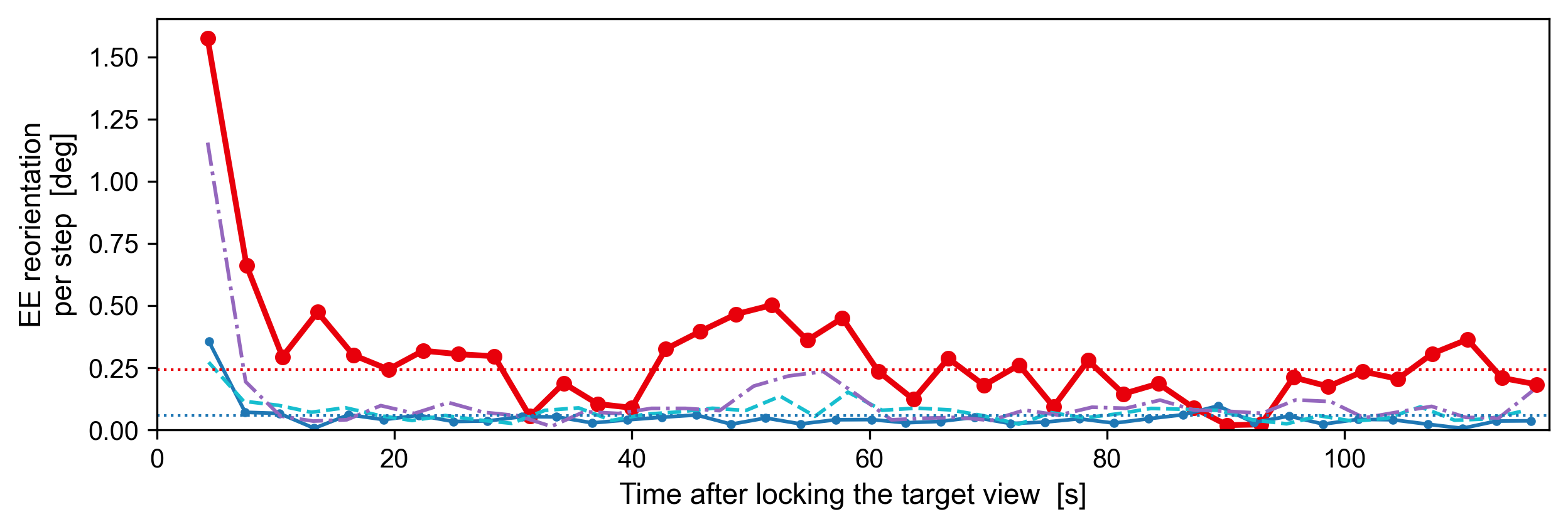}
        \subcaption{Per-step end-effector reorientation}
    \end{minipage}
    \caption{Representative closed-loop pose-hold traces, one run per marker type, in the order quARtet-D, quARtet-P, quARtet-rP, and Single. Over the common ${\sim}117$~s window, the plotted medians are $0.07$--$0.08\tcdegree$ for the three quARtet layouts and $0.47\tcdegree$ for Single in panel~(a), and $0.04$--$0.08\tcdegree$ and $0.24\tcdegree$, respectively, in panel~(b). Dotted horizontal lines mark the pooled quARtet (blue) and Single (red) medians.}
    \label{fig:closedloop_hold}
\end{figure}

\FloatBarrier
\textbf{Closed-loop pose-hold.}
Figure~\ref{fig:closedloop_hold} shows the closed-loop pose-hold traces of one representative run per marker type (the run whose mean hold error is closest to the median of its five runs) over the common ${\sim}117$~s window. All twenty runs are shown in Supplementary Section~S5.

The closed-loop columns of Table~\ref{tab:summary} summarize the five runs of each marker type by the mean and SD of the within-run means over the same window. The run means exceed the plotted medians because the error distributions are right-skewed. Single held the target with a mean orientation error of $0.68\tcdegree$ and reoriented by $0.31\tcdegree$ per step, whereas the three quARtet layouts held it within $0.08$--$0.11\tcdegree$ and reoriented by $0.05$--$0.10\tcdegree$ per step. With the run as the unit of replication, two-sided Welch $t$-tests against Single (Holm-corrected within each metric) confirm the separation for every layout: adjusted $p<10^{-4}$ for the orientation hold error and adjusted $p=0.012$ for the per-step reorientation. The run-to-run SDs ($0.01$--$0.02\tcdegree$ for the quARtet layouts and $0.06\tcdegree$ for Single in hold error) show that the separation is repeatable across re-placed and re-locked runs. The three quARtet layouts differ from one another by at most $0.05\tcdegree$ in either metric, and the test does not define a formal settling criterion.

\begin{table}[t]
    \centering
    \caption{Summary of the RQ1 results, with the Single control last. Fixed camera: orientation is the mean$\pm$SD over the 10 consecutive frames at the frontal cell of the broad grid, and position is the RMSE over the translation grid, both under the operational-reference and constant-alignment conventions of Section~\ref{sec:setup_fixed}. Closed-loop pose hold: mean$\pm$SD over five runs per marker type of the within-run mean orientation hold error and per-step end-effector reorientation, evaluated over the common ${\sim}117$~s window of Fig.~\ref{fig:closedloop_hold}. The fixed-camera entries summarize technical repeats within one setup and carry no inferential marks. In the closed-loop columns, asterisks mark quARtet entries that differ significantly from Single (two-sided Welch $t$-tests on the five run means, Holm-corrected within each metric, $^{*}$: adjusted $p<0.05$, $^{**}$: adjusted $p<0.01$). The smallest value in each column is bold.}
    \label{tab:summary}
    \setlength{\tabcolsep}{5pt}
    \renewcommand{\arraystretch}{1.2}
    \begin{tabular}{@{}l cc cc@{}}
    \toprule
    & \multicolumn{2}{c}{Fixed camera} & \multicolumn{2}{c}{Closed-loop pose hold} \\
    \cmidrule(lr){2-3}\cmidrule(l){4-5}
    Marker & \shortstack{Frontal orientation\\error [$\tcdegree$]} & \shortstack{Position\\RMSE [mm]} & \shortstack{Orientation hold\\error [$\tcdegree$]} & \shortstack{Per-step end-effector\\reorientation [$\tcdegree$]} \\
    \midrule
    quARtet-D  & $\mathbf{0.24\pm0.03}$ & $\mathbf{0.17}$ & $\mathbf{0.08\pm0.01}^{**}$ & $\mathbf{0.05\pm0.01}^{*}$ \\
    quARtet-P  & $0.27\pm0.03$ & $0.20$ & $0.09\pm0.02^{**}$ & $0.07\pm0.02^{*}$ \\
    quARtet-rP & $0.47\pm0.07$ & $0.19$ & $0.11\pm0.01^{**}$ & $0.10\pm0.01^{*}$ \\
    Single     & $2.18\pm0.80$ & $1.50$ & $0.68\pm0.06$ & $0.31\pm0.10$ \\
    \bottomrule
    \end{tabular}
\end{table}

\FloatBarrier
\subsection{Graspability Evaluation}\label{sec:rq2}
RQ2 is assessed experimentally through a physical swing-down test, while an idealized planar-grasp model provides complementary geometric insight into the observed differences among layouts (Supplementary Section~S6). The operational outcome is retention and in-grasp slip after the specified swing, where slip is the change of the object's pose relative to the gripper. Single, whose face is entirely flat, serves as the control. The evaluation is intended to compare the layouts under this specific grasping condition rather than to characterize general graspability.

\subsubsection{Settings}
\begin{figure}[t]
    \centering
    \setlength{\figcolgap}{0.02\linewidth}%
    \setlength{\figrowgap}{1.2ex}%
    \sbox{\aabcmeasbox}{\begin{minipage}[b]{2cm}\centering\rule{2cm}{2cm}\subcaption*{(x) Xg}\end{minipage}}%
    \setlength{\figcapht}{\dimexpr\ht\aabcmeasbox+\dp\aabcmeasbox-2cm\relax}%
    \setlength{\figtilew}{\dimexpr(\linewidth-2\figcolgap-\figcapht-\figrowgap-1pt)/4\relax}%
    \setlength{\figherow}{\dimexpr2\figtilew+\figcapht+\figrowgap\relax}%
    \begin{minipage}[b]{\figherow}\centering
        \includegraphics[width=\linewidth]{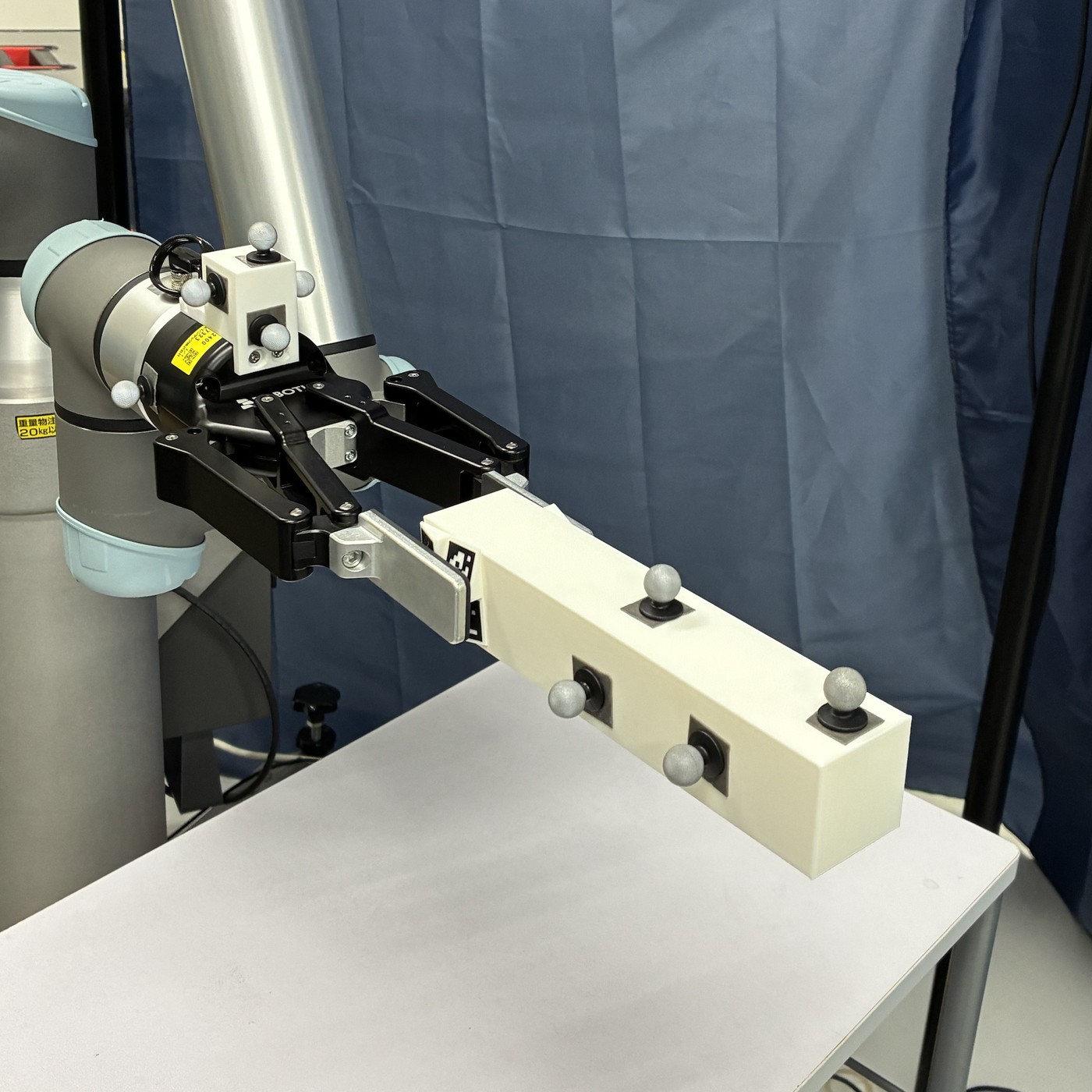}
        \subcaption{Grasped test object}
    \end{minipage}\hspace{\figcolgap}%
    \begin{minipage}[b]{\dimexpr2\figtilew+\figcolgap\relax}
        \setlength{\parskip}{0pt}\setlength{\lineskip}{0pt}\setlength{\lineskiplimit}{0pt}%
        \noindent
        \begin{minipage}[b]{\figtilew}\centering
            \includegraphics[width=\linewidth]{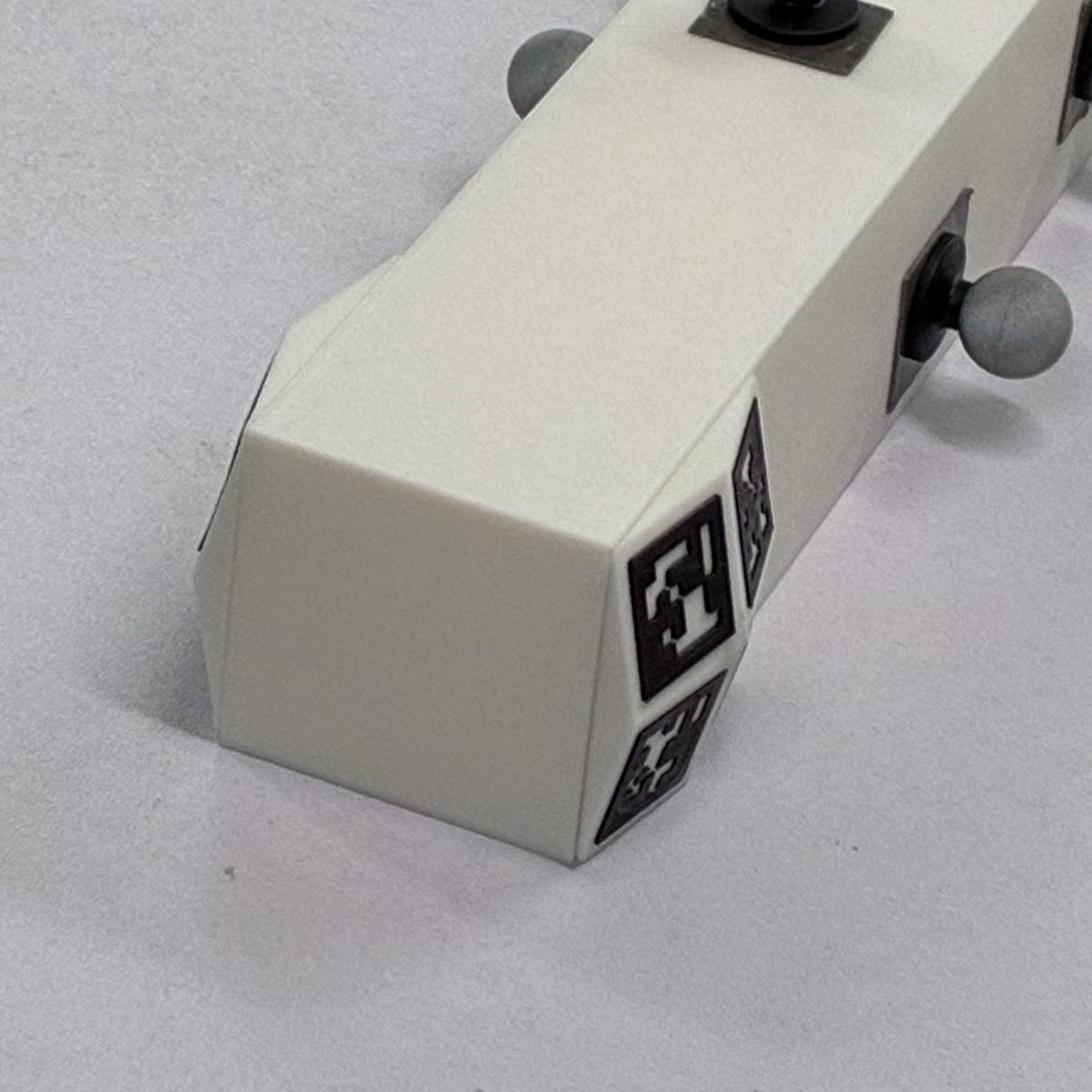}
            \subcaption{quARtet-D}
        \end{minipage}\hfill
        \begin{minipage}[b]{\figtilew}\centering
            \includegraphics[width=\linewidth]{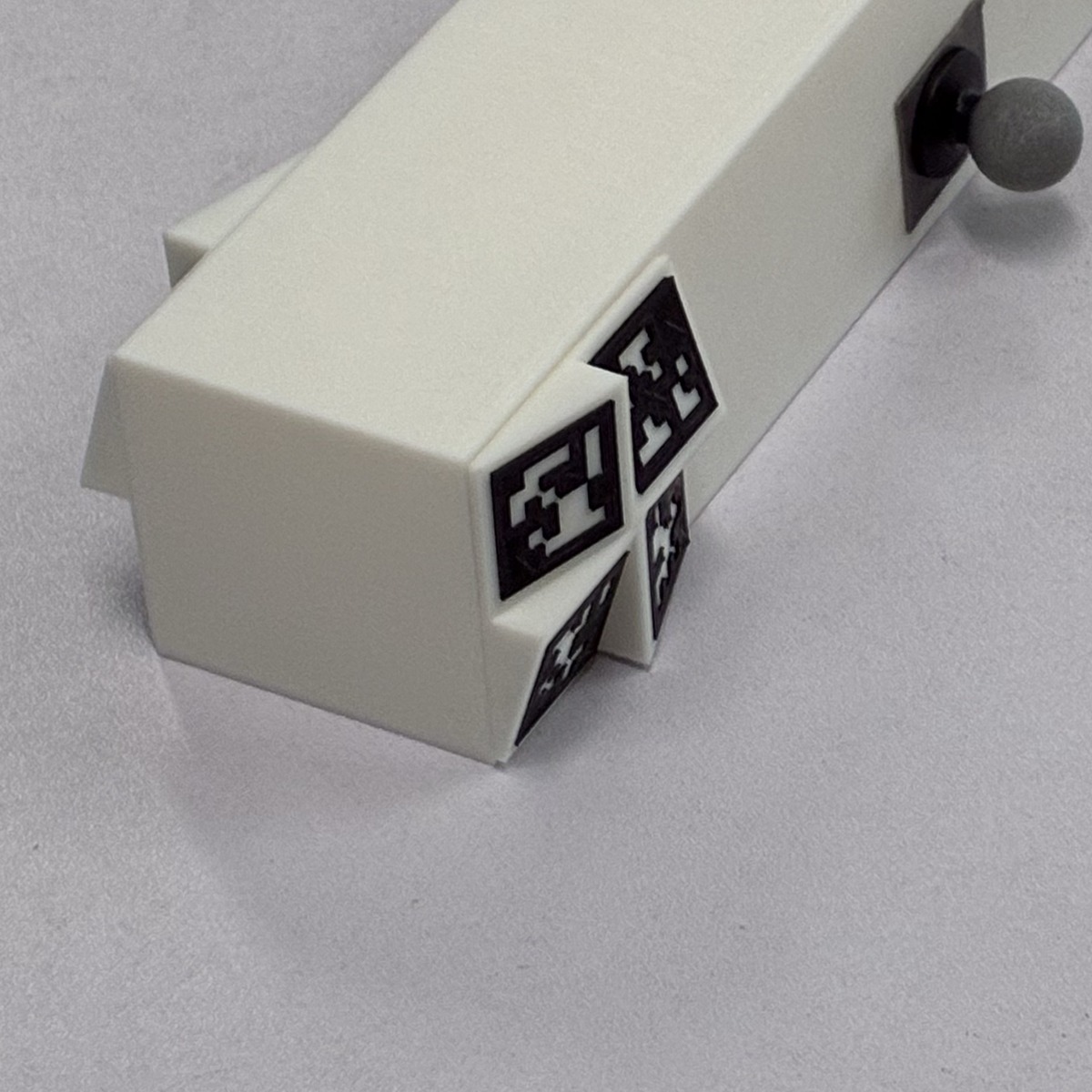}
            \subcaption{quARtet-P}
        \end{minipage}\par
        \vspace{\figrowgap}%
        \noindent
        \begin{minipage}[b]{\figtilew}\centering
            \includegraphics[width=\linewidth]{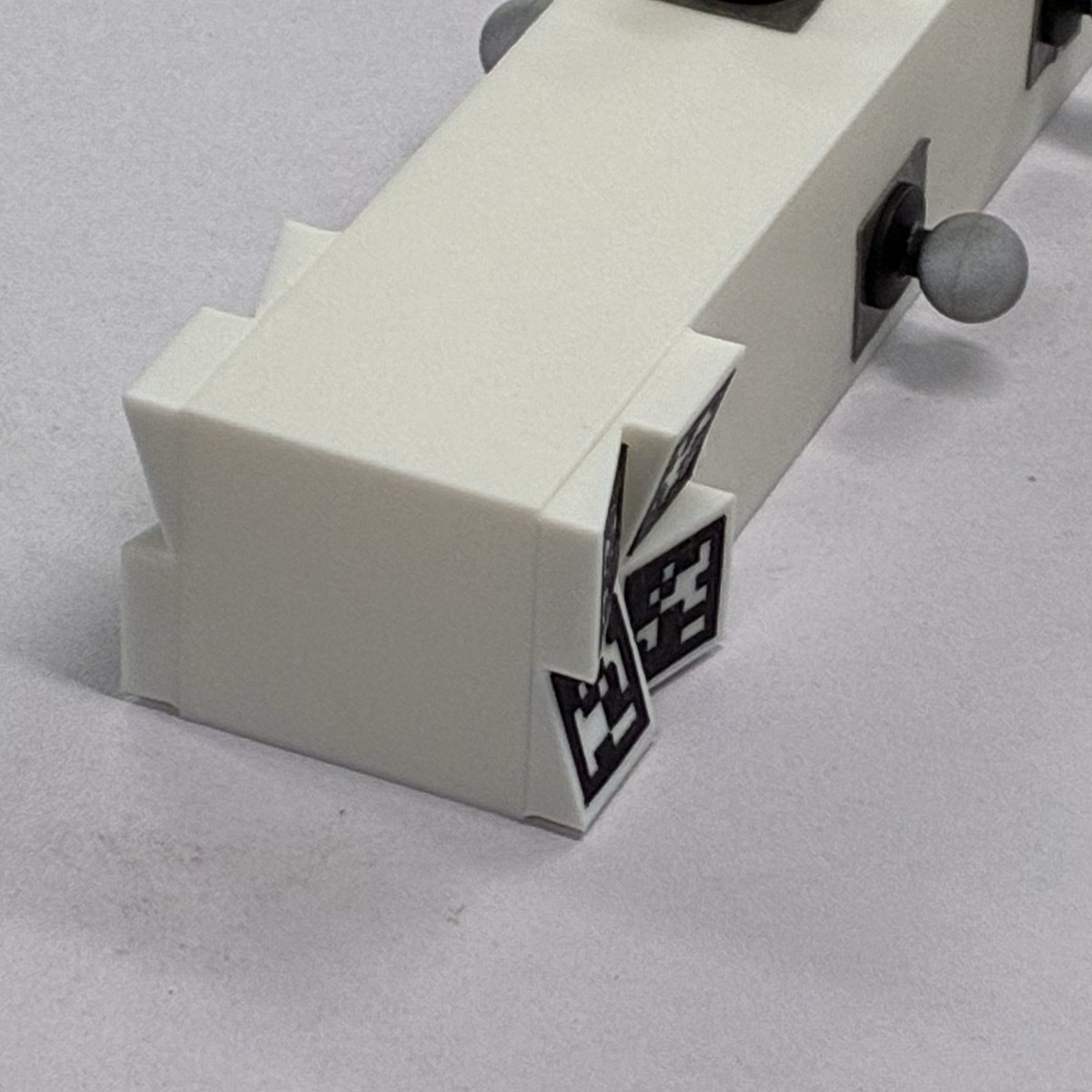}
            \subcaption{quARtet-rP}
        \end{minipage}\hfill
        \begin{minipage}[b]{\figtilew}\centering
            \includegraphics[width=\linewidth]{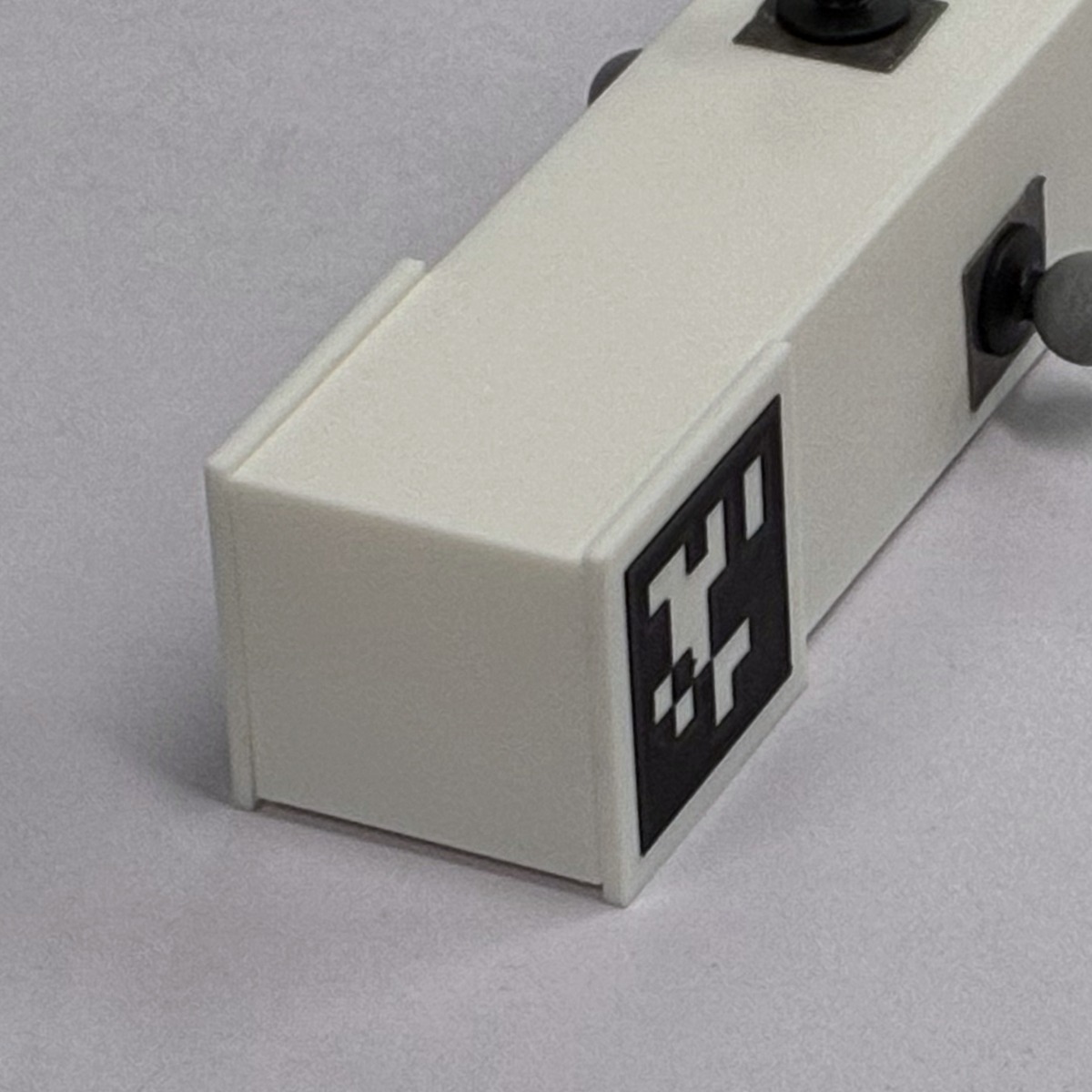}
            \subcaption{Single}
        \end{minipage}\par
    \end{minipage}
    \caption{Swing-down test object and marker contact surfaces. (a)~The $40\times40\times200$~mm object grasped near its marker end by the two-finger gripper. The flat fingers close directly on the marker faces, and the silver spheres on the gripper and object are the retroreflective spheres of the motion-capture system. (b)--(e)~Close-ups of the markers affixed to the grasped end, in the order quARtet-D, quARtet-P, quARtet-rP, and Single (control).}
    \label{fig:swingdown_setup}
\end{figure}

\begin{figure}[t]
    \centering
    \begin{tikzpicture}[>={Latex[length=2.6mm]}]
      \node[inner sep=0] (fa) at (0,0) {\includegraphics[width=0.29\linewidth]{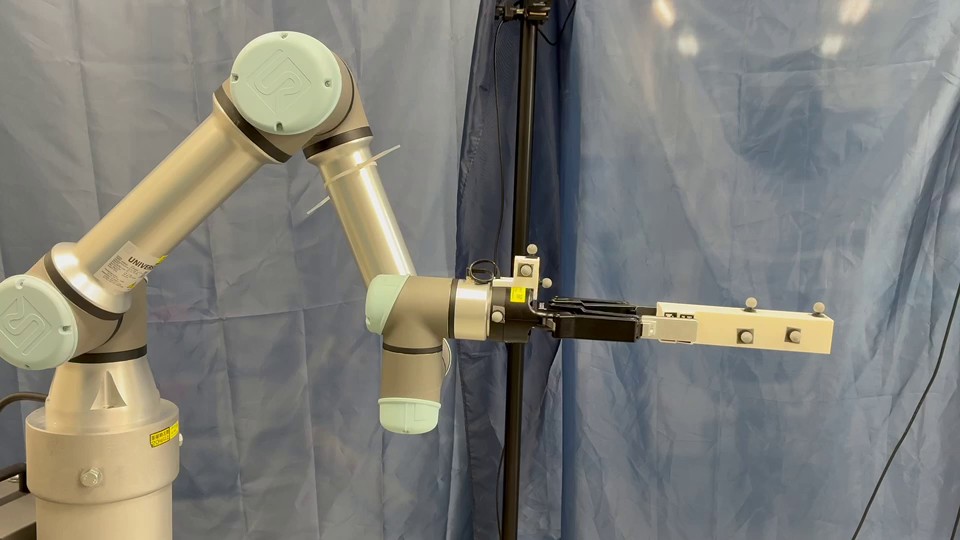}};
      \node[inner sep=0, right=0.045\linewidth of fa] (fb) {\includegraphics[width=0.29\linewidth]{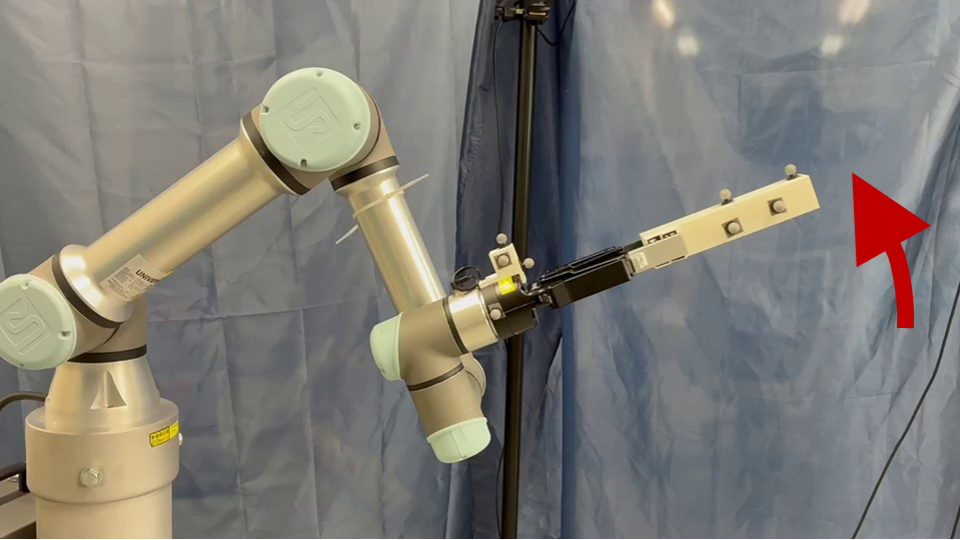}};
      \node[inner sep=0, right=0.045\linewidth of fb] (fc) {\includegraphics[width=0.29\linewidth]{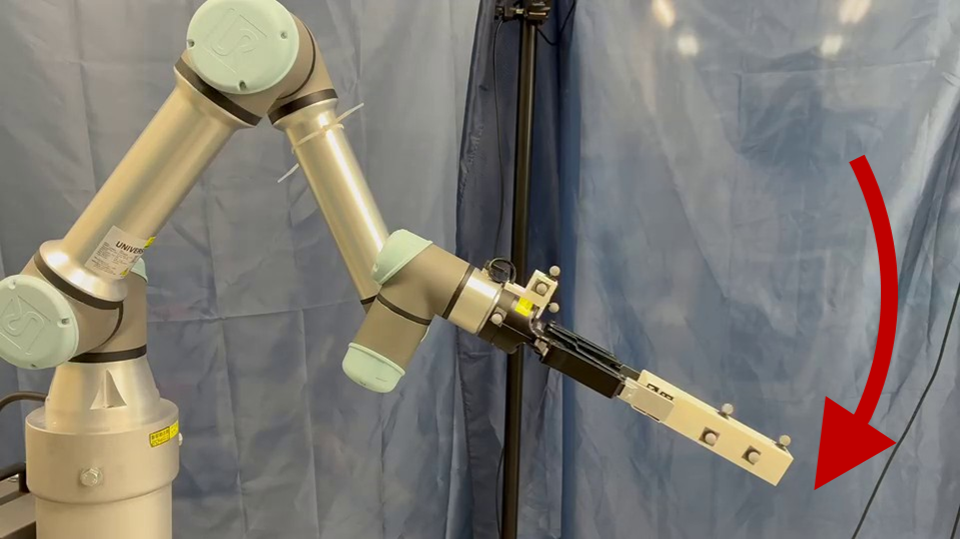}};
      \draw[->, very thick] (fa) -- (fb);
      \draw[->, very thick] (fb) -- (fc);
      \node[below=1pt of fa, font=\small] {Grasped};
      \node[below=1pt of fb, font=\small] {Raised to $+25\tcdegree$};
      \node[below=1pt of fc, font=\small] {Swung down to $-25\tcdegree$};
    \end{tikzpicture}
    \caption{One-way swing-down motion (quARtet-P shown). The red arcs indicate the commanded rotation of the gripper about the swing axis: raising to $+25\tcdegree$, then swinging down to $-25\tcdegree$.}
    \label{fig:swingdown_motion}
\end{figure}

The test object was a $40\times40\times200$~mm PLA rectangular prism ($92$~g, or about $102$~g with the two markers and double-sided tape attached). A marker was affixed with thin double-sided tape to each of the two opposing $40\times200$~mm faces at one end (Fig.~\ref{fig:swingdown_setup}). A Robotiq 2F-140 parallel-jaw gripper mounted on the UR5e grasped that end so that its flat fingers closed directly on the two marker faces, at a $10\%$ force setting on the gripper's control interface (the manufacturer specifies a $10$--$125$~N grip-force range for this gripper, and the force at each marker's contact surfaces was not measured in situ). The gripper was centered on the marker with the distal edge of the fingertips aligned to the marker edge farthest from the palm, and the same grasp pose was used for every marker type. Keeping the end-effector position fixed, the robot first raised the object to $+25\tcdegree$ and then swung it down to $-25\tcdegree$ about a horizontal axis through the gripper, hereafter the swing axis (Fig.~\ref{fig:swingdown_motion}). The command duration was 0.1~s, but the arm's speed and acceleration limits stretched this into a realized swing that motion capture measured at $0.63$--$0.65$~s ($5$--$95\%$ of amplitude) with a sustained peak angular speed of about $180\tcdegree$/s about the swing axis (brief frame-level peaks reached ${\sim}230\tcdegree$/s for at most $17$~ms). The realized motion was highly repeatable: the swing amplitude was $50.0\tcdegree$ with a trial-to-trial SD below $0.02\tcdegree$ for every marker type, so all four marker types experienced the same commanded motion and inertial load (Supplementary Section~S7). The object was re-grasped before each of five trials per marker type, so the trials are independent replicates.

An OptiTrack system (six Flex~13 cameras, Motive~2.3.0, 120~Hz) tracked rigid bodies defined by six retroreflective spheres on the gripper and six on the object. These motion-capture spheres are distinct from the printed fiducial markers under test. In-grasp slip was the change in the object's pose relative to the gripper between the static holds before and after the swing. Position slip is the Euclidean relative translation, and orientation slip is the geodesic relative rotation angle. Supplementary Section~S7 gives the frame-level calculation and all five trial traces.

\subsubsection{Results}

\begin{figure}[tp]
    \centering
    \begin{minipage}[b]{0.40\linewidth}\centering
        \includegraphics[width=\linewidth]{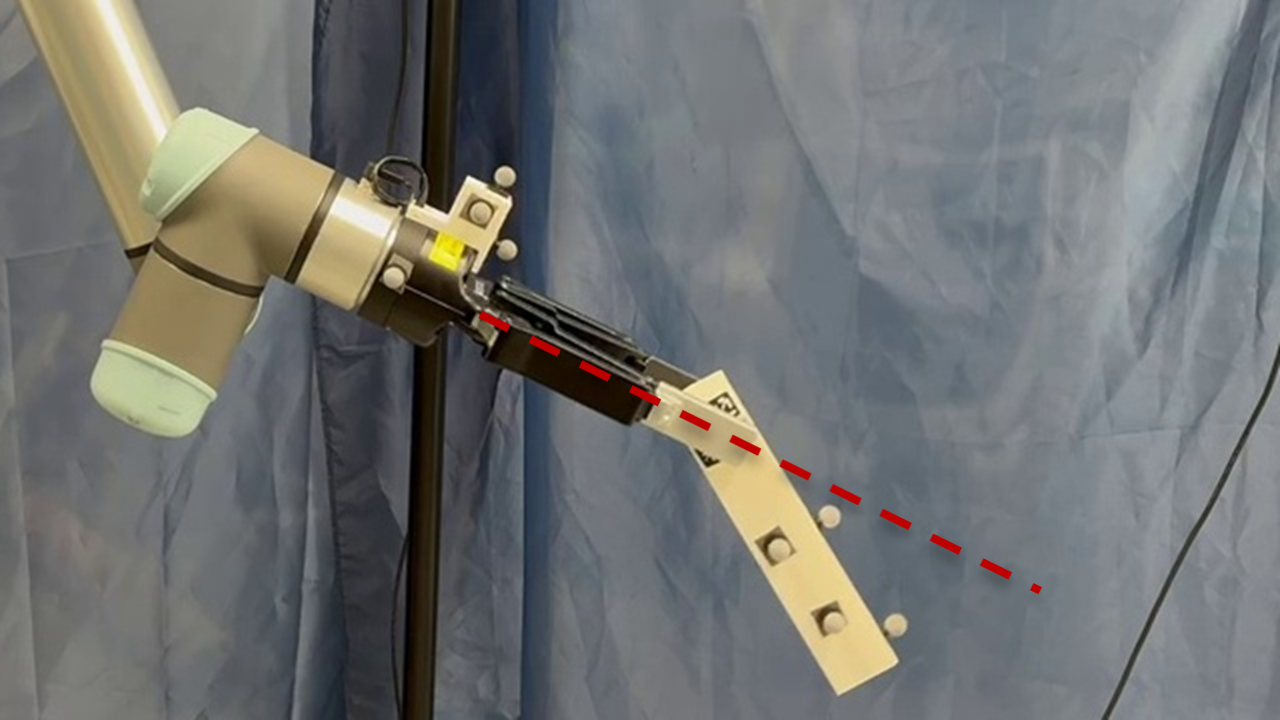}
        \subcaption{quARtet-D: post-swing snapshot}
    \end{minipage}\hfill
    \begin{minipage}[b]{0.40\linewidth}\centering
        \includegraphics[width=\linewidth]{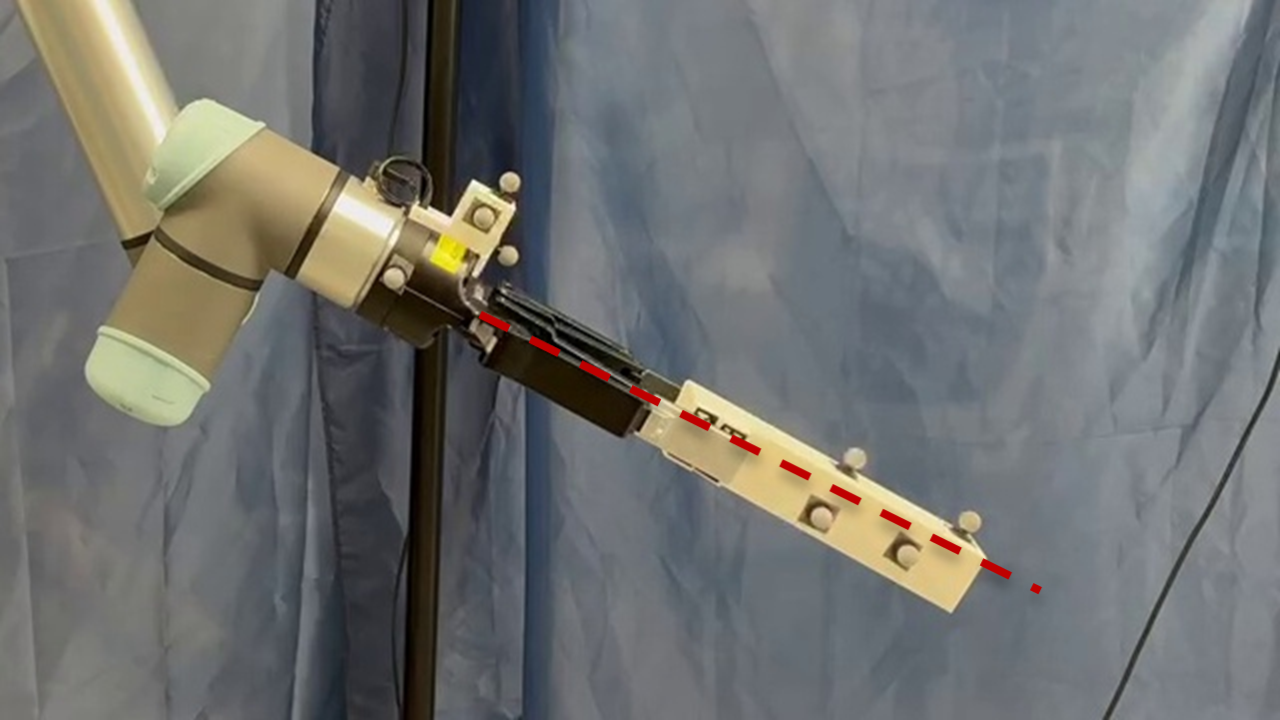}
        \subcaption{quARtet-P: post-swing snapshot}
    \end{minipage}
    \vspace{0.5ex}
    \begin{minipage}[b]{0.48\linewidth}\centering
        \includegraphics[width=.92\linewidth]{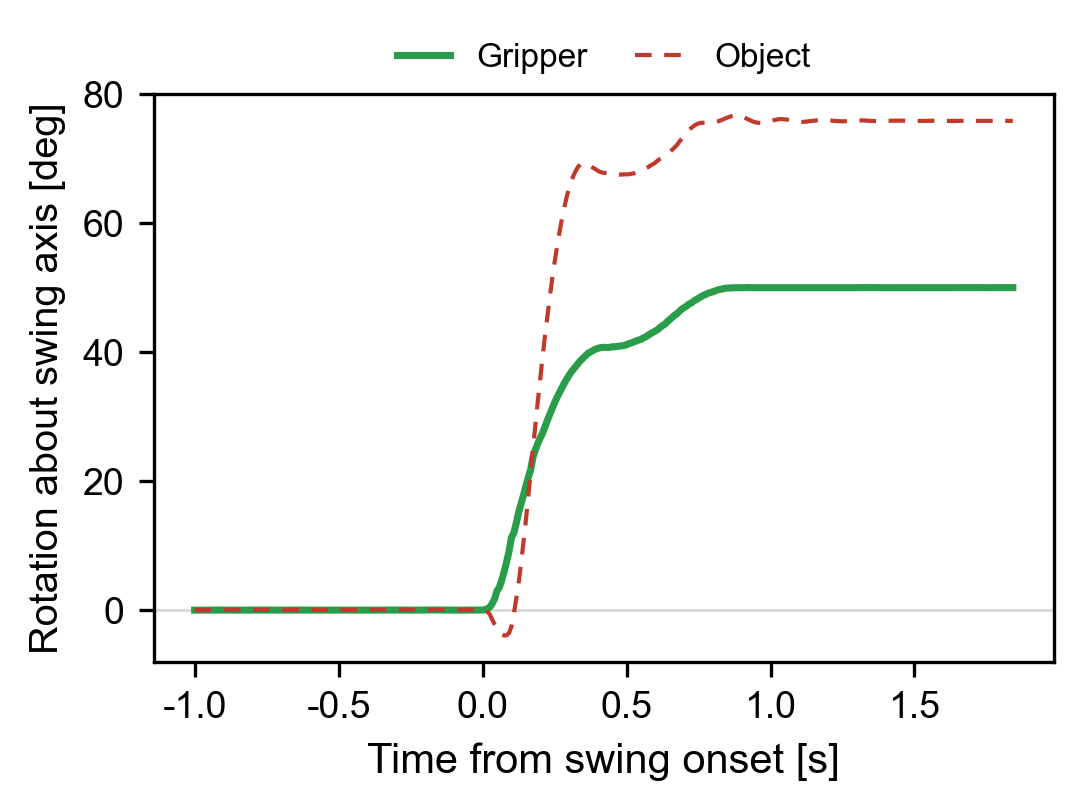}
        \subcaption{quARtet-D}
    \end{minipage}\hfill
    \begin{minipage}[b]{0.48\linewidth}\centering
        \includegraphics[width=.92\linewidth]{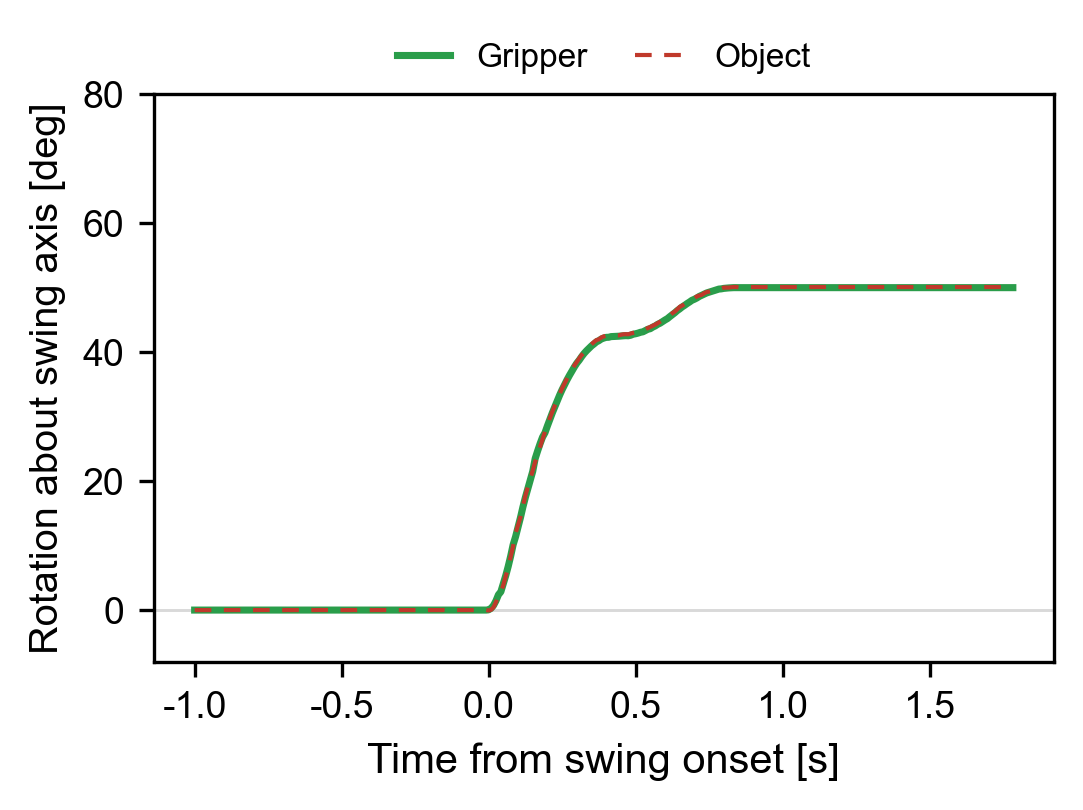}
        \subcaption{quARtet-P}
    \end{minipage}
    \vspace{0.5ex}
    \begin{minipage}[b]{0.48\linewidth}\centering
        \includegraphics[width=.92\linewidth]{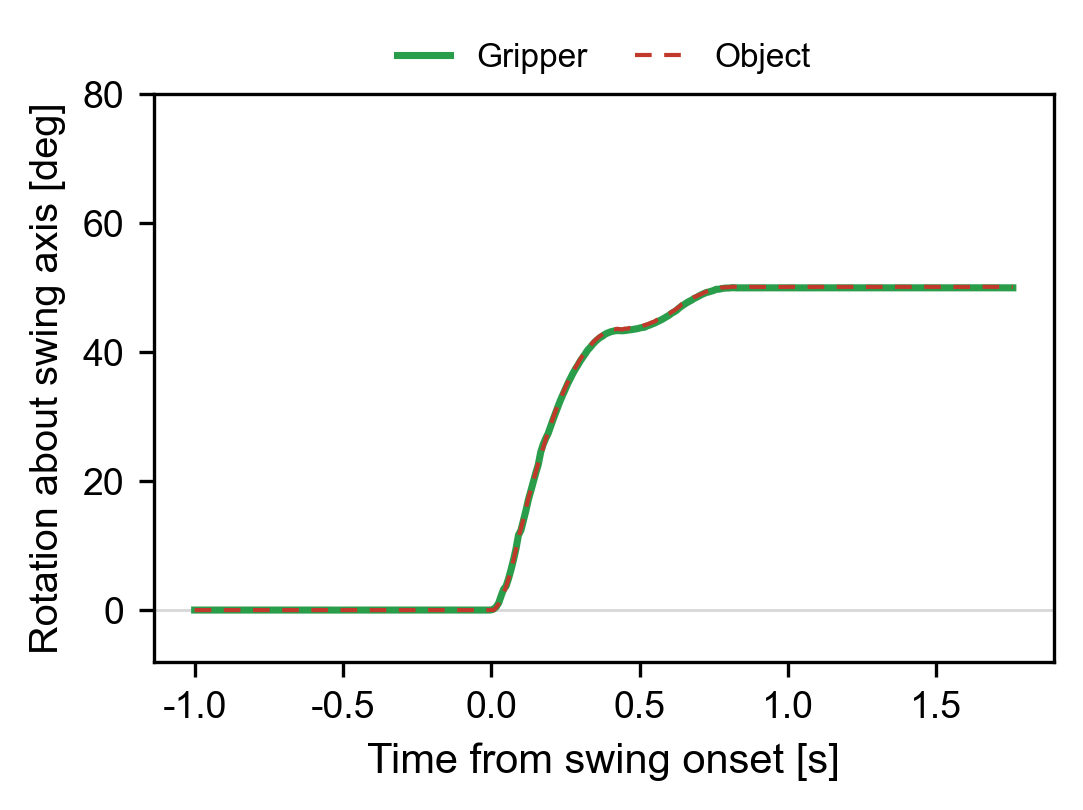}
        \subcaption{quARtet-rP}
    \end{minipage}\hfill
    \begin{minipage}[b]{0.48\linewidth}\centering
        \includegraphics[width=.92\linewidth]{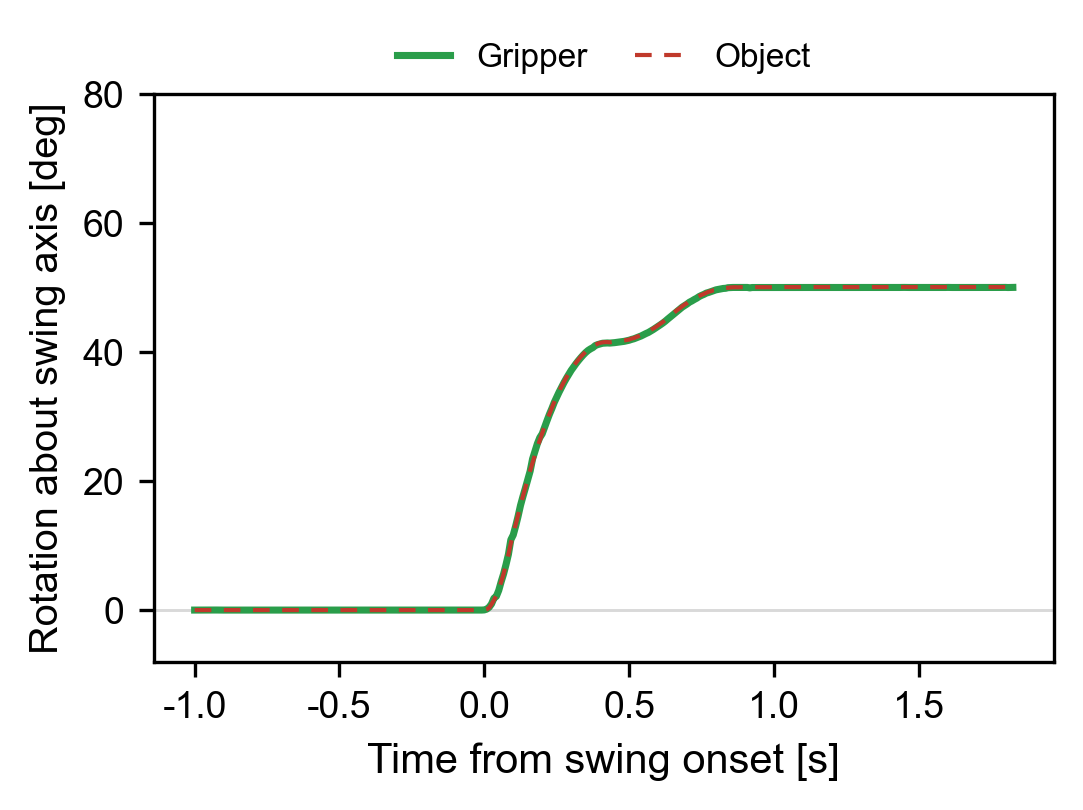}
        \subcaption{Single}
    \end{minipage}
    \caption{Swing-down outcome and representative time series. (a),(b)~Snapshots at the post-swing hold for quARtet-D and quARtet-P. The dashed line extends the gripper's finger axis, along which the object would lie if the grasp had remained rigid: the quARtet-P object stays on it, whereas the quARtet-D object has pivoted away and slid outward (quARtet-rP and Single behaved like quARtet-P, Table~\ref{tab:swingdown}). (c)--(f)~Gripper (solid green) and object (dashed red) rotation about the swing axis relative to the pre-swing hold, for the trial closest to the median position slip of each marker type, with the downward swing plotted as a positive rotation. The dashed style keeps both curves visible where they coincide. Time is aligned at swing onset, each window spans the two bracketing static holds, and all panels share one scale. All five trials per marker type are shown in Supplementary Section~S7.}
    \label{fig:swingdown_outcome}
\end{figure}

The post-swing snapshots in Fig.~\ref{fig:swingdown_outcome}(a) and (b) show the outcome directly: the quARtet-P object still lies on the extension of the gripper's finger axis, whereas the quARtet-D object has pivoted away from it and slid outward within the fingers. The representative time series in Fig.~\ref{fig:swingdown_outcome}\mbox{(c)--(f)} show how this arises: coincident gripper and object rotations indicate a stable grasp, and persistent separation indicates slip. The quARtet-D object separates from the gripper during the swing and stays separated through the post-swing hold, whereas quARtet-P, quARtet-rP, and Single remain closely aligned. The representative trial is the trial closest to the median position slip of each marker type.

\FloatBarrier
\begin{table}[!htb]
    \centering
    \setlength{\tabcolsep}{6pt}
    \renewcommand{\arraystretch}{1.2}
    \caption{In-grasp slip after the one-way swing-down (mean$\pm$SD, $N=5$ independently re-grasped trials per marker type). Smaller is better, the best value in each column is bold, and Single is last as the control. Asterisks mark quARtet entries that differ significantly from Single (two-sided Welch $t$-tests, Holm-corrected within each metric, $^{**}$: adjusted $p<0.01$). All starred entries have adjusted $p<0.001$ (Supplementary Section~S7).}
    \label{tab:swingdown}
    \begin{tabular}{@{}l c c@{}}
    \toprule
    Marker & Position slip [mm] & Orientation slip [$\tcdegree$] \\
    \midrule
    quARtet-D  & $51.76\pm7.48^{**}$ & $24.85\pm3.69^{**}$ \\
    quARtet-P  & $0.86\pm0.06^{**}$ & $0.31\pm0.02$ \\
    quARtet-rP & $1.01\pm0.07^{**}$ & $0.32\pm0.05$ \\
    Single     & $\mathbf{0.51\pm0.02}$ & $\mathbf{0.30\pm0.03}$ \\
    \bottomrule
    \end{tabular}
\end{table}

Across five trials, quARtet-D slipped by $51.76\pm7.48$~mm and $24.85\pm3.69\tcdegree$. quARtet-P, quARtet-rP, and Single showed $0.51$--$1.01$~mm and approximately $0.3\tcdegree$ of slip (Table~\ref{tab:swingdown}). Welch $t$-tests on the per-trial slips against Single (Holm-corrected within each metric) confirm the visual impression: quARtet-D slips significantly more in both position and orientation (adjusted $p<0.001$). quARtet-P and quARtet-rP also differ significantly from Single in position slip (adjusted $p<0.001$), but the difference is small, $+0.35$ and $+0.50$~mm on average ($95\%$ confidence intervals $[0.27,0.42]$ and $[0.41,0.59]$~mm), and their orientation slip is not detectably different from Single's (adjusted $p=0.77$). Marking part of the face with a pitch-based quARtet therefore costs about half a millimeter of additional slip under this load, whereas quARtet-D costs about a hundred times more. In no trial did the object leave the gripper entirely: with quARtet-D it pivoted by ${\sim}25\tcdegree$ and shifted by ${\sim}50$~mm relative to the gripper, which we classify as loss of stable retention. All five time-series overlays, per-trial slip traces, and the per-trial dot plot are given in Supplementary Section~S7.

The supplementary idealized model gives the same coarse explanation: quARtet-D has no effective flat grasp region, whereas quARtet-P and quARtet-rP retain accessible strips (Supplementary Section~S6). It is not used to rank the small difference between the two pitch-based layouts because it omits friction, compliance, force closure, and full-gripper collision. The model's area-ratio metric favors quARtet-rP, whereas the physical test slightly favors quARtet-P, plausibly because the tested grasp pose engaged mainly the central region of the face, where the pitch layout retains its accessible inner-edge strips (Supplementary Sections~S6 and~S7).

In direct answer to RQ2 under the specified gripper, force, object, and swing, quARtet-P and quARtet-rP retained the object in all five trials with small slip, whereas quARtet-D did not provide stable retention. The result establishes a design trade-off rather than a universally best layout: quARtet-D favors pose-estimation consistency, and the pitch-based layouts preserve grasp access.

\section{Discussion}\label{sec:discussion}
\textbf{Main findings.} The evaluation supports two distinct conclusions. For RQ1, all three quARtet layouts showed far smaller near-frontal orientation and position errors than Single within the robot-referenced, same-setup protocol. quARtet-D was the most consistent over the tested viewpoint grid, while the pitch-based layouts traded away some consistency at oblique views where a planar tag already has small errors (Supplementary Section~S2.3). In closed-loop use, five runs per marker type showed the same separation with statistical support: every quARtet layout held the target with a significantly smaller orientation error and per-step reorientation than Single. For RQ2, quARtet-P and quARtet-rP retained the object during all five swing-down trials with about half a millimeter of additional slip relative to Single, whereas quARtet-D slipped about a hundred times more than Single. These results should therefore be read as a trade-off, not as evidence for one universally superior layout.

\textbf{Design trade-off and implications.} The physical mechanism makes the layout choice interpretable. Tilting the four tag faces in different directions supplies non-coplanar perspective information near a frontal view. quARtet-D distributes those directions most symmetrically and with the least mutual occlusion (Section~\ref{sec:detection}), but leaves no effective flat contact strip. The pitch-based layouts sacrifice some off-axis consistency to preserve regions that a planar finger can contact. For the tested marker size and gripper, quARtet-D is the appropriate choice when the marked face is not grasped, and quARtet-P or quARtet-rP is preferable when that face must remain a contact surface. Relative to the specialized optical markers and learned detection pipelines of Section~\ref{sec:related}, near-frontal stabilization and grasp access are obtained here with nothing beyond an FDM-printable geometry and a standard detector.

\textbf{Limitations.} Several limitations bound the claims. First, the study evaluates marker-level pose errors and grasp-level slip rather than success in an end-to-end laboratory task such as insertion or pick-and-place. Second, the fixed-camera analysis uses robot-commanded relative pose changes as an operational reference (Section~\ref{sec:setup_fixed}), not an independently calibrated external ground truth. The $\pm0.03$~mm manufacturer figure is translational repeatability, is comparable to the smallest within-cell variations reported, and does not quantify orientation-reference uncertainty. Third, the constant frame alignment is fitted on the same dataset it evaluates. It removes global frame mismatch together with any constant marker-specific bias, so the results support comparisons of pose-dependent errors after alignment but not absolute pose bias or held-out calibration accuracy. Fourth, the ten images of each static condition are consecutive frames at a single mounting and quantify within-acquisition variation. Independent re-mounted or re-acquired runs would be needed for setup-level inference. Fifth, the five closed-loop runs per marker type were recorded in one session with the camera fixture unchanged, so their statistics address repeatability across manual re-placements of the marker and re-locked targets, not variability across sessions or camera re-mounting. Sixth, each experiment used a single configuration of working distance (about $220$~mm for the fixed camera and about $200$~mm for the robot-mounted camera), marker size ($35$~mm, whose $16$~mm sub-tags decode at shorter range than the full-size tag of Single), illumination, camera, test object, and tape attachment (shared by all marker types), and the detection and joint-PnP runtime was not evaluated. Seventh, fabrication fidelity, optical contrast, reflections, and contamination may affect detection. Finally, multiple simultaneous quARtet instances and progressive loss of visible tags are supported in principle by ID-based matching but were not evaluated experimentally. These caveats bound absolute-accuracy and setup-level claims. The layout comparison itself was made within one shared setup, so reference and setup uncertainties are expected to be largely common to all marker types, and the observed separations are large: roughly five- to ninefold in the near-frontal errors and about a hundredfold in grasp retention.

\textbf{Overall implication.} Within these limits, the contribution is a configurable marker family whose geometry makes the consistency--graspability choice explicit: the evaluation delineates where the tested layouts reduce errors relative to a planar tag under the common protocol, and what physical access is lost in return.

\section{Conclusions}\label{sec:conclusion}
This paper presented quARtet, a 3D-printable marker that combines four tilted AprilTags in one compact footprint and estimates their common pose in a joint PnP solve. Under the tested fixed-camera conditions and robot-referenced protocol, all three quARtet layouts showed much smaller near-frontal orientation and position errors than Single, and a five-run closed-loop pose-hold test confirmed that separation under visual feedback. The swing-down experiment showed that the benefit must be balanced against grasp access: quARtet-D was the most consistent across viewpoints but slipped about a hundred times more than Single when grasped, whereas quARtet-P and quARtet-rP preserved flat contact regions and held the object in all five trials at a cost of a small additional slip.

The resulting layout rule for the tested marker size and gripper is to select quARtet-D when pose-estimation consistency is primary and the marked face is not a grasp surface, and to select a pitch-based layout when the gripper must contact that face. Future work should evaluate the selected layout in end-to-end laboratory manipulation and under multiple-marker and partial-occlusion conditions.

\backmatter

\bmhead{Supplementary Information}

The Supplementary Information (Sections~S1--S7, referenced throughout the text) follows the references.

\bmhead{Data and Code Availability}

The Fusion~360 add-in, the shared configuration file, the pose-estimation code, the STL files of the evaluated markers, the per-measurement comma-separated values (CSV) datasets of the fixed-camera and closed-loop experiments, and the per-trial slip values of the swing-down experiment are available at \url{https://github.com/Wa-Araki/quARtet-marker}. Supplementary Section~S3 documents the names of the released datasets, the alignment conventions, and the analysis steps by which the fixed-camera and closed-loop entries of Table~\ref{tab:summary} and the cell-wise summaries of Supplementary Section~S2.3 are obtained from them.

\bmhead{Acknowledgements}

The authors gratefully acknowledge the Materials Informatics Laboratory, the Division of Materials Science, Nara Institute of Science and Technology, for kindly lending the UR5e six-axis robotic arm and the Raise3D E2 3D printer used in this work.

\bmhead{Declaration of AI Use}

The authors used ChatGPT (OpenAI) and Claude (Anthropic) during the preparation of this manuscript. The concept, the marker design, the experiments, and the data are the authors' own. The tools were used to edit and rephrase the authors' text, to draft passages from the authors' notes, data, and instructions, to suggest alternative phrasing, framing, and structure, and to check the consistency of numbers and terminology and whether each claim was supported by the data. They were also used to assist in writing the scripts that compute the summary statistics, generate the plots, and format the released data files, and to prepare the released code for publication (a configuration-file wrapper for the detector and English code comments). The authors decided which suggestions to adopt, revised the text themselves, and verified all AI-assisted text, code, and outputs. All scientific judgments, including the interpretation of the results and the conclusions, were made by the authors, who take full responsibility for the content of the article.

\bmhead{Author Contributions}

The individual author contributions, following the CRediT taxonomy, are as follows.
\begin{itemize}
\item \textbf{Araki Wakiuchi}: Conceptualization, Methodology, Software, Investigation, Formal analysis, Visualization, Writing -- original draft.
\item \textbf{Hikaru Sasaki}: Methodology, Supervision, Writing -- review \& editing.
\item \textbf{Takamitsu Matsubara}: Methodology, Supervision, Writing -- review \& editing.
\end{itemize}
The graspability evaluation arose from discussions among all authors. All authors discussed the results and approved the final manuscript.

\bmhead{Funding}

The authors declare that no funds, grants, or other support were received during the preparation of this manuscript.

\bmhead{Competing Interests}

The authors declare no competing interests.

\bmhead{Ethics Approval and Consent}

Not applicable. This study involved no human participants or animals.

\bibliography{reference}%

\clearpage
\phantomsection
\pdfbookmark[1]{Supplementary Information}{si}
\graphicspath{{si_figures/}}
\setcounter{figure}{0}\setcounter{table}{0}\setcounter{equation}{0}
\renewcommand{\thefigure}{S\arabic{figure}}\renewcommand{\thetable}{S\arabic{table}}\renewcommand{\theequation}{S\arabic{equation}}
\setlength{\parindent}{0pt}\setlength{\parskip}{0.5em}

\begin{center}
{\Large\bfseries Supplementary Information}\\[1em]
{\Large quARtet Marker: A 3D-Printable Multi-Tag Fiducial for Robust Near-Frontal Pose Estimation}\\[1em]
Araki Wakiuchi$^{1,*}$, Hikaru Sasaki$^{2}$, Takamitsu Matsubara$^{2}$\\[1em]
$^{1}$Materials Informatics Initiative, RD Technology and Digital Transformation Center, JSR Corporation, 3-103-9 Tonomachi, Kawasaki-ku, Kawasaki, Kanagawa, 210-0821, Japan.\\
$^{2}$Division of Information Science, Graduate School of Science and Technology, Nara Institute of Science and Technology, 8916-5 Takayama-cho, Ikoma, Nara, 630-0192, Japan.\\
$^{*}$Corresponding author: araki\_wakiuchi@jsrgr.com
\end{center}

\section*{Notation and abbreviations}
This list makes the Supplementary Information readable on its own. The full definitions are given in the main text at the points indicated.

\textbf{Marker types.} Four marker types are compared throughout, in the order used in every result display. The quARtet abbreviations are introduced in main-text Section~3.1 and used from main-text Fig.~2 onward, and the four types, including the control, are listed in main-text Section~4.1:
\begin{itemize}\setlength{\itemsep}{0pt}
\item \textbf{quARtet-D}: the quARtet marker in the \emph{diagonal} layout.
\item \textbf{quARtet-P}: the quARtet marker in the \emph{pitch} layout.
\item \textbf{quARtet-rP}: the quARtet marker in the \emph{reversed-pitch} layout.
\item \textbf{Single}: the control, one planar AprilTag on the same footprint.
\end{itemize}
The three layout geometries are described in main-text Section~3.2 (Fig.~3, with the configurable parameters in Table~1) and specified numerically in Section~S3.4. \emph{Pitch-based layouts} denotes quARtet-P and quARtet-rP together. \emph{Marker type} refers to all four, and \emph{quARtet layout} refers only to the three quARtet variants. RQ1 and RQ2 are the two research questions stated in main-text Section~4.

\textbf{On the word ``pitch''.} In Section~S1 only, \emph{roll} and \emph{pitch} denote the two rotation angles of the simulated planar marker, and they are unrelated to the pitch layout.

\textbf{Abbreviations.} SD, standard deviation (sample, $\mathrm{ddof}=1$); RMSE, root-mean-square error; CI, confidence interval; PnP, Perspective-n-Point; CSV, comma-separated values; $\mathrm{SO}(3)$ and $\mathrm{SE}(3)$, the rotation and rigid-motion groups. Grid symbols: $\theta_x$, $\theta_y$, and $\theta_z$ are the commanded rotations about the $x$-, $y$-, and $z$-axes of the initial pose ($\theta_x$ and $\theta_y$ are called pitch and roll in main-text Section~4.2.1, and the axis labels of Figs.~\ref{fig:S8} and~\ref{fig:S9} give the same angles as $\Delta\theta_x$ and $\Delta\theta_y$), and $\Delta x$, $\Delta y$, and $\Delta z$ are the commanded translations along the same axes (mm).

\section*{S1. Simulation-based rationale for tilt-angle selection}
This section provides a simplified simulation used to support the selection of the base tilt described in Section~3.2 of the main text. The purpose is not to reproduce the full quARtet geometry, but to examine how pose-estimation accuracy of a square planar marker changes with viewing angle under idealized imaging conditions and pixel-level quantization perturbations.

\subsection*{S1.1. Purpose and simulation conditions}
The simulation evaluates pose-estimation sensitivity for a square marker tilted in roll and pitch. A pinhole camera model is assumed, and projected image points are quantized to integer pixel coordinates before pose estimation. Conditions were: image width $W=800$ pixels, image height $H=600$ pixels, focal length $f=800$ pixels, marker side length $s=80$ and camera-to-marker distance $d=800$ (both in arbitrary length units, since only their ratio matters), roll range $0\tcdegree$ to $72\tcdegree$ in $3\tcdegree$ increments, and pitch range $0\tcdegree$ to $72\tcdegree$ in $3\tcdegree$ increments. All angles are reported in degrees unless otherwise stated.

\subsection*{S1.2. Camera model and square-marker geometry}
The camera intrinsic matrix is
\begin{equation}
K=\begin{bmatrix} f&0&c_x\\0&f&c_y\\0&0&1\end{bmatrix},\qquad c_x=\frac{W}{2},\quad c_y=\frac{H}{2}.
\end{equation}
The square marker is centered at the origin on the plane $z=0$, and its four corners are
\begin{equation}
\begin{aligned}
\bm X_0&=[-s/2,\ s/2,\ 0]^\mathsf{T},&\bm X_1&=[s/2,\ s/2,\ 0]^\mathsf{T},\\
\bm X_2&=[s/2,\ -s/2,\ 0]^\mathsf{T},&\bm X_3&=[-s/2,\ -s/2,\ 0]^\mathsf{T}.
\end{aligned}
\end{equation}

\subsection*{S1.3. Projection under roll--pitch tilts}
For a roll--pitch pair $(\phi,\theta)$, the ground-truth rotation is
\begin{equation}R_{\rm true}=R_y(\theta)R_x(\phi),\end{equation}
where $R_x$ and $R_y$ are standard right-handed rotations. Yaw is fixed to zero. The ground-truth translation is
\begin{equation}\bm t_{\rm true}=[0,0,d]^\mathsf{T}.\end{equation}
Each corner is transformed as
\begin{equation}\bm X_{{\rm cam},i}=R_{\rm true}\bm X_i+\bm t_{\rm true}=[x_i,y_i,z_i]^\mathsf{T}.\end{equation}
Normalized image coordinates and pixel coordinates are
\begin{equation}\tilde u_i=x_i/z_i,\quad \tilde v_i=y_i/z_i,\end{equation}
\begin{equation}[u_i,v_i,1]^\mathsf{T}=K[\tilde u_i,\tilde v_i,1]^\mathsf{T}.\end{equation}
To mimic pixel quantization, projected coordinates are rounded:
\begin{equation}\hat u_i={\rm round}(u_i),\quad \hat v_i={\rm round}(v_i).\end{equation}

\subsection*{S1.4. Pixel-perturbation protocol}
For each $(\phi,\theta)$ pair, 16 perturbed cases were generated by perturbing exactly one of the four corners by one of the four axis-aligned one-pixel offsets $(+1,0)$, $(-1,0)$, $(0,+1)$, and $(0,-1)$, while the remaining corners were left unchanged:
\begin{equation}4\ {\rm corners}\times4\ {\rm offsets}=16\ {\rm perturbed\ cases}.\end{equation}

\subsection*{S1.5. Pose estimation and rotation error metric}
Pose estimation was performed using OpenCV's \texttt{solvePnP} with \texttt{cv2.SOLVEPNP\_ITERATIVE}. The solver returns $R_{\rm rec}$ and $\bm t_{\rm rec}$. Rotational accuracy is evaluated by the geodesic angular distance on $\SO$:
\begin{equation}\theta_{\rm err}=\arccos\left(\frac{\operatorname{tr}(R_{\rm true}^{\mathsf T}R_{\rm rec})-1}{2}\right).\end{equation}
The argument of $\arccos$ is clamped to $[-1,1]$. The angular distance from the frontal reference orientation is
\begin{equation}\alpha(\phi,\theta)=d_{\SO}(I,R_{\rm true})=\arccos\left(\frac{\operatorname{tr}(R_{\rm true})-1}{2}\right).\end{equation}

\subsection*{S1.6. Aggregation and visualization}
For each roll--pitch pair, the reference-case estimation error is
\begin{equation}E_{\rm ref}(\phi,\theta)=\theta_{\rm err}^{\rm ref}(\phi,\theta),\end{equation}
and the perturbed-case mean and standard deviation are
\begin{equation}\bar E(\phi,\theta)=\frac{1}{16}\sum_{k=1}^{16}\theta_{\rm err}^{(k)}(\phi,\theta),\end{equation}
\begin{equation}\sigma_E(\phi,\theta)=\sqrt{\frac{1}{16}\sum_{k=1}^{16}\left(\theta_{\rm err}^{(k)}(\phi,\theta)-\bar E(\phi,\theta)\right)^2}.\end{equation}
\begin{figure}[!htbp]\centering\includegraphics[width=.95\linewidth]{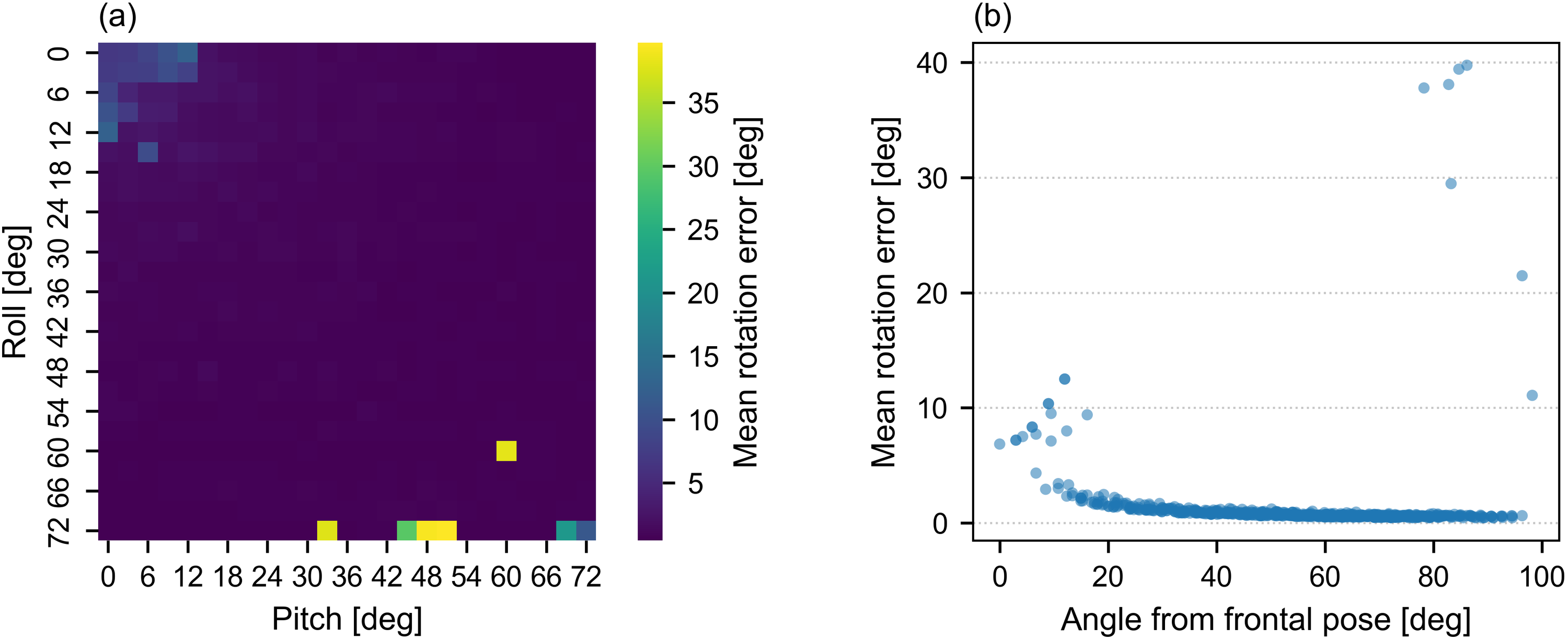}\caption{Simulation-based rationale for selecting a moderate tag tilt. (a) Heatmap of the mean rotation error $\bar E$ as a function of roll and pitch, computed from the 16 one-corner pixel-perturbation cases at each pose. (b) Scatter plot of the same mean rotation error against the geodesic angular distance from the frontal reference orientation.}\label{fig:S1}\end{figure}

\subsection*{S1.7. Implications for selecting the base tilt}
Figure~\ref{fig:S1} indicates that, over the moderate inclinations at which a planar tag remains well resolved, the mean pose-estimation error is highest near exactly frontal observation and decreases markedly once the marker is viewed at a modest inclination, especially around $15$--$20\tcdegree$ from the frontal pose. The few much larger values at the oblique end of the simulated range arise where the square is seen nearly edge-on and would not be decoded reliably in practice, so they lie outside the operating range that motivates the design. This motivated the $18\tcdegree$ base tilt in the fabricated quARtet layouts as a practical compromise between pose-estimation stability and compact, fabrication-friendly geometry. The simulation is intended as a qualitative design guide and does not model multi-tag occlusion, printing defects, lens distortion, or other non-ideal effects.

\section*{S2. Per-image pose estimates for the fixed-camera experiments}
\begin{figure}[!htbp]\centering
\begin{minipage}[b]{0.48\linewidth}\centering\includegraphics[width=.95\linewidth]{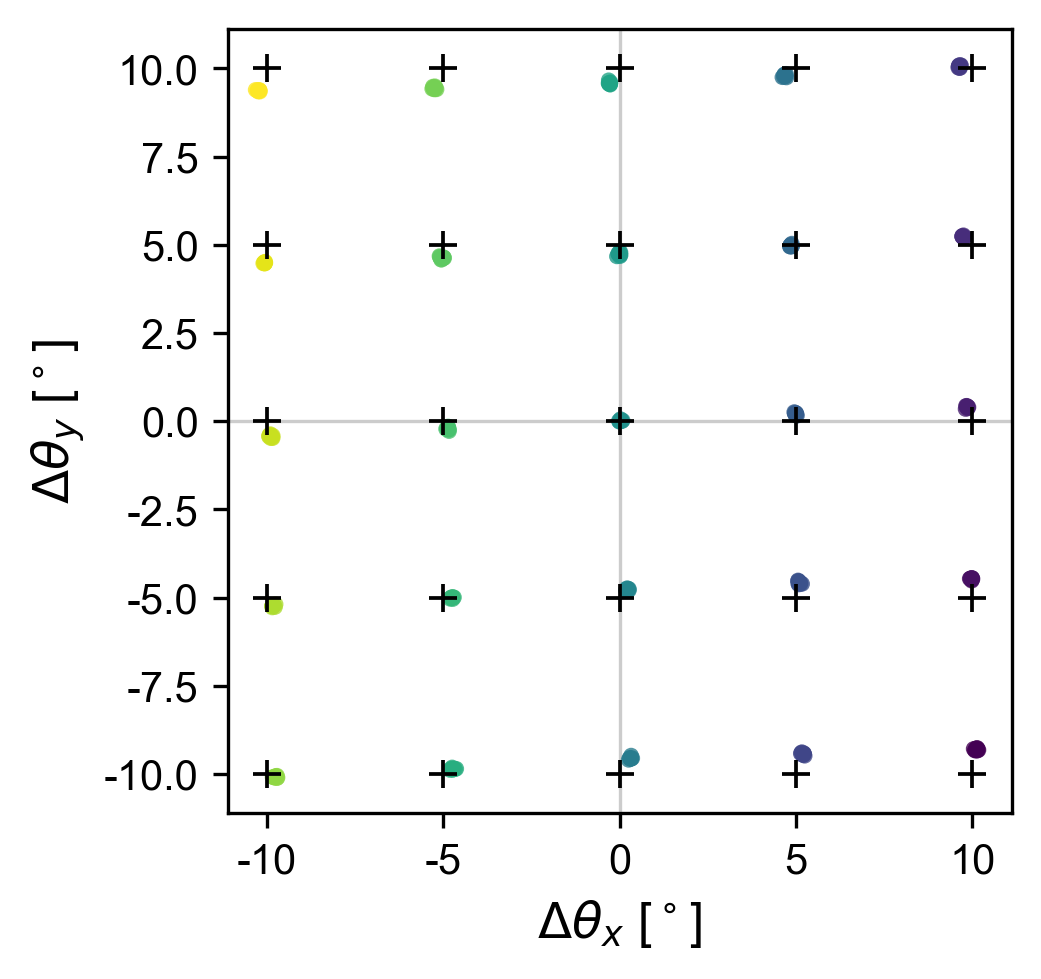}\\{\footnotesize (a) quARtet-D}\end{minipage}\hfill
\begin{minipage}[b]{0.48\linewidth}\centering\includegraphics[width=.95\linewidth]{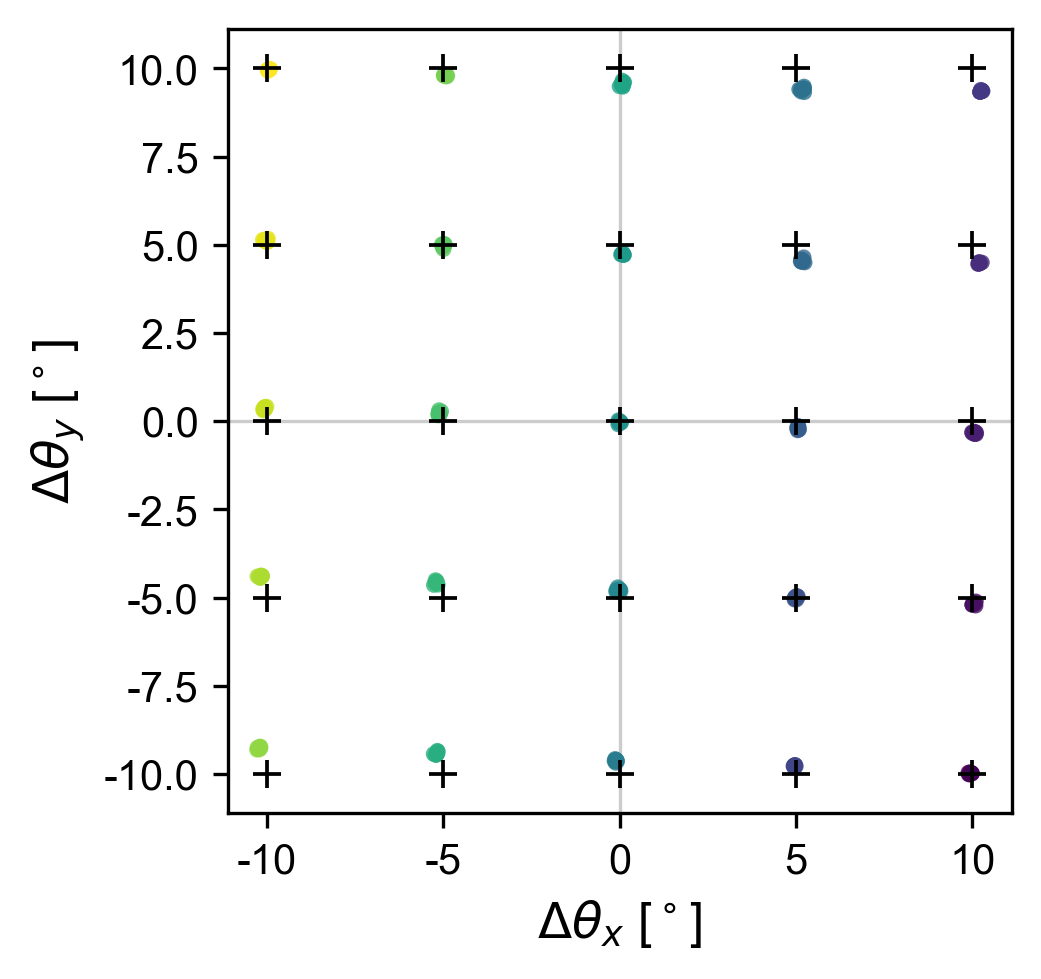}\\{\footnotesize (b) quARtet-P}\end{minipage}
\vspace{1ex}
\begin{minipage}[b]{0.48\linewidth}\centering\includegraphics[width=.95\linewidth]{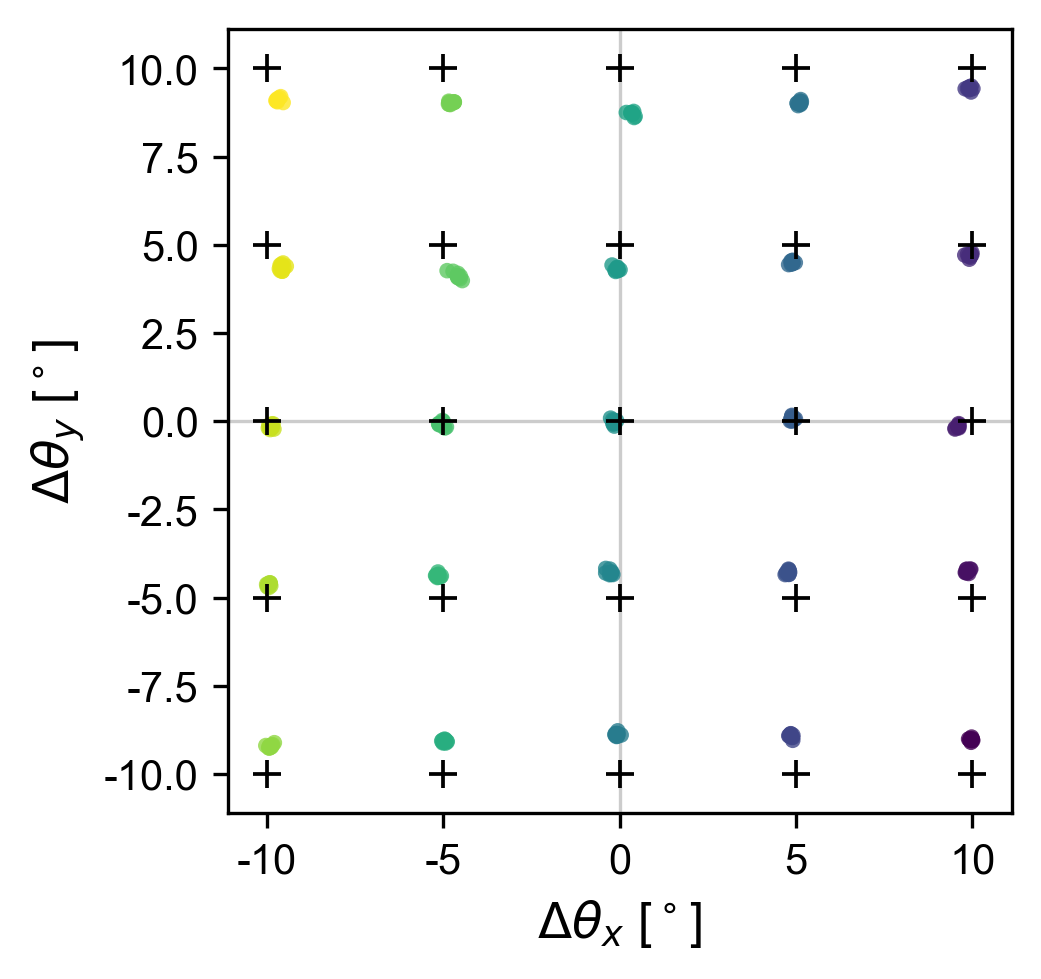}\\{\footnotesize (c) quARtet-rP}\end{minipage}\hfill
\begin{minipage}[b]{0.48\linewidth}\centering\includegraphics[width=.95\linewidth]{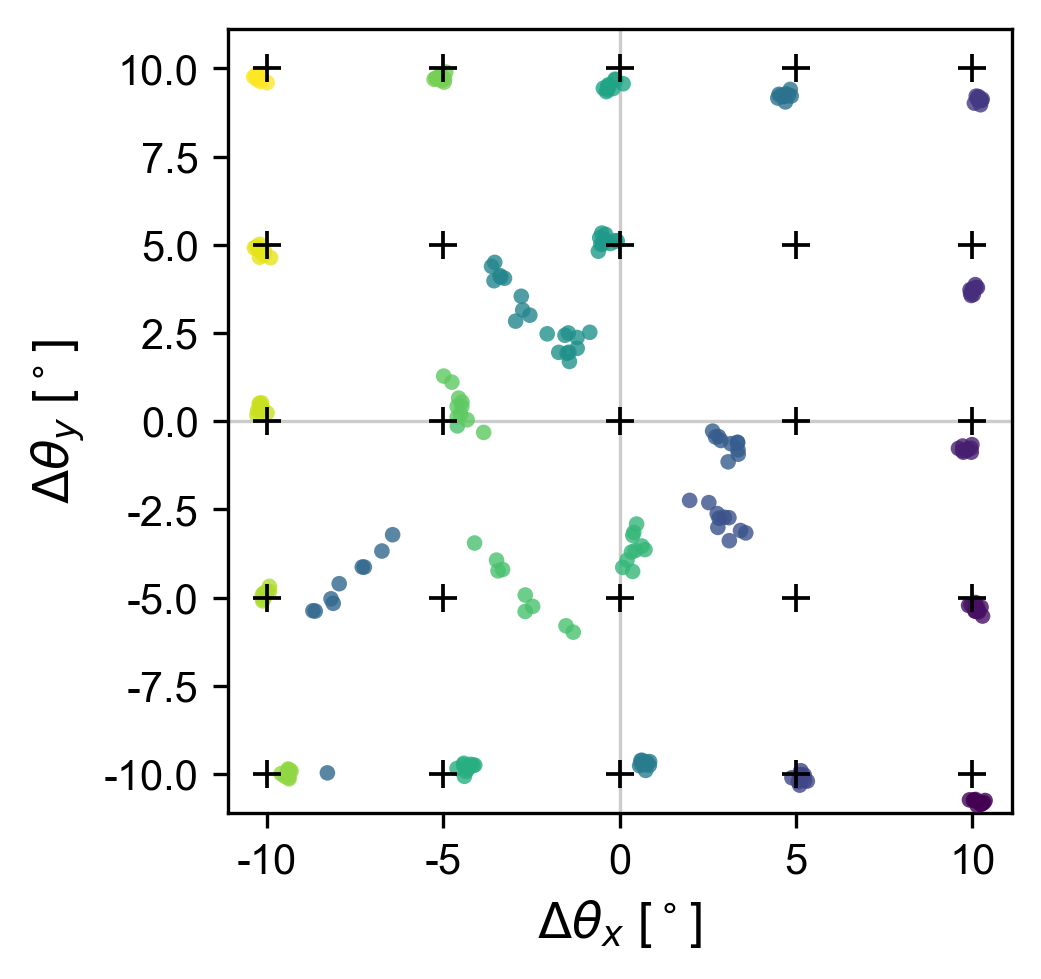}\\{\footnotesize (d) Single}\end{minipage}
\caption{Per-image orientation estimates on the narrow grid ($-10\tcdegree$ to $+10\tcdegree$ in $5\tcdegree$ increments). Black plus signs mark commanded orientations, and dot color distinguishes the commanded cells without encoding error. Coordinates follow the operational-reference and constant-alignment conventions of main-text Section~4.2.1 and Section~S3.}\label{fig:S8}\end{figure}

\begin{figure}[!htbp]\centering
\begin{minipage}[b]{0.48\linewidth}\centering\includegraphics[width=.95\linewidth]{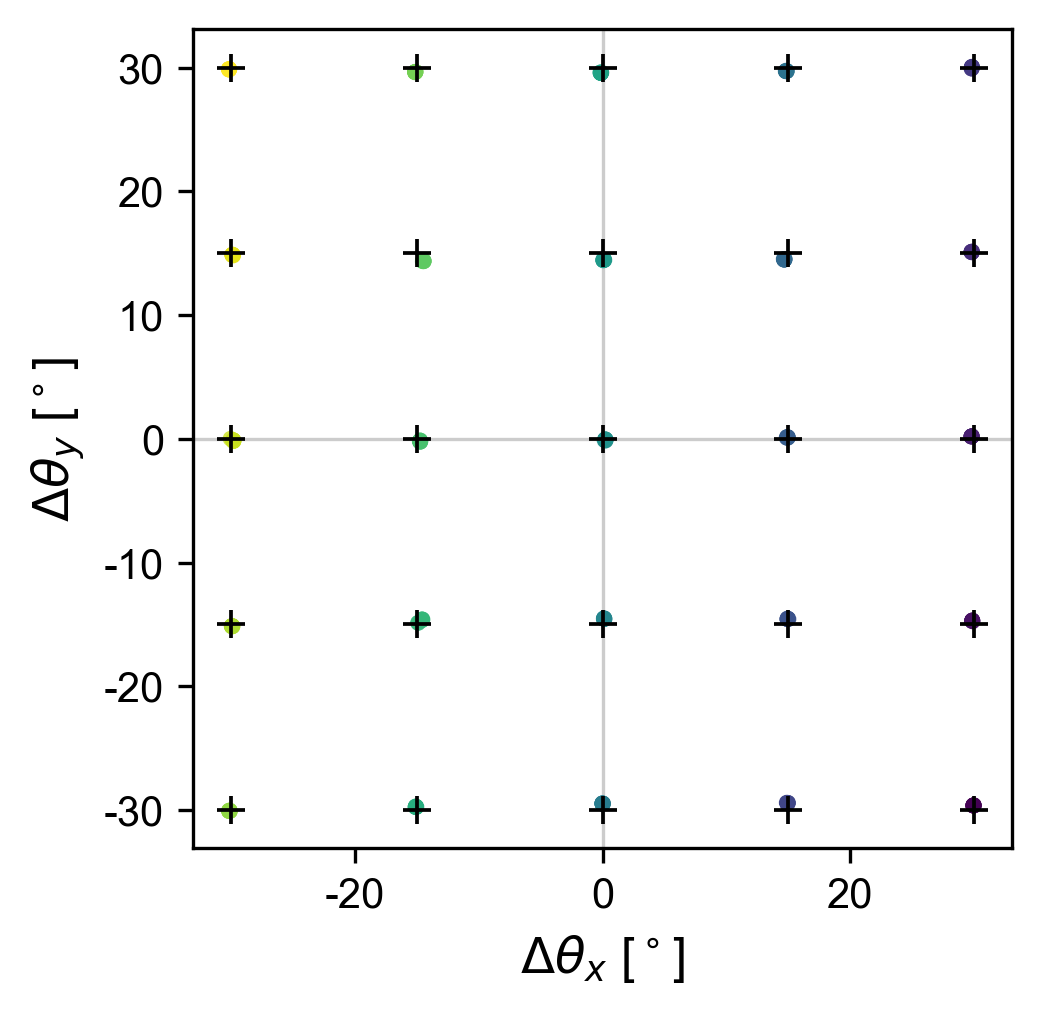}\\{\footnotesize (a) quARtet-D}\end{minipage}\hfill
\begin{minipage}[b]{0.48\linewidth}\centering\includegraphics[width=.95\linewidth]{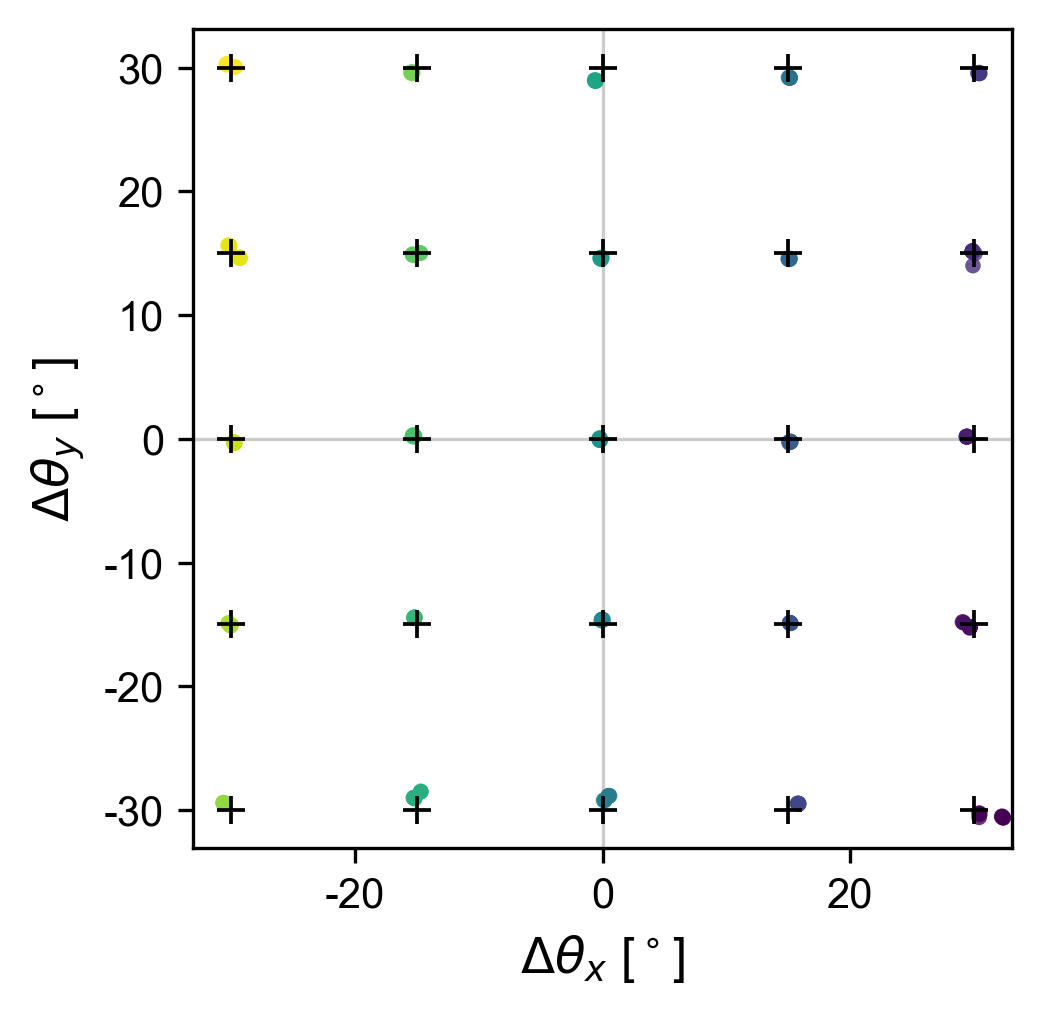}\\{\footnotesize (b) quARtet-P}\end{minipage}
\vspace{1ex}
\begin{minipage}[b]{0.48\linewidth}\centering\includegraphics[width=.95\linewidth]{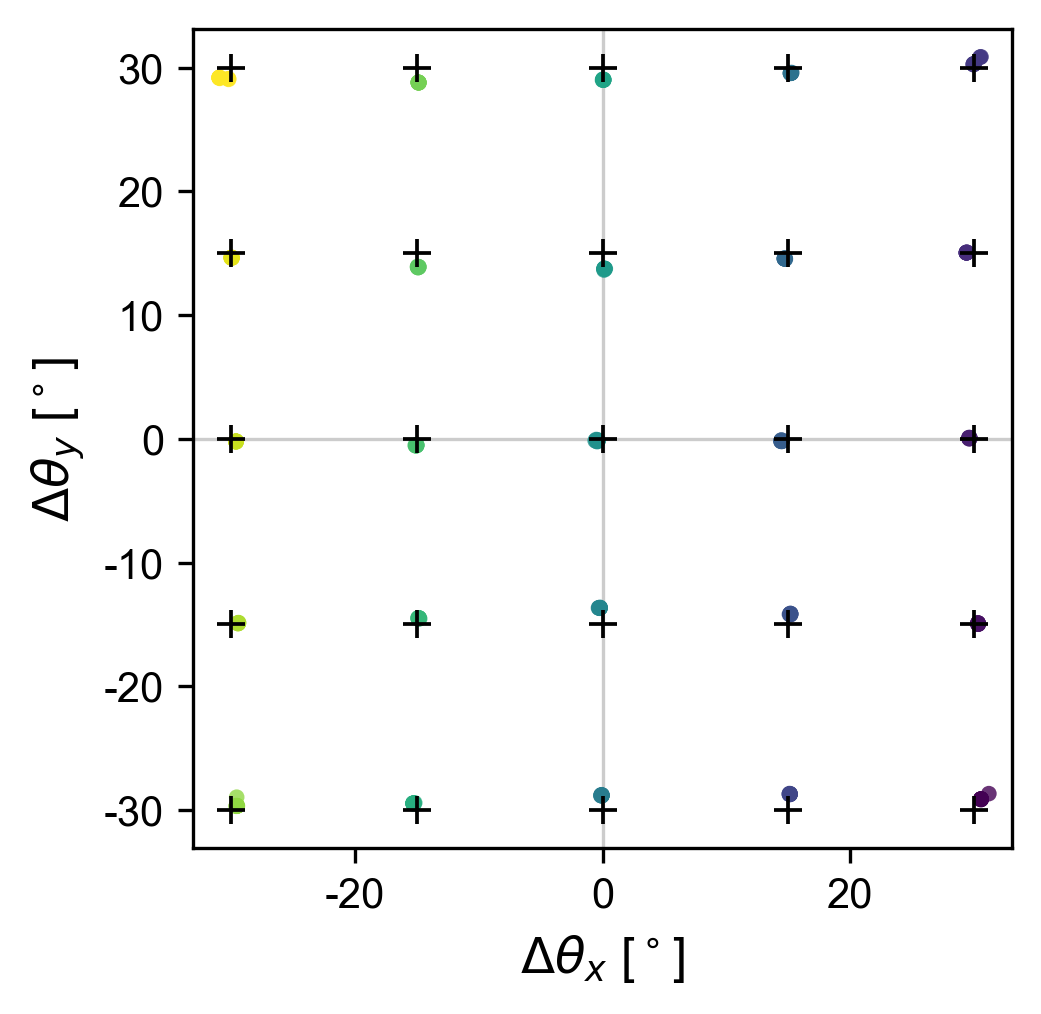}\\{\footnotesize (c) quARtet-rP}\end{minipage}\hfill
\begin{minipage}[b]{0.48\linewidth}\centering\includegraphics[width=.95\linewidth]{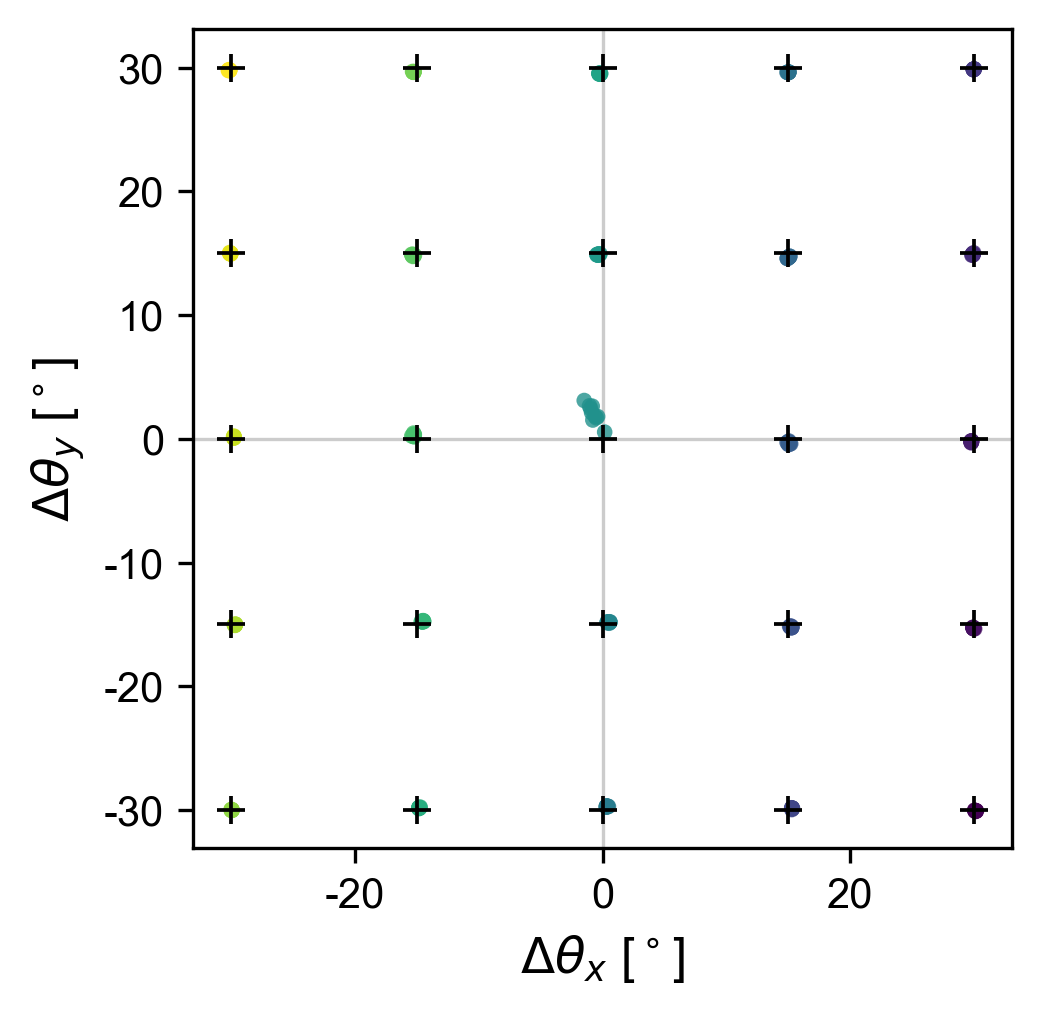}\\{\footnotesize (d) Single}\end{minipage}
\caption{Per-image orientation estimates on the broad grid ($-30\tcdegree$ to $+30\tcdegree$ in $15\tcdegree$ increments). Black plus signs mark commanded orientations. Conventions as in Fig.~\ref{fig:S8}.}\label{fig:S9}\end{figure}

This section provides per-image scatter for the orientation (Figs.~\ref{fig:S8} and~\ref{fig:S9}) and position (Fig.~\ref{fig:S10}) experiments in main-text Section~4.2, including the additional narrow-angle grid. Within every figure the panel order is quARtet-D, quARtet-P, quARtet-rP, and Single, and the axes are shared for direct comparison.

\subsection*{S2.1. Orientation estimates}

The narrow grid shows that Single can exhibit its largest spread within a small band near, rather than necessarily at, the exactly frontal pose. The quARtet layouts do not show this near-frontal spread.

\subsection*{S2.2. Position estimates}
\begin{figure}[!htbp]\centering
\begin{minipage}[b]{0.48\linewidth}\centering\includegraphics[width=.95\linewidth]{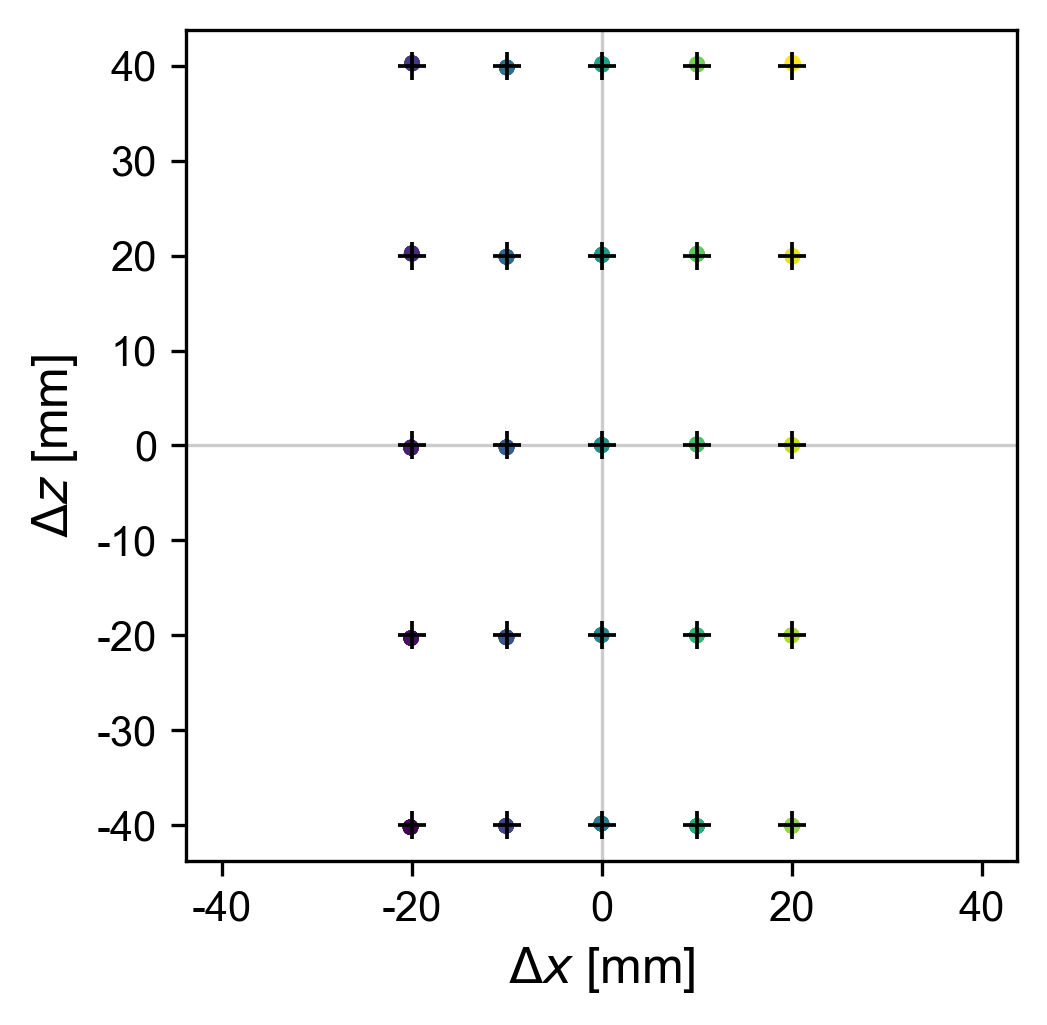}\\{\footnotesize (a) quARtet-D}\end{minipage}\hfill
\begin{minipage}[b]{0.48\linewidth}\centering\includegraphics[width=.95\linewidth]{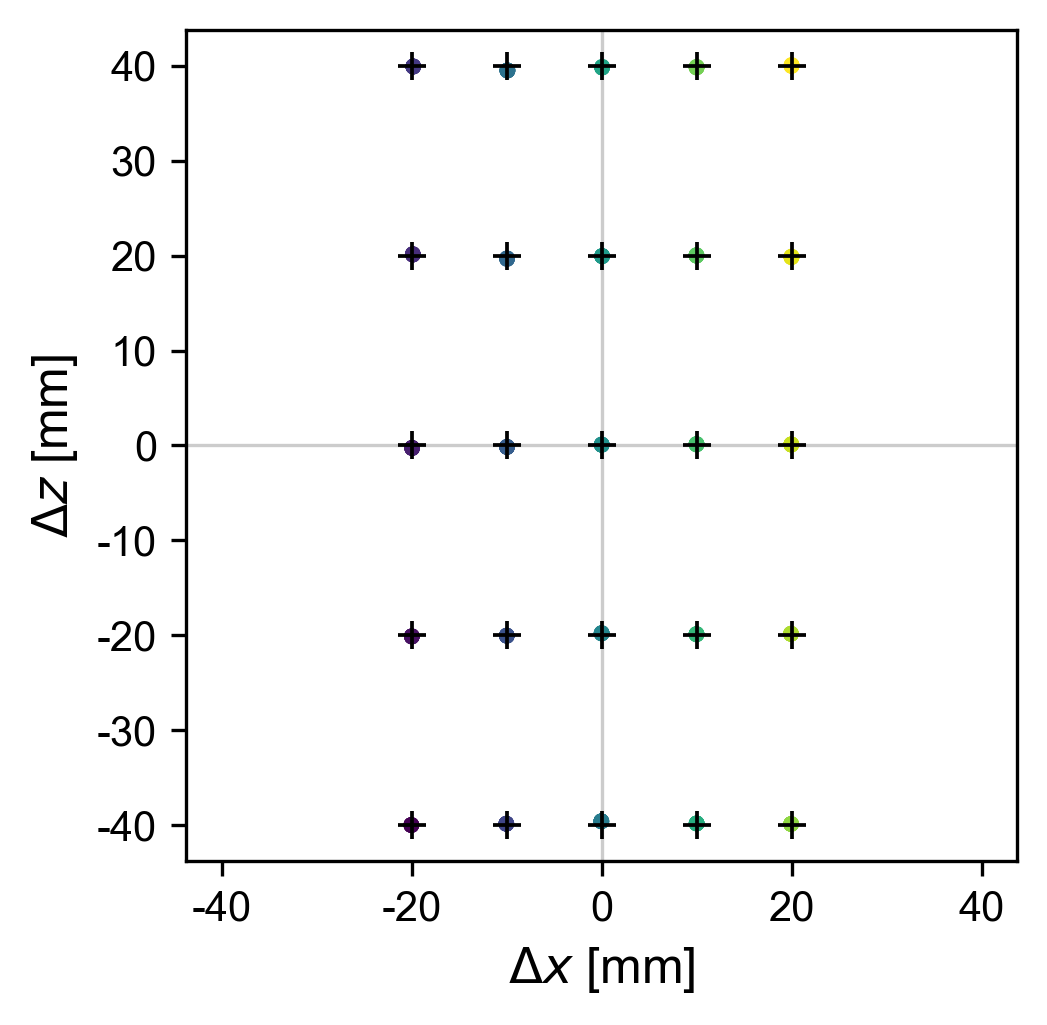}\\{\footnotesize (b) quARtet-P}\end{minipage}
\vspace{1ex}
\begin{minipage}[b]{0.48\linewidth}\centering\includegraphics[width=.95\linewidth]{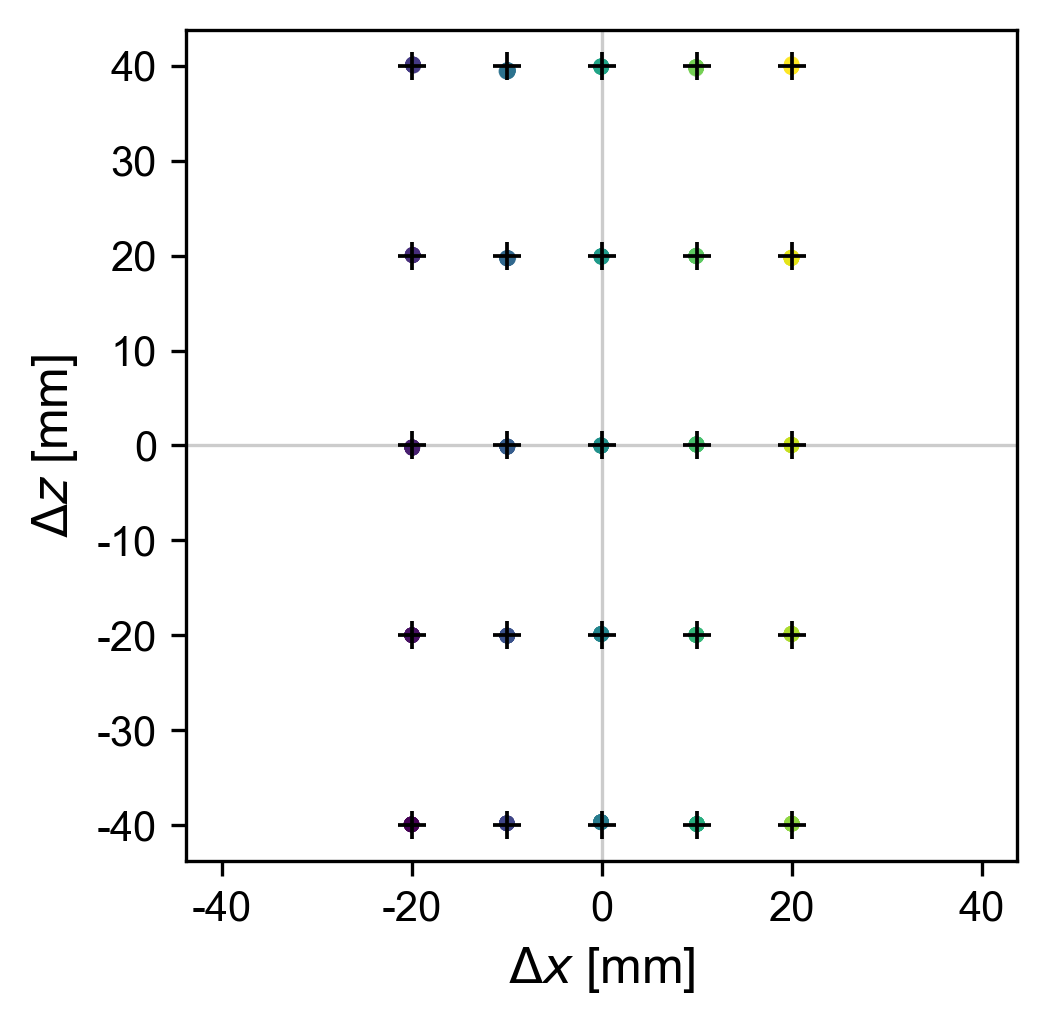}\\{\footnotesize (c) quARtet-rP}\end{minipage}\hfill
\begin{minipage}[b]{0.48\linewidth}\centering\includegraphics[width=.95\linewidth]{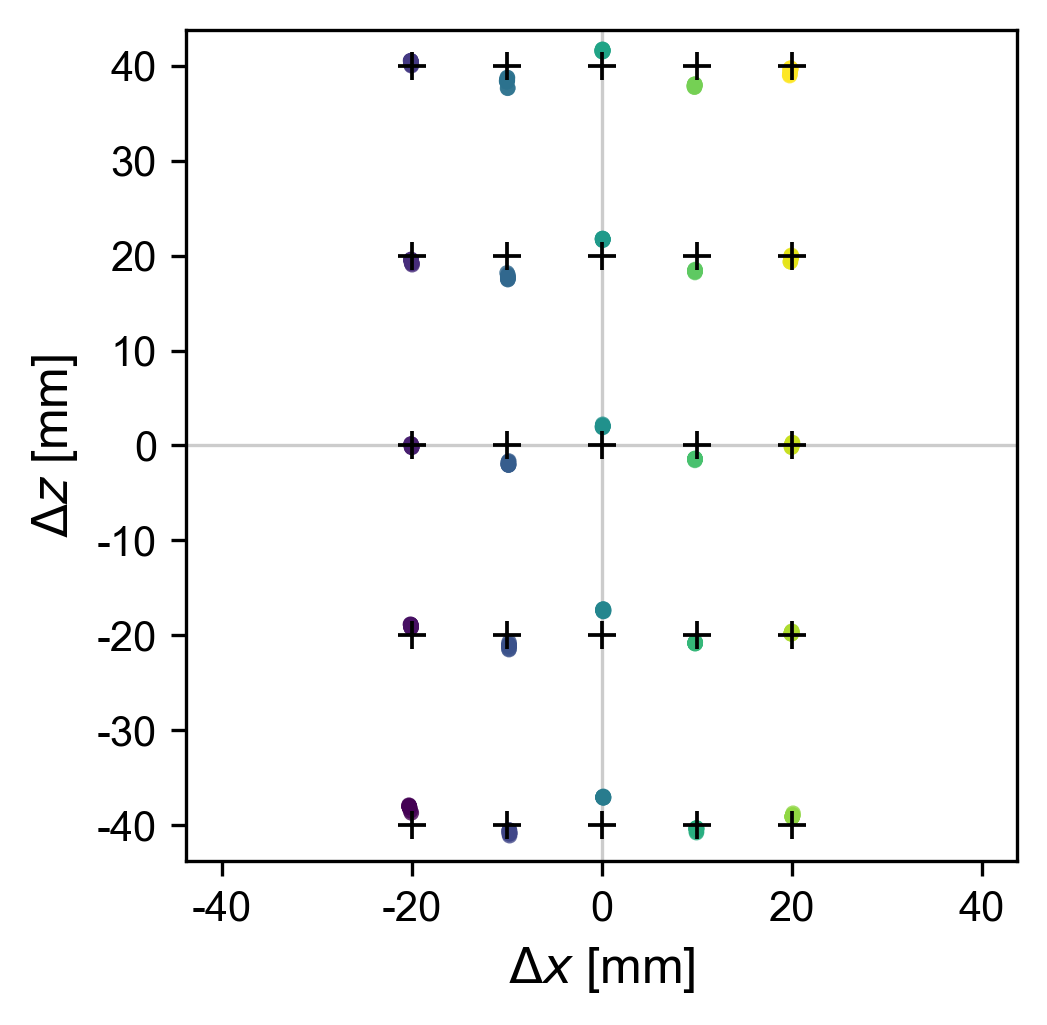}\\{\footnotesize (d) Single}\end{minipage}
\caption{Per-image position estimates over the commanded translation grid. Black plus signs mark commanded positions, and dot color distinguishes the commanded cells without encoding error. Coordinates follow the operational-reference and constant-alignment conventions of main-text Section~4.2.1 and Section~S3.}\label{fig:S10}\end{figure}

\subsection*{S2.3. Cell-wise comparison against Single (within-acquisition, descriptive)}
Tables~\ref{tab:cellwise_ori} and~\ref{tab:cellwise_pos} report, for every commanded cell, the per-image mean$\pm$SD error of each quARtet layout and of Single. Because the 10 images of a cell are consecutive technical repeats of one held pose at one mounting, these are descriptive within-acquisition summaries: no frame-level significance claims are attached, and independent re-mounting or re-acquisition runs would be required for inference beyond this setup. The pattern runs in both directions: all three layouts have far smaller errors than Single at the frontal orientation cell and over almost the entire translation grid, whereas at oblique orientation cells, where a planar tag already has small errors, quARtet-P and quARtet-rP are often modestly worse (cell-mean increases of up to $+1.3\tcdegree$, with absolute cell means of up to $1.6\tcdegree$). quARtet-D is also above Single at 10 of the 24 oblique cells, by at most about $+0.4\tcdegree$, and stays at or below $0.8\tcdegree$. The one exception in the position grid is the cell at $(\Delta x,\Delta z)=(-20,0)$, where all three layouts, including quARtet-D, are slightly worse (at most $+0.11$~mm).

\begin{figure}[!htbp]\centering
\begin{minipage}[b]{0.32\linewidth}\centering\includegraphics[width=\linewidth]{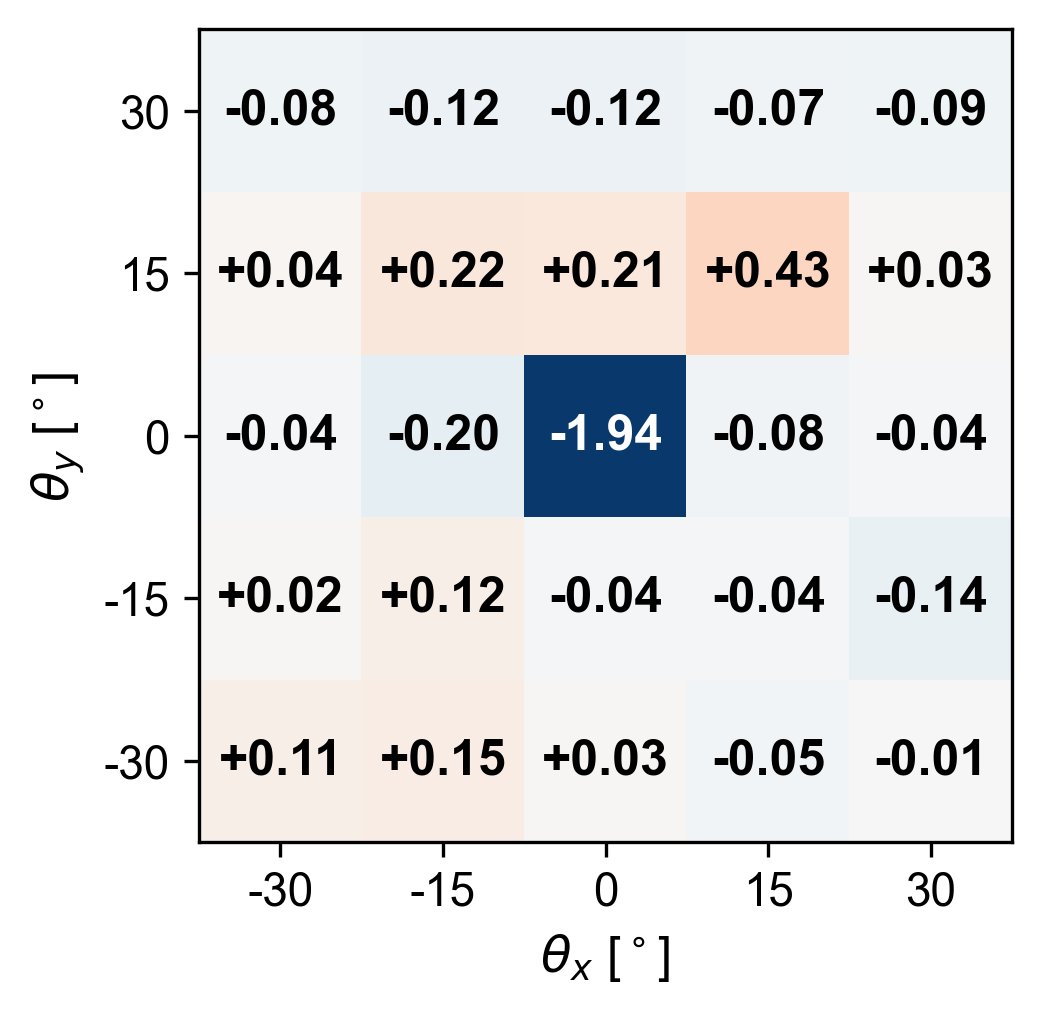}\\{\footnotesize (a) quARtet-D $-$ Single}\end{minipage}\hfill
\begin{minipage}[b]{0.32\linewidth}\centering\includegraphics[width=\linewidth]{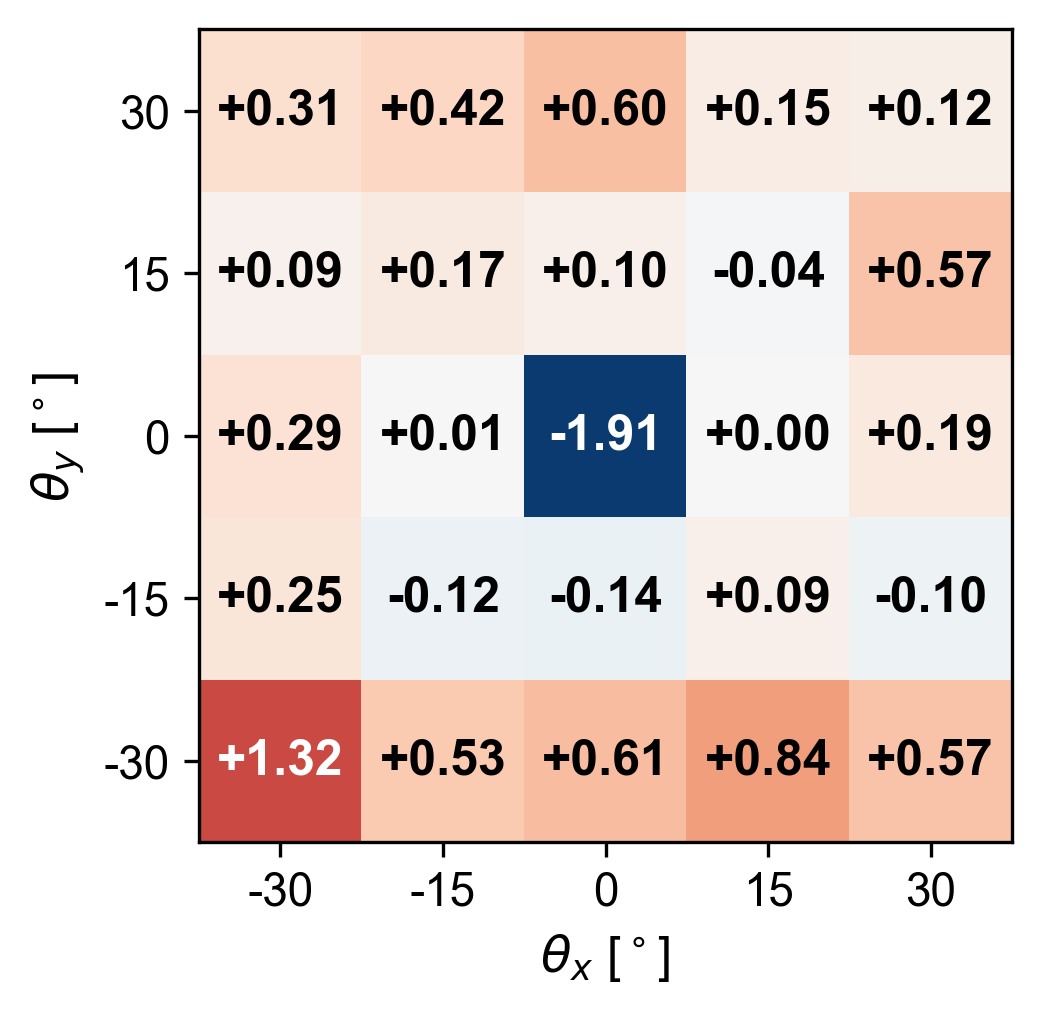}\\{\footnotesize (b) quARtet-P $-$ Single}\end{minipage}\hfill
\begin{minipage}[b]{0.32\linewidth}\centering\includegraphics[width=\linewidth]{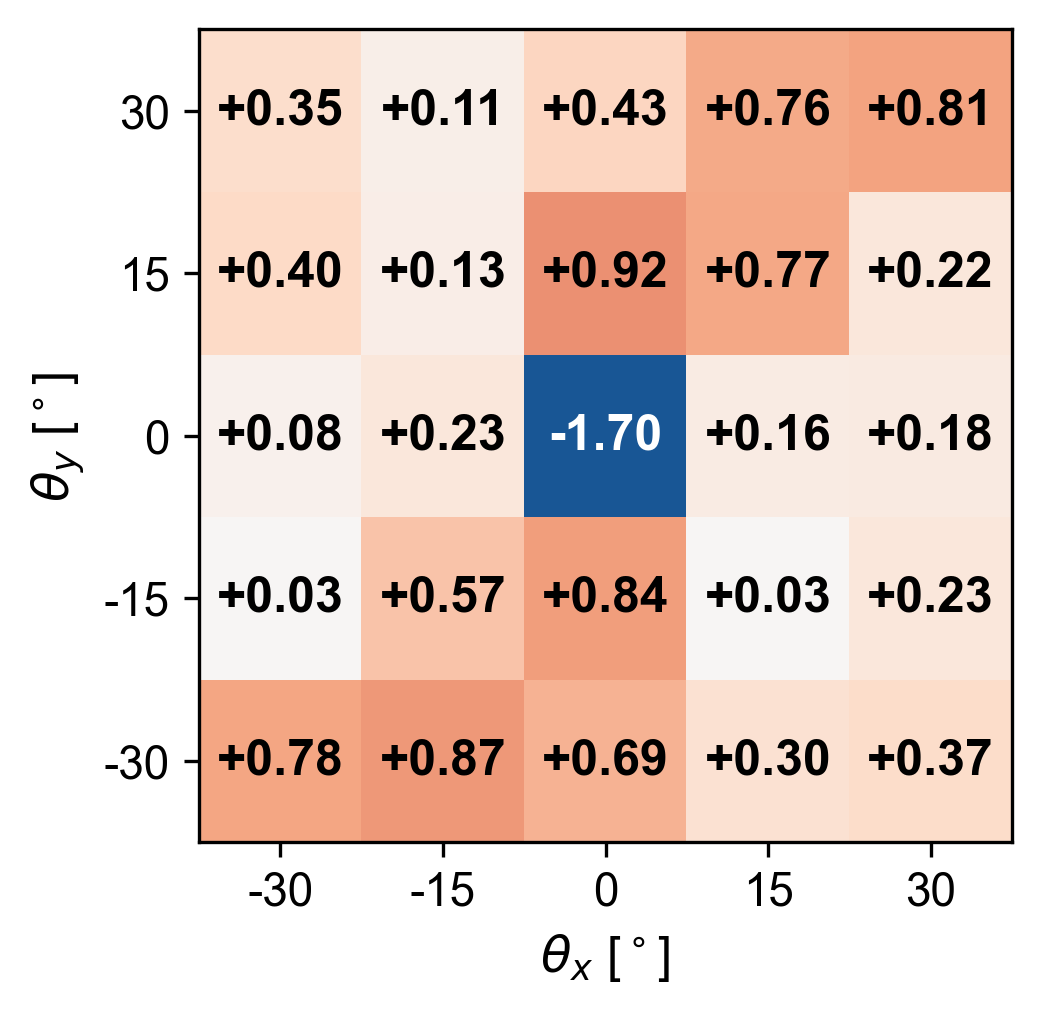}\\{\footnotesize (c) quARtet-rP $-$ Single}\end{minipage}\\[0.6ex]
\includegraphics[width=0.55\linewidth]{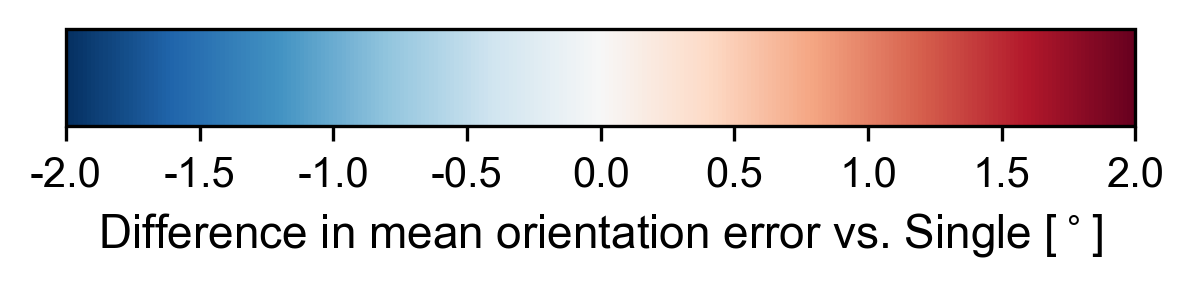}
\caption{Difference maps of the per-cell mean orientation error on the broad grid (quARtet layout minus Single, so blue means the layout's error is smaller). The map is a descriptive summary of the cell-mean differences. No frame-level test is applied, because the frames of a cell are consecutive technical repeats of one held pose. The large improvement is concentrated at the frontal cell, while at oblique cells the pitch-based layouts pay penalties of up to about $+1.3\tcdegree$ in cell-mean error and quARtet-D of up to about $+0.4\tcdegree$. Differences are computed from unrounded cell means, so end digits can differ by $0.01$ from the rounded values of Table~\ref{tab:cellwise_ori}.}\label{fig:diffmap_ori}\end{figure}

\begin{figure}[!htbp]\centering
\begin{minipage}[b]{0.32\linewidth}\centering\includegraphics[width=\linewidth]{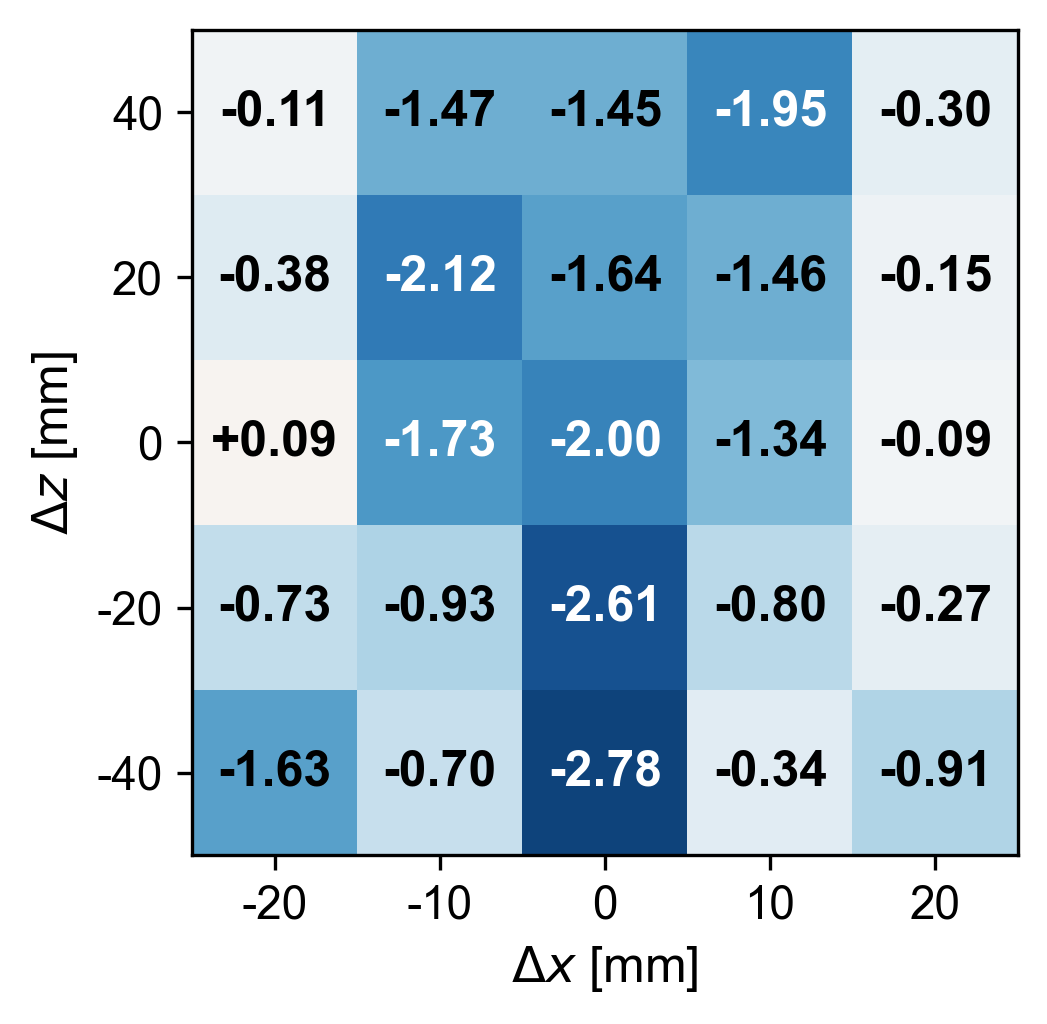}\\{\footnotesize (a) quARtet-D $-$ Single}\end{minipage}\hfill
\begin{minipage}[b]{0.32\linewidth}\centering\includegraphics[width=\linewidth]{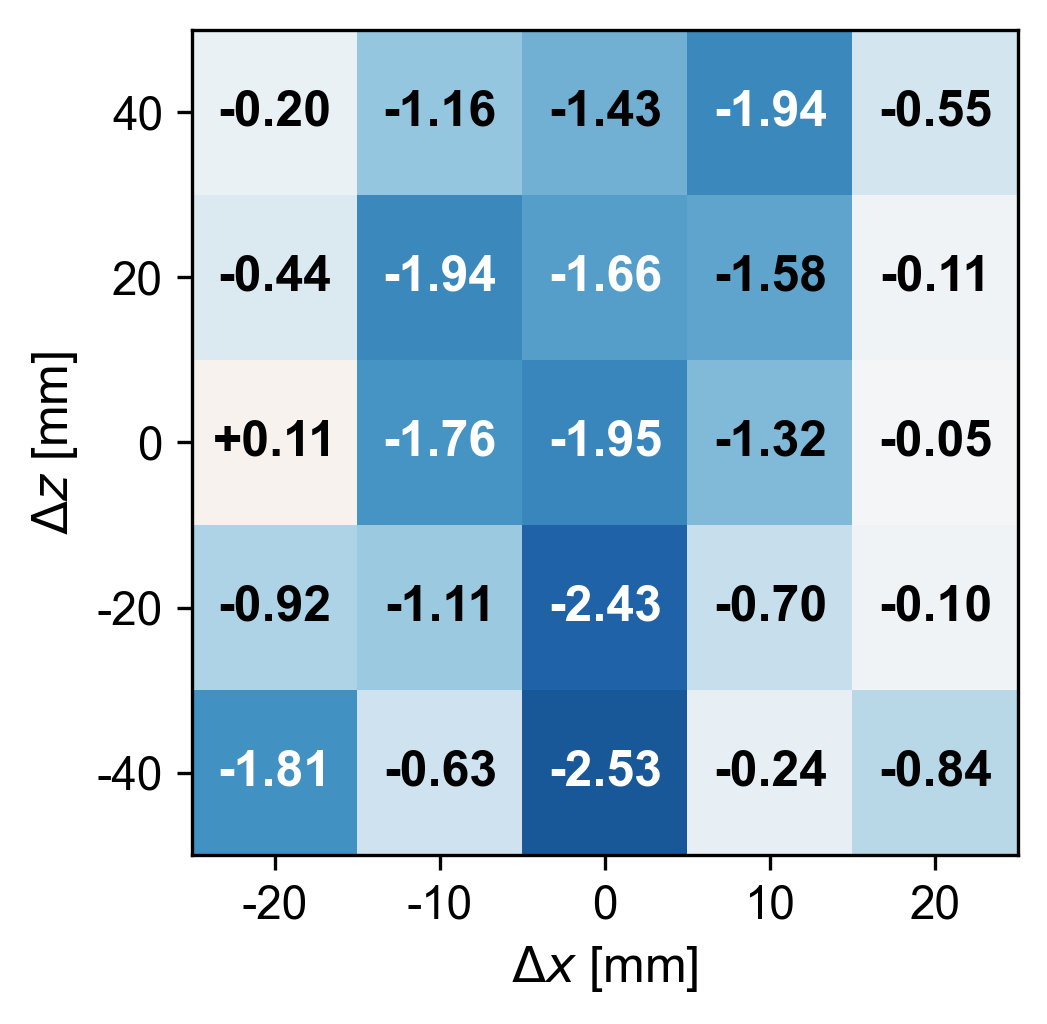}\\{\footnotesize (b) quARtet-P $-$ Single}\end{minipage}\hfill
\begin{minipage}[b]{0.32\linewidth}\centering\includegraphics[width=\linewidth]{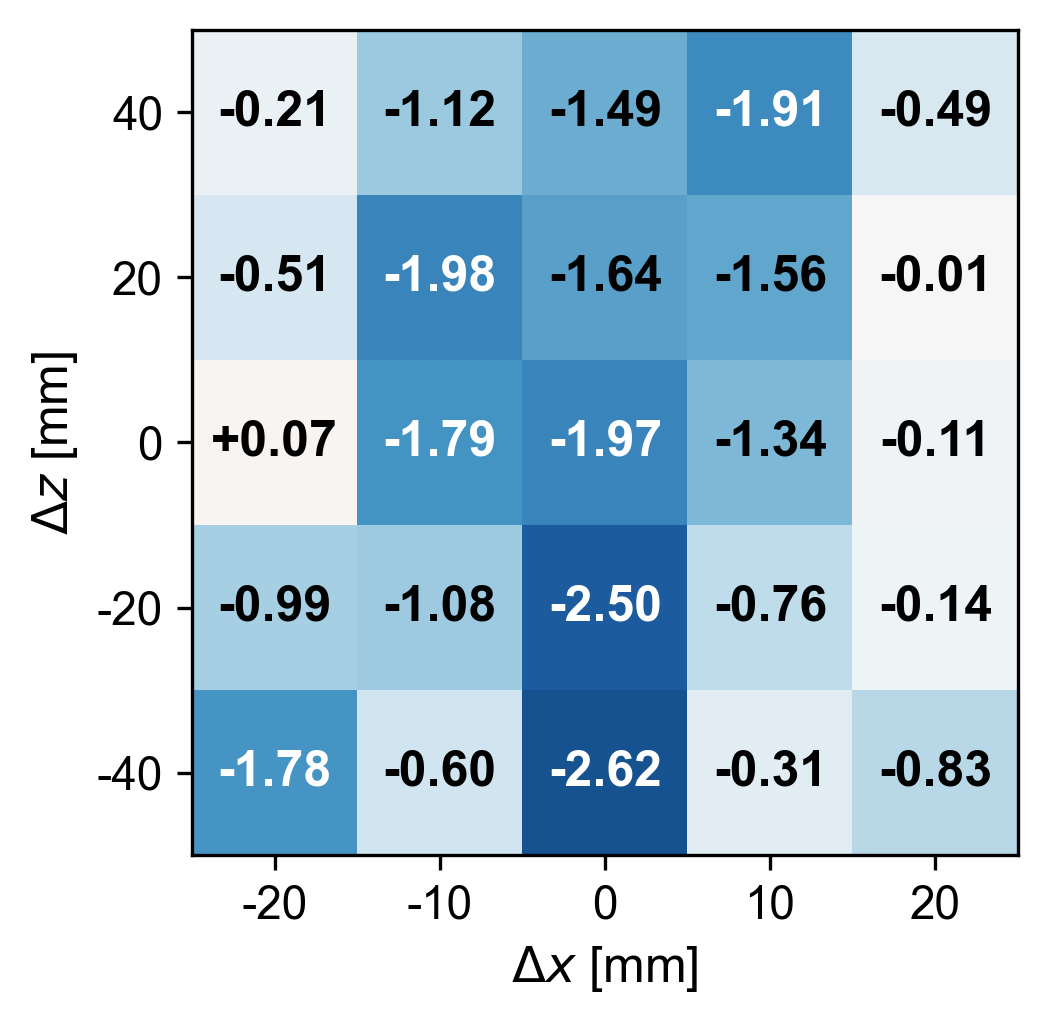}\\{\footnotesize (c) quARtet-rP $-$ Single}\end{minipage}\\[0.6ex]
\includegraphics[width=0.55\linewidth]{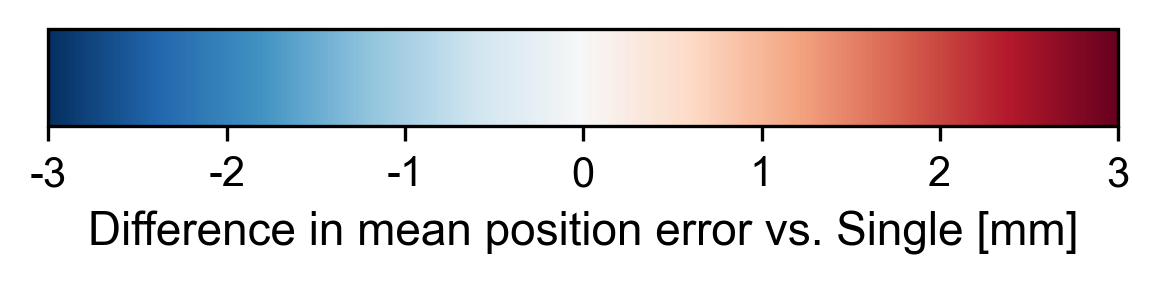}
\caption{Difference maps of the per-cell mean position error (quARtet layout minus Single, so blue means the layout's error is smaller). The map is a descriptive summary of the cell-mean differences. No frame-level test is applied, because the frames of a cell are consecutive technical repeats of one held pose. All three layouts have smaller errors over almost the entire grid, the single exception being the cell at $(-20,0)$ (at most $+0.11$~mm). Differences are computed from unrounded cell means.}\label{fig:diffmap_pos}\end{figure}

Figures~\ref{fig:diffmap_ori} and~\ref{fig:diffmap_pos} visualize the same comparisons at a glance. Each panel maps the difference in per-cell mean error between one quARtet layout and Single (blue where the layout's error is smaller, red where it is larger). The maps are descriptive: they display the same cell means as Tables~\ref{tab:cellwise_ori} and~\ref{tab:cellwise_pos}, with no frame-level test and no significance marking. These maps are an auxiliary, pairwise view. The absolute per-marker error maps in main-text Figs.~6--7 remain the characterization of each marker on its own.

\begin{table}[!htbp]
\centering\scriptsize
\setlength{\tabcolsep}{4pt}
\caption{Cell-wise orientation residual on the broad grid: mean$\pm$SD over the valid consecutive frames of each commanded cell, in degrees, in the common result order with the Single control last. The smallest mean in each row is bold (ties resolved on unrounded means). Frames within a cell are technical repeats of one held pose at one mounting, so these are descriptive within-acquisition summaries, and no inferential asterisks are attached. The companion difference maps (Figs.~\ref{fig:diffmap_ori} and~\ref{fig:diffmap_pos}) display the same cell-mean differences graphically.}
\label{tab:cellwise_ori}
\begin{tabular}{@{}lcccc@{}}
\toprule
$(\theta_x,\theta_y)$ [$\tcdegree$] & quARtet-D & quARtet-P & quARtet-rP & Single \\
\midrule
$(-30,-30)$ & $0.42\pm0.02$ & $1.63\pm0.88$ & $1.09\pm0.21$ & $\mathbf{0.31\pm0.03}$ \\
$(-30,-15)$ & $0.33\pm0.03$ & $0.56\pm0.21$ & $0.34\pm0.02$ & $\mathbf{0.31\pm0.04}$ \\
$(-30,+0)$ & $\mathbf{0.27\pm0.01}$ & $0.60\pm0.03$ & $0.39\pm0.03$ & $0.31\pm0.05$ \\
$(-30,+15)$ & $0.22\pm0.02$ & $0.27\pm0.28$ & $0.58\pm0.01$ & $\mathbf{0.18\pm0.04}$ \\
$(-30,+30)$ & $\mathbf{0.24\pm0.01}$ & $0.63\pm0.03$ & $0.67\pm0.26$ & $0.32\pm0.01$ \\
$(-15,-30)$ & $0.60\pm0.02$ & $0.97\pm0.02$ & $1.32\pm0.02$ & $\mathbf{0.45\pm0.02}$ \\
$(-15,-15)$ & $0.45\pm0.03$ & $\mathbf{0.22\pm0.02}$ & $0.90\pm0.03$ & $0.33\pm0.06$ \\
$(-15,+0)$ & $\mathbf{0.14\pm0.03}$ & $0.35\pm0.02$ & $0.57\pm0.04$ & $0.34\pm0.10$ \\
$(-15,+15)$ & $0.61\pm0.03$ & $0.57\pm0.02$ & $0.52\pm0.02$ & $\mathbf{0.39\pm0.05}$ \\
$(-15,+30)$ & $\mathbf{0.38\pm0.02}$ & $0.92\pm0.02$ & $0.61\pm0.01$ & $0.50\pm0.01$ \\
$(+0,-30)$ & $0.54\pm0.02$ & $1.12\pm0.22$ & $1.19\pm0.03$ & $\mathbf{0.51\pm0.04}$ \\
$(+0,-15)$ & $0.48\pm0.02$ & $\mathbf{0.39\pm0.03}$ & $1.37\pm0.02$ & $0.52\pm0.08$ \\
$(+0,+0)$ & $\mathbf{0.24\pm0.03}$ & $0.27\pm0.03$ & $0.47\pm0.07$ & $2.18\pm0.80$ \\
$(+0,+15)$ & $0.57\pm0.02$ & $0.46\pm0.04$ & $1.28\pm0.04$ & $\mathbf{0.36\pm0.07}$ \\
$(+0,+30)$ & $\mathbf{0.47\pm0.03}$ & $1.18\pm0.03$ & $1.01\pm0.03$ & $0.58\pm0.03$ \\
$(+15,-30)$ & $\mathbf{0.28\pm0.02}$ & $1.16\pm0.25$ & $0.62\pm0.01$ & $0.33\pm0.03$ \\
$(+15,-15)$ & $\mathbf{0.48\pm0.16}$ & $0.61\pm0.03$ & $0.54\pm0.02$ & $0.52\pm0.06$ \\
$(+15,+0)$ & $\mathbf{0.33\pm0.03}$ & $0.40\pm0.05$ & $0.56\pm0.03$ & $0.40\pm0.07$ \\
$(+15,+15)$ & $0.80\pm0.04$ & $\mathbf{0.34\pm0.04}$ & $1.15\pm0.03$ & $0.37\pm0.05$ \\
$(+15,+30)$ & $\mathbf{0.39\pm0.02}$ & $0.61\pm0.03$ & $1.22\pm0.02$ & $0.46\pm0.02$ \\
$(+30,-30)$ & $\mathbf{0.20\pm0.02}$ & $0.77\pm0.04$ & $0.57\pm0.19$ & $0.20\pm0.01$ \\
$(+30,-15)$ & $\mathbf{0.19\pm0.02}$ & $0.23\pm0.05$ & $0.56\pm0.02$ & $0.33\pm0.03$ \\
$(+30,+0)$ & $\mathbf{0.22\pm0.06}$ & $0.45\pm0.04$ & $0.44\pm0.01$ & $0.26\pm0.03$ \\
$(+30,+15)$ & $0.18\pm0.03$ & $0.72\pm0.06$ & $0.37\pm0.02$ & $\mathbf{0.15\pm0.02}$ \\
$(+30,+30)$ & $\mathbf{0.21\pm0.02}$ & $0.42\pm0.04$ & $1.11\pm0.11$ & $0.30\pm0.03$ \\
\bottomrule
\end{tabular}
\end{table}

\begin{table}[!htbp]
\centering\scriptsize
\setlength{\tabcolsep}{4pt}
\caption{Cell-wise Euclidean position residual on the translation grid: mean$\pm$SD over the valid consecutive frames of each commanded cell, in mm, in the common result order with the Single control last. The smallest mean in each row is bold (ties resolved on unrounded means). Frames within a cell are technical repeats of one held pose at one mounting, so these are descriptive within-acquisition summaries, and no inferential asterisks are attached.}
\label{tab:cellwise_pos}
\begin{tabular}{@{}lcccc@{}}
\toprule
$(\Delta x,\Delta z)$ [mm] & quARtet-D & quARtet-P & quARtet-rP & Single \\
\midrule
$(-20,-40)$ & $0.25\pm0.01$ & $\mathbf{0.06\pm0.01}$ & $0.10\pm0.01$ & $1.87\pm0.26$ \\
$(-20,-20)$ & $0.29\pm0.01$ & $0.11\pm0.02$ & $\mathbf{0.03\pm0.01}$ & $1.02\pm0.14$ \\
$(-20,+0)$ & $0.24\pm0.02$ & $0.26\pm0.02$ & $0.21\pm0.03$ & $\mathbf{0.15\pm0.04}$ \\
$(-20,+20)$ & $0.23\pm0.03$ & $0.16\pm0.02$ & $\mathbf{0.10\pm0.02}$ & $0.61\pm0.20$ \\
$(-20,+40)$ & $0.26\pm0.03$ & $0.17\pm0.01$ & $\mathbf{0.16\pm0.02}$ & $0.37\pm0.20$ \\
$(-10,-40)$ & $\mathbf{0.10\pm0.01}$ & $0.17\pm0.01$ & $0.21\pm0.01$ & $0.80\pm0.18$ \\
$(-10,-20)$ & $0.21\pm0.01$ & $\mathbf{0.03\pm0.02}$ & $0.06\pm0.01$ & $1.14\pm0.22$ \\
$(-10,+0)$ & $0.19\pm0.02$ & $0.15\pm0.02$ & $\mathbf{0.13\pm0.02}$ & $1.92\pm0.16$ \\
$(-10,+20)$ & $\mathbf{0.14\pm0.01}$ & $0.32\pm0.03$ & $0.29\pm0.03$ & $2.26\pm0.23$ \\
$(-10,+40)$ & $\mathbf{0.18\pm0.02}$ & $0.50\pm0.01$ & $0.54\pm0.07$ & $1.66\pm0.40$ \\
$(+0,-40)$ & $\mathbf{0.18\pm0.01}$ & $0.42\pm0.01$ & $0.34\pm0.01$ & $2.96\pm0.05$ \\
$(+0,-20)$ & $\mathbf{0.04\pm0.01}$ & $0.22\pm0.02$ & $0.15\pm0.02$ & $2.65\pm0.09$ \\
$(+0,+0)$ & $\mathbf{0.04\pm0.02}$ & $0.09\pm0.02$ & $0.07\pm0.01$ & $2.04\pm0.10$ \\
$(+0,+20)$ & $0.09\pm0.02$ & $\mathbf{0.07\pm0.01}$ & $0.09\pm0.04$ & $1.73\pm0.06$ \\
$(+0,+40)$ & $0.15\pm0.03$ & $0.17\pm0.03$ & $\mathbf{0.11\pm0.03}$ & $1.61\pm0.09$ \\
$(+10,-40)$ & $\mathbf{0.09\pm0.01}$ & $0.19\pm0.01$ & $0.13\pm0.01$ & $0.43\pm0.16$ \\
$(+10,-20)$ & $\mathbf{0.02\pm0.00}$ & $0.13\pm0.02$ & $0.06\pm0.01$ & $0.82\pm0.05$ \\
$(+10,+0)$ & $0.12\pm0.01$ & $0.13\pm0.02$ & $\mathbf{0.11\pm0.03}$ & $1.45\pm0.06$ \\
$(+10,+20)$ & $0.16\pm0.03$ & $\mathbf{0.04\pm0.01}$ & $0.06\pm0.02$ & $1.62\pm0.09$ \\
$(+10,+40)$ & $\mathbf{0.15\pm0.03}$ & $0.16\pm0.03$ & $0.19\pm0.06$ & $2.10\pm0.11$ \\
$(+20,-40)$ & $\mathbf{0.08\pm0.01}$ & $0.14\pm0.01$ & $0.16\pm0.01$ & $0.99\pm0.13$ \\
$(+20,-20)$ & $\mathbf{0.02\pm0.01}$ & $0.18\pm0.01$ & $0.15\pm0.02$ & $0.28\pm0.09$ \\
$(+20,+0)$ & $0.07\pm0.01$ & $0.11\pm0.01$ & $\mathbf{0.05\pm0.03}$ & $0.16\pm0.10$ \\
$(+20,+20)$ & $\mathbf{0.11\pm0.01}$ & $0.14\pm0.03$ & $0.25\pm0.03$ & $0.26\pm0.23$ \\
$(+20,+40)$ & $0.29\pm0.03$ & $\mathbf{0.04\pm0.03}$ & $0.10\pm0.05$ & $0.58\pm0.29$ \\
\bottomrule
\end{tabular}
\end{table}

\section*{S3. Data sources and reproduction of the quantitative summary}
This section records the datasets, alignment conventions, and analysis steps that reproduce the fixed-camera and closed-loop entries of the main-text summary table (main-text Table~2) and the cell-wise summaries of Section~S2.3. The fixed-camera results are errors relative to robot-commanded pose changes after a same-dataset constant alignment. They are not measurements against an independently calibrated ground truth. The conventions below are identical to those of main-text Section~4.2.1.

\subsection*{S3.1. Canonical experimental datasets}
The fixed-camera and closed-loop values reported in the main-text summary table and in the supplementary analyses (e.g., Figs.~\ref{fig:S8}--\ref{fig:S10} and Tables~\ref{tab:cellwise_ori}--\ref{tab:cellwise_pos}) are derived from the following datasets (CSV files unless noted), which are released on GitHub together with the marker design and detection code (the repository is given in the Data and Code Availability statement of the main text):
\begin{itemize}
\item \texttt{tilt\_narrow\_5deg/250414\_3\_tilt\_results\_\textless marker\textgreater.csv}: orientation experiment under narrow-angle conditions ($-10\tcdegree$ to $+10\tcdegree$ in $5\tcdegree$ increments, up to 10 consecutive frames per $(\theta_x,\theta_y)$ cell).
\item \texttt{tilt\_broad\_15deg/250414\_2\_tilt\_results\_\textless marker\textgreater.csv}: orientation experiment under broad-angle conditions ($-30\tcdegree$ to $+30\tcdegree$ in $15\tcdegree$ increments, up to 10 consecutive frames per cell).
\item \texttt{coord\_position/250403\_\allowbreak{}coord\_\allowbreak{}results\_\allowbreak{}\textless marker\textgreater.csv}:\newline position experiment ($\pm 20$~mm horizontal, $\pm 40$~mm depth). The commanded displacements are stored in meters in the columns \texttt{dx} (horizontal) and \texttt{dy} (depth), which correspond to $\Delta x$ and $\Delta z$ of Table~\ref{tab:cellwise_pos}.
\item \texttt{hand\_eye\_recordings/260910\_\allowbreak{}\textless marker\textgreater\_\allowbreak{}robot\_\allowbreak{}marker.csv}:\newline the ten-pose robot-kinematics and marker-observation recordings used for the per-marker-type hand--eye estimation (Section~S4), acquired on the day of the pose-hold runs. These files are whitespace-delimited without a header, one row per pose: the end-effector translation (m) and rotation vector (rad) reported by the controller, followed by the marker translation and rotation vector observed in the camera frame in the same units.
\item \texttt{hand\_eye\_transforms/260910\_\allowbreak{}\textless marker\textgreater\_\allowbreak{}ee\_cam\_X.txt}:\newline the estimated hand--eye transforms of Table~\ref{tab:S7}.
\item \texttt{closed\_loop\_pose\_hold\_5runs/260910\_\allowbreak{}\textless marker\textgreater\_\allowbreak{}live\_\allowbreak{}tracking\_\allowbreak{}log\_\allowbreak{}run\textless k\textgreater.csv}:\newline the 20 closed-loop pose-hold logs (five runs per marker type, Section~S5): elapsed time, camera-to-marker position and orientation error relative to the locked target, and the end-effector and marker poses at each control step. \texttt{closed\_loop\_pose\_hold\_5runs/260910\_\allowbreak{}closedloop\_\allowbreak{}pose\_\allowbreak{}hold\_\allowbreak{}summary.csv} lists the per-run values of Table~\ref{tab:closedloop_runs}. In the logs, poses are in meters and radians and errors in millimeters and degrees, as named in the header row.
\item \texttt{260619\_swingdown\_pertrial.csv},\newline \texttt{260619\_swingdown\_summary.csv}: per-swing and per-marker-type in-grasp slip of the swing-down grasp-stability test (main-text swing-down table, five swings per marker type), extracted from the motion-capture recordings as described in Section~S7. The \texttt{layout} column uses the written-out names Diagonal, Pitch, Reversed-pitch, and Single planar tag, and the \texttt{file} column names the original motion-capture takes, which are not part of the release.
\end{itemize}
Here \texttt{\textless marker\textgreater} takes one of \texttt{\{single, quartet\_\allowbreak{}diagonal, quartet\_\allowbreak{}pitch, quartet\_\allowbreak{}pitch\_\allowbreak{}r\}}, and \texttt{\textless k\textgreater} is the run number 1--5 used in Table~\ref{tab:closedloop_runs}. The per-cell sample counts are nominal: occasional frames without a valid detection were dropped (at most two per dataset), so a few commanded cells contain nine valid samples. The analysis averages the valid samples of each cell, and every frontal orientation cell entering the main-text summary contains the full 10 samples. In the position datasets the only frontal cell with nine valid frames is that of quARtet-rP, which enters only the grid-wide RMSE.

\subsection*{S3.2. Unit conventions}
All translation columns of the released CSVs are in meters and are converted to millimeters before the position RMSE entry of the main-text summary table is computed (Section~S3.3). The translation columns of the orientation-experiment CSVs are not used in the orientation analysis. Orientation CSVs report the measured rotation as Rodrigues vectors in radians, and orientation error is computed from the geodesic angular distance between the measured and reference rotations.

\subsection*{S3.3. Analysis steps}
The fixed-camera entries of the main-text summary table, namely the frontal-cell ($\theta_x=\theta_y=0$) mean$\pm$SD orientation errors of the broad grid and the position RMSEs, are obtained from the CSV files listed above by the following steps:
\begin{itemize}
\item the evaluation of the orientation error as the $\SO$ geodesic distance between the measured and the robot-commanded relative orientation, computed from a fixed ideal reference orientation after removing a single constant camera-to-command frame offset estimated robustly across all poses: the offset is the rotation minimizing the mean $\SO$ log-residual (the residual rotation angle) over all samples, refitted with samples whose residual exceeds $12\tcdegree$ (pose flips) excluded from the fit (at most three passes) while all samples are retained in the reported error.
\item the reduction of the position measurements to a common millimeter scale (Section~S3.2), followed by a global axis-sign resolution (sign of the command--measurement correlation per axis) and the removal of a single constant in-plane frame offset, fitted as a two-dimensional rigid Procrustes (Kabsch) alignment of all measured positions onto the commanded grid. The fit is made robust by excluding samples whose residual exceeds three times the current root-mean-square residual and refitting (at most three passes), while all samples are retained in the reported error, taken as the Euclidean residual to the commanded relative translation.
\end{itemize}
Sample standard deviation ($\mathrm{ddof}=1$) is used throughout. The cell-wise tables and difference maps of Section~S2.3 apply the same conventions to every commanded cell. The closed-loop entries of the main-text summary table and Table~\ref{tab:closedloop_runs} are the within-run means over the common window defined in Section~S5, which also specifies the representative-run rule and the statistical tests.

The alignment parameters are fitted to the same samples on which the errors are evaluated. They therefore remove the global frame mismatch together with any constant marker-specific bias. The resulting values do not measure absolute pose bias or held-out calibration accuracy. Because one constant transform is shared by all commanded cells, pose-dependent patterns across the grid remain in the errors. The 10 frames at each cell are consecutive images of one held pose and quantify within-acquisition variation, not variability across independent re-mountings. No frame-level significance inference is used for the main fixed-camera claims.

For reference, two auxiliary quantities quoted in Section~4.2 of the main text are obtained from the same position datasets: the mean within-cell standard deviation of the measured positions is $0.02$--$0.03$~mm for the three quARtet layouts and $0.15$~mm for Single, and the fitted constant in-plane frame offset includes an in-plane rotation of approximately $-3.1\tcdegree$ in all four datasets. This agreement is compatible with a common camera-to-robot frame mismatch but is not an independent validation, because each offset is fitted to its evaluated dataset. The manufacturer's $\pm0.03$~mm repeatability specification is translational and comparable to the smallest within-cell variations. No external orientation-reference uncertainty was measured.

\subsection*{S3.4. Detector-side 3D corner coordinates}
Table~\ref{tab:corners} lists the 3D corner coordinates that the detector reconstructs from the shared configuration for the implemented layouts, so that the marker geometry is fully specified independently of the code. The construction follows main-text Section~3.3: the local tag square of side $\texttt{tag\_size\_mm}$ is tilted about the layout-specific axis, rotated in-plane by $90\tcdegree t$ for tag $t$, and translated to quadrant $t$ of the $\texttt{whole\_size\_mm}$ footprint. The reversed-pitch layout additionally applies the in-plane block rotation described in main-text Section~3.2. In this frame the tilted tag faces rise toward $+z$, which points from the mounting face toward the tag side and therefore toward a camera that views the marker. The released pose logs use the same frame. The add-in and the detector define the tilt with opposite signs, so the detector reconstructs a marker generated with $\texttt{tilt\_degree}=18$ from a tilt argument of $-18$, which is the negation applied by the released wrapper and by the experiments of this study. Relative to the pitch layout, the in-plane block rotation of the reversed-pitch layout moves the projected footprint of each tag by $0.78$~mm toward the footprint center ($17.50$ versus $16.72$~mm in Table~\ref{tab:corners}), which equals $\texttt{tag\_size\_mm}\,(1-\cos\texttt{tilt\_degree})$.

\begin{table}[!htbp]
\centering\scriptsize
\setlength{\tabcolsep}{3.6pt}
\caption{Detector-side 3D corner coordinates (mm) of the three implemented layouts for the configuration used in this study ($\texttt{whole\_size\_mm}=35$, $\texttt{tag\_size\_mm}=16$, $\texttt{tilt\_degree}=18$), generated by the released detector code from the shared configuration (with the detector's tilt argument set to $-\texttt{tilt\_degree}$, see the text). Coordinates are in the marker frame of Section~3.3 (origin at the footprint center on the base plane, $z$ normal to it and pointing toward the tag side). Tag $t\in\{0,1,2,3\}$ is the AprilTag with ID $t$; corners $c_1$--$c_4$ follow the AprilTag corner convention of the local tag square before tilting. These 16 points per layout are the object points of the joint PnP solve.}
\label{tab:corners}
\begin{tabular}{@{}cc rrr rrr rrr@{}}
\toprule
 & & \multicolumn{3}{c}{quARtet-D} & \multicolumn{3}{c}{quARtet-P} & \multicolumn{3}{c}{quARtet-rP} \\
\cmidrule(lr){3-5}\cmidrule(lr){6-8}\cmidrule(lr){9-11}
Tag & Corner & $x$ & $y$ & $z$ & $x$ & $y$ & $z$ & $x$ & $y$ & $z$ \\
\midrule
0 & $c_1$ & -17.50 & -17.50 & 0.00 & -17.50 & -17.50 & 0.00 & -1.50 & -1.50 & 0.00 \\
 & $c_2$ & -1.89 & -17.89 & 3.50 & -1.50 & -17.50 & 0.00 & -17.50 & -1.50 & 0.00 \\
 & $c_3$ & -2.28 & -2.28 & 6.99 & -1.50 & -2.28 & 4.94 & -17.50 & -16.72 & 4.94 \\
 & $c_4$ & -17.89 & -1.89 & 3.50 & -17.50 & -2.28 & 4.94 & -1.50 & -16.72 & 4.94 \\
\addlinespace[2pt]
1 & $c_1$ & 17.50 & -17.50 & 0.00 & 17.50 & -17.50 & 0.00 & 1.50 & -1.50 & 0.00 \\
 & $c_2$ & 17.89 & -1.89 & 3.50 & 17.50 & -1.50 & 0.00 & 1.50 & -17.50 & 0.00 \\
 & $c_3$ & 2.28 & -2.28 & 6.99 & 2.28 & -1.50 & 4.94 & 16.72 & -17.50 & 4.94 \\
 & $c_4$ & 1.89 & -17.89 & 3.50 & 2.28 & -17.50 & 4.94 & 16.72 & -1.50 & 4.94 \\
\addlinespace[2pt]
2 & $c_1$ & 17.50 & 17.50 & 0.00 & 17.50 & 17.50 & 0.00 & 1.50 & 1.50 & 0.00 \\
 & $c_2$ & 1.89 & 17.89 & 3.50 & 1.50 & 17.50 & 0.00 & 17.50 & 1.50 & 0.00 \\
 & $c_3$ & 2.28 & 2.28 & 6.99 & 1.50 & 2.28 & 4.94 & 17.50 & 16.72 & 4.94 \\
 & $c_4$ & 17.89 & 1.89 & 3.50 & 17.50 & 2.28 & 4.94 & 1.50 & 16.72 & 4.94 \\
\addlinespace[2pt]
3 & $c_1$ & -17.50 & 17.50 & 0.00 & -17.50 & 17.50 & 0.00 & -1.50 & 1.50 & 0.00 \\
 & $c_2$ & -17.89 & 1.89 & 3.50 & -17.50 & 1.50 & 0.00 & -1.50 & 17.50 & 0.00 \\
 & $c_3$ & -2.28 & 2.28 & 6.99 & -2.28 & 1.50 & 4.94 & -16.72 & 17.50 & 4.94 \\
 & $c_4$ & -1.89 & 17.89 & 3.50 & -2.28 & 17.50 & 4.94 & -16.72 & 1.50 & 4.94 \\
\bottomrule
\end{tabular}
\end{table}

\section*{S4. Hand--eye transform estimation details for the robot-mounted-camera configuration}
This section provides details for the robot-mounted-camera configuration used in the closed-loop pose-hold experiment of main-text Section~4.2. The goal is to determine the constant rigid transformation between the robot end-effector frame and the camera frame.

\subsection*{S4.1. Coordinate frames and SE(3) formulation}
Rigid motions are represented by $4\times4$ homogeneous matrices in $\mathrm{SE}(3)$. The notation $\T{A}{B}$ maps coordinates from frame $B$ to frame $A$. For pose $i$, $\T{\rm base}{\rm ee,i}$ is obtained from the robot controller, $\T{\rm ee}{\rm cam}$ is the unknown hand--eye transform, and $\T{\rm cam}{\rm marker,i}$ is the marker pose estimated from the image. Because the marker is fixed,
\begin{equation}\T{\rm base}{\rm marker,i}=\T{\rm base}{\rm ee,i}\T{\rm ee}{\rm cam}\T{\rm cam}{\rm marker,i}={\rm constant}.\end{equation}
Each transform has the standard form
\begin{equation}T=\begin{bmatrix}R&\bm t\\\bm0^\mathsf{T}&1\end{bmatrix}.\end{equation}

\subsection*{S4.2. Optimization-based estimation}
For a candidate hand--eye transform $X$, the implied base-to-marker transforms are
\begin{equation}M_i(X)=\T{\rm base}{\rm ee,i}X\T{\rm cam}{\rm marker,i},\quad i=1,\ldots,N.\end{equation}
The objective minimizes their dispersion around the sample mean:
\begin{equation}\bar M(X)=\frac{1}{N}\sum_{i=1}^{N}M_i(X),\qquad J(X)=\sum_{i=1}^{N}\|M_i(X)-\bar M(X)\|_F^2.\end{equation}
The transform was parameterized by three translations and a rotation vector. The optimization was initialized from $(0,0,0.05,0,0,0)$, with translation bounds $[-0.3,0.3]$~m and rotation-vector bounds $[-\pi,\pi]$~rad.

\subsection*{S4.3. Experimental estimation setup and poses}
The marker was fixed to a table while the camera was mounted to the robot end-effector using a 3D-printed fixture (main-text Fig.~5). The nominal initial pose was near-frontal to the marker, at a camera-to-marker distance of about $200$~mm, as recorded in the hand--eye and pose-hold logs. Data were collected by moving the robot from this initial pose to the displaced poses listed in Table~\ref{tab:S1}. In that table the translational entries are absolute displacements from the initial pose, whereas the rotational entries are increments added to the components of the initial pose's rotation vector. Because the initial pose is far from the identity orientation, these increments merely generate a diverse set of orientations and are not exact axis-angle rotations from the initial pose. This parameterization does not affect the estimate: the optimization uses the \emph{actual} end-effector poses reported by the robot controller (not the nominal commands), so only the realized pose diversity matters. The hand--eye transform was estimated separately for each marker type. The resulting numerical estimates are reported in Table~\ref{tab:S7}. All four fitted transforms are close to the same nominal rigid offset, essentially a ${\sim}40$~mm translation along the camera optical axis with rotations of at most about $4\tcdegree$.
\small
\setlength{\tabcolsep}{4pt}
\begin{longtable}{ccccccc}
\caption{Robot pose displacements used for hand--eye estimation in the robot-mounted-camera configuration (Section~S4). The robot was moved from a near-frontal initial pose to the displaced poses listed in the table. $\Delta x$, $\Delta y$, and $\Delta z$ are translational displacements from the initial pose in meters, and $\Delta r_x$, $\Delta r_y$, and $\Delta r_z$ are increments added to the components of the initial pose's rotation vector in degrees. At each pose, the robot kinematics and the marker observation were recorded and used for hand--eye transform estimation. Entry~1 is the undisplaced initial pose.}\label{tab:S1}\\
\toprule
Entry & $\Delta x$ [m] & $\Delta y$ [m] & $\Delta z$ [m] & $\Delta r_x$ [$\tcdegree$] & $\Delta r_y$ [$\tcdegree$] & $\Delta r_z$ [$\tcdegree$]\\
\midrule
\endfirsthead
\caption[]{Robot pose displacements used for hand--eye estimation in the robot-mounted-camera configuration (Section~S4). The robot was moved from a near-frontal initial pose to the displaced poses listed in the table. $\Delta x$, $\Delta y$, and $\Delta z$ are translational displacements from the initial pose in meters, and $\Delta r_x$, $\Delta r_y$, and $\Delta r_z$ are increments added to the components of the initial pose's rotation vector in degrees. At each pose, the robot kinematics and the marker observation were recorded and used for hand--eye transform estimation. (continued)}\\
\toprule
Entry & $\Delta x$ [m] & $\Delta y$ [m] & $\Delta z$ [m] & $\Delta r_x$ [$\tcdegree$] & $\Delta r_y$ [$\tcdegree$] & $\Delta r_z$ [$\tcdegree$]\\
\midrule
\endhead
\midrule
\multicolumn{7}{r}{Continued on next page}\\
\endfoot
\bottomrule
\endlastfoot
1 & 0.00 & 0.00 & 0.00 & 0.0 & 0.0 & 0.0\\
2 & 0.03 & 0.00 & 0.00 & 0.0 & 10.0 & 0.0\\
3 & -0.03 & 0.00 & 0.00 & 0.0 & -10.0 & 0.0\\
4 & 0.00 & 0.03 & 0.00 & 10.0 & 0.0 & 0.0\\
5 & 0.00 & -0.03 & 0.00 & -10.0 & 0.0 & 0.0\\
6 & 0.00 & 0.00 & 0.03 & 0.0 & 0.0 & 10.0\\
7 & 0.00 & 0.00 & -0.03 & 0.0 & 0.0 & -10.0\\
8 & 0.02 & 0.02 & 0.00 & 5.0 & 5.0 & 0.0\\
9 & -0.02 & -0.02 & 0.00 & -5.0 & 0.0 & 5.0\\
10 & 0.02 & -0.02 & 0.02 & 10.0 & -5.0 & 5.0\\
\end{longtable}
\normalsize

\subsection*{S4.4. Pose update toward the locked target}
Let $\T{\rm base}{\rm ee,c}$ be the current end-effector pose, $\T{\rm ee}{\rm cam}$ the hand--eye transform, $\T{\rm cam}{\rm marker,c}$ the current marker observation, and $\T{\rm cam}{\rm marker,d}$ the desired marker pose, which in the pose-hold experiment is the initial near-frontal observation locked as the target (Section~S5) and is held fixed throughout the run. Since the marker is fixed,
\begin{equation}\T{\rm base}{\rm marker}=\T{\rm base}{\rm ee,c}\T{\rm ee}{\rm cam}\T{\rm cam}{\rm marker,c}=\T{\rm base}{\rm ee,n}\T{\rm ee}{\rm cam}\T{\rm cam}{\rm marker,d},\end{equation}
and therefore
\begin{equation}\T{\rm base}{\rm ee,n}=\T{\rm base}{\rm marker}\left(\T{\rm ee}{\rm cam}\T{\rm cam}{\rm marker,d}\right)^{-1}.\end{equation}

\subsection*{S4.5. Estimated hand--eye transforms}
Table~\ref{tab:S7} lists the estimated hand--eye transforms $\T{\rm ee}{\rm cam}$ obtained from the optimization of Section~S4.2 (using the data-collection protocol of Section~S4.3) for each of the four marker types.
\small
\setlength{\tabcolsep}{4pt}
\begin{longtable}{cc}
\caption{Estimated hand--eye transforms $\T{\rm ee}{\rm cam}$ for the robot-mounted-camera configuration, expressed as $4\times4$ homogeneous matrices. Translations are in meters. A separate estimate was performed for each marker type on the day of the pose-hold runs, because the marker observations enter the optimization objective described in Section~S4.}\label{tab:S7}\\
\toprule
Marker & Estimated $\T{\rm ee}{\rm cam}$ \\
\midrule
\endfirsthead
\caption[]{Estimated hand--eye transforms (continued)}\\
\toprule
Marker & Estimated $\T{\rm ee}{\rm cam}$ \\
\midrule
\endhead
\midrule
\multicolumn{2}{r}{Continued on next page}\\
\endfoot
\bottomrule
\endlastfoot
quARtet-D &
$\begin{bmatrix}
 0.99997 &  0.00750 & -0.00258 &  0.00140\\
-0.00752 &  0.99996 & -0.00424 & -0.00257\\
 0.00255 &  0.00426 &  0.99999 &  0.03923\\
 0       &  0       &  0       &  1
\end{bmatrix}$ \\[1.5em]
quARtet-P &
$\begin{bmatrix}
 0.99961 &  0.02491 &  0.01251 & -0.00165\\
-0.02477 &  0.99963 & -0.01138 & -0.00015\\
-0.01279 &  0.01107 &  0.99986 &  0.03974\\
 0       &  0       &  0       &  1
\end{bmatrix}$ \\[1.5em]
quARtet-rP &
$\begin{bmatrix}
 0.99953 & -0.00056 & -0.03073 &  0.00464\\
-0.00010 &  0.99977 & -0.02133 & -0.00151\\
 0.03074 &  0.02132 &  0.99930 &  0.03711\\
 0       &  0       &  0       &  1
\end{bmatrix}$ \\[1.5em]
Single &
$\begin{bmatrix}
 0.99758 &  0.05210 & -0.04613 &  0.00446\\
-0.05299 &  0.99843 & -0.01817 &  0.00449\\
 0.04511 &  0.02057 &  0.99877 &  0.03957\\
 0       &  0       &  0       &  1
\end{bmatrix}$ \\
\end{longtable}
\normalsize

\section*{S5. Closed-loop pose-hold protocol}
Main-text Section~4.2.2 presents the closed-loop pose-hold traces and their interpretation. This section records the corresponding protocol without duplicating the figure.

The robot-mounted camera and fixed marker of Section~S4 (main-text Fig.~5) were used. An initial near-frontal marker view was locked as the target. The end-effector was displaced to $(\theta_x,\theta_y,\theta_z)=(10,-8,6)\tcdegree$, corresponding to a $12.8\tcdegree$ viewing-axis tilt plus a $6\tcdegree$ rotation about the viewing axis, and was commanded to restore the target continuously for 120~s. At each control step, the camera-to-marker orientation error relative to the target and the geodesic reorientation between consecutive end-effector orientations were logged. Five runs were acquired for each of quARtet-D, quARtet-P, quARtet-rP, and Single, in that order, within one session with the camera fixture unchanged. Between runs the marker was re-placed by hand, approximately centered and near-frontal in the camera view, and before every run the target view was locked afresh from the current observation and the displacement was applied again, so each run is an independent restart of the closed loop from a re-placed marker. All quantities are evaluated over the ${\sim}117$~s window shared by the 20 logs, i.e., the duration of the shortest run. Main-text Fig.~8 shows, for each marker type, the run whose within-run mean hold error is closest to the median of its five runs.

\begin{figure}[!htbp]
\centering
\begin{minipage}[b]{0.48\linewidth}\centering\includegraphics[width=\linewidth]{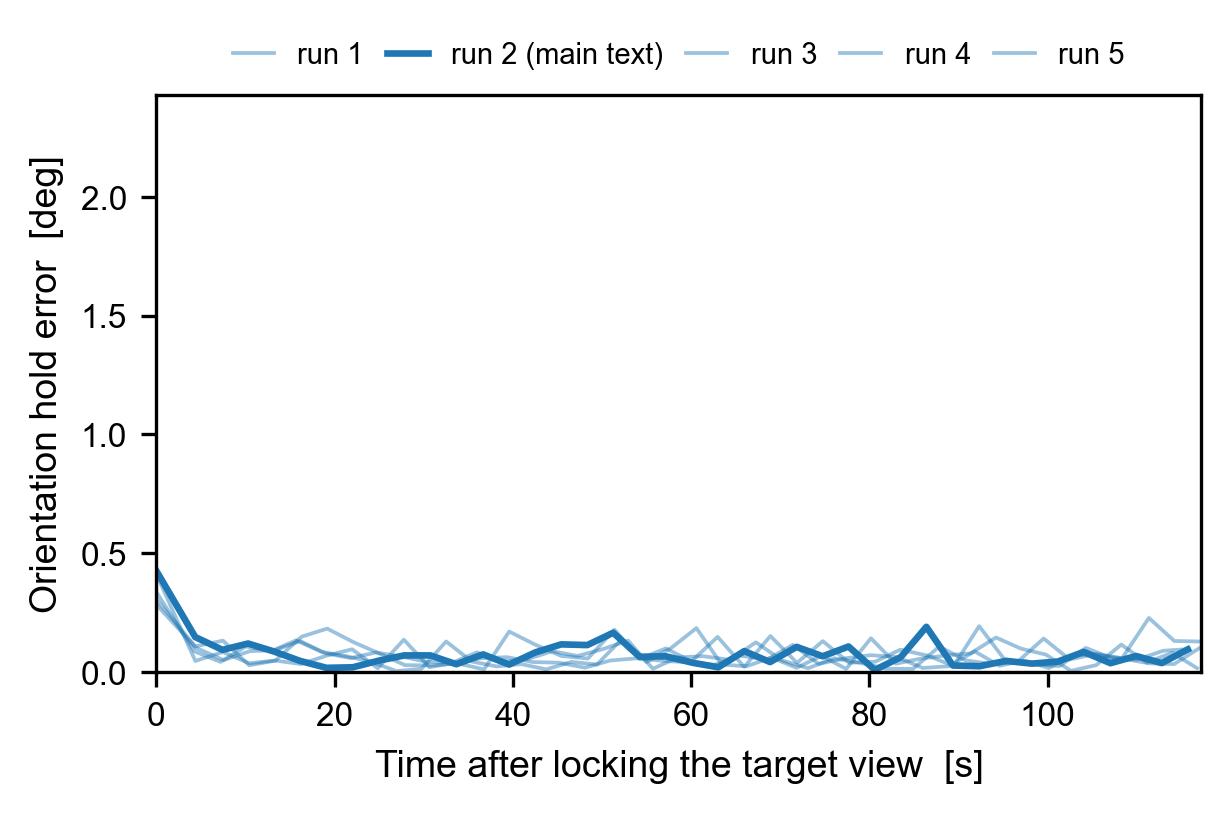}\\{\footnotesize (a) quARtet-D}\end{minipage}\hfill
\begin{minipage}[b]{0.48\linewidth}\centering\includegraphics[width=\linewidth]{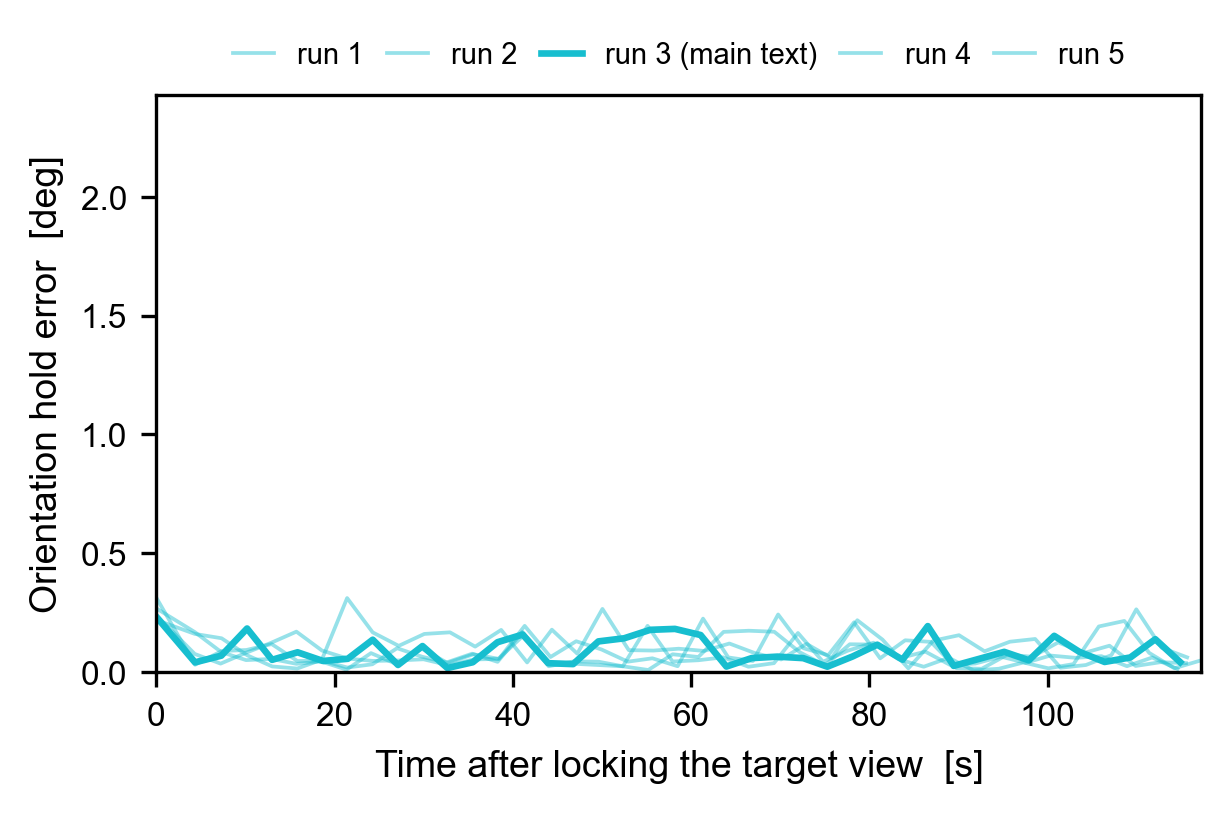}\\{\footnotesize (b) quARtet-P}\end{minipage}\par\vspace{1ex}
\begin{minipage}[b]{0.48\linewidth}\centering\includegraphics[width=\linewidth]{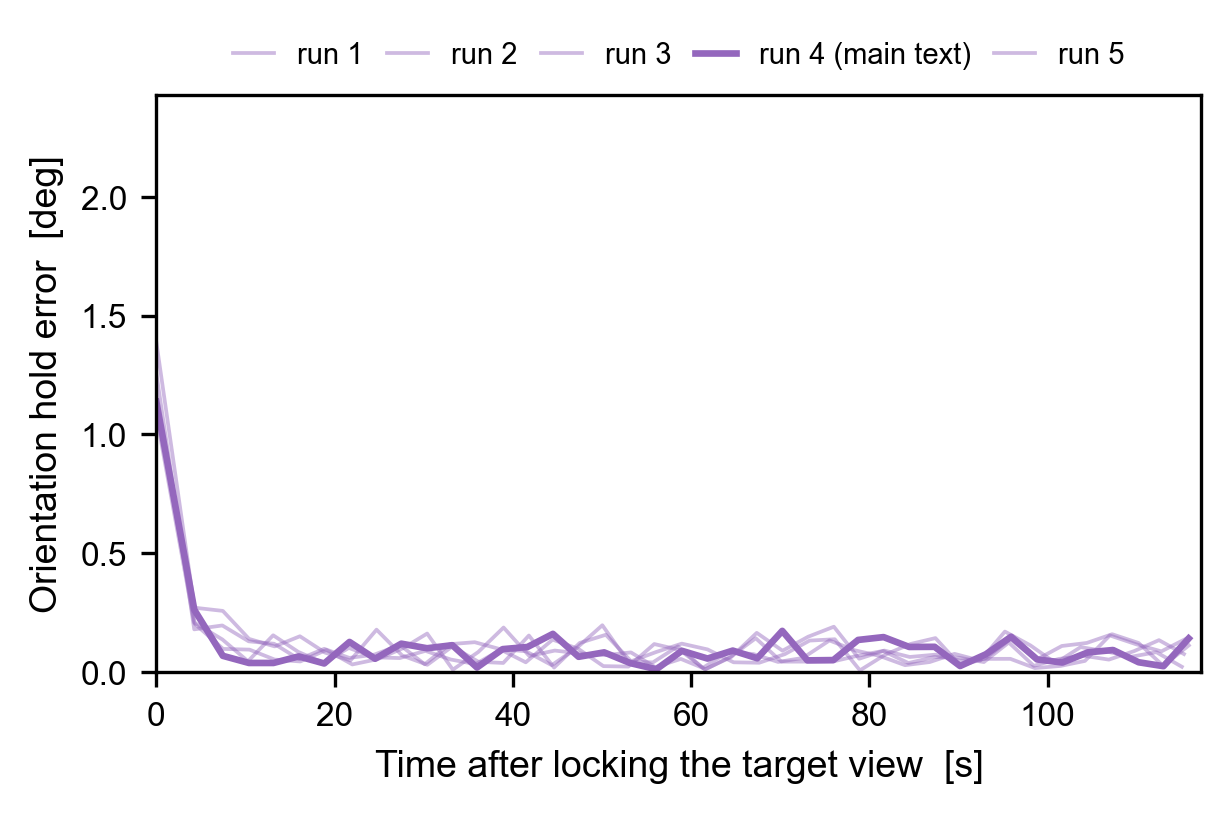}\\{\footnotesize (c) quARtet-rP}\end{minipage}\hfill
\begin{minipage}[b]{0.48\linewidth}\centering\includegraphics[width=\linewidth]{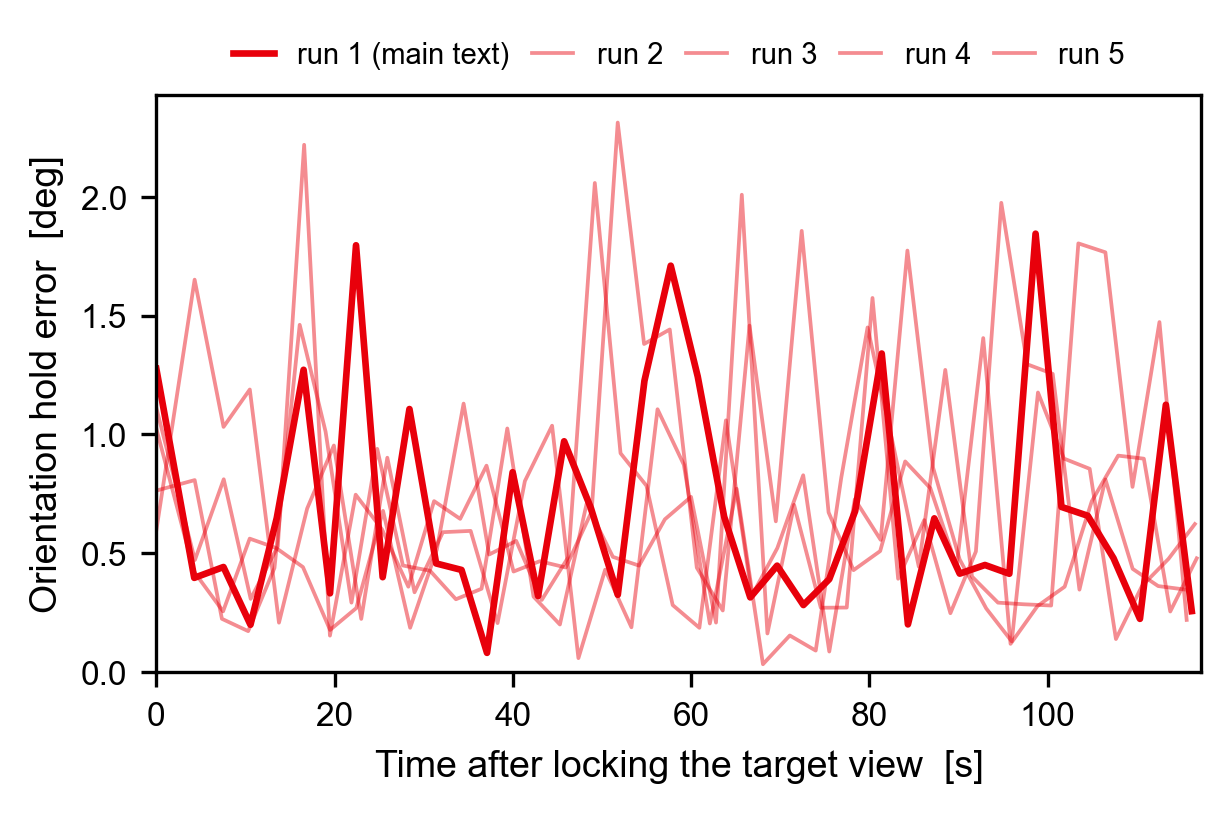}\\{\footnotesize (d) Single}\end{minipage}
\caption{Orientation hold error of all five closed-loop runs per marker type over the common ${\sim}117$~s window, in the order quARtet-D, quARtet-P, quARtet-rP, and Single, with shared axis limits. The bold trace is the representative run shown in the main text.}\label{fig:closedloop_all_hold}
\end{figure}

\begin{figure}[!htbp]
\centering
\begin{minipage}[b]{0.48\linewidth}\centering\includegraphics[width=\linewidth]{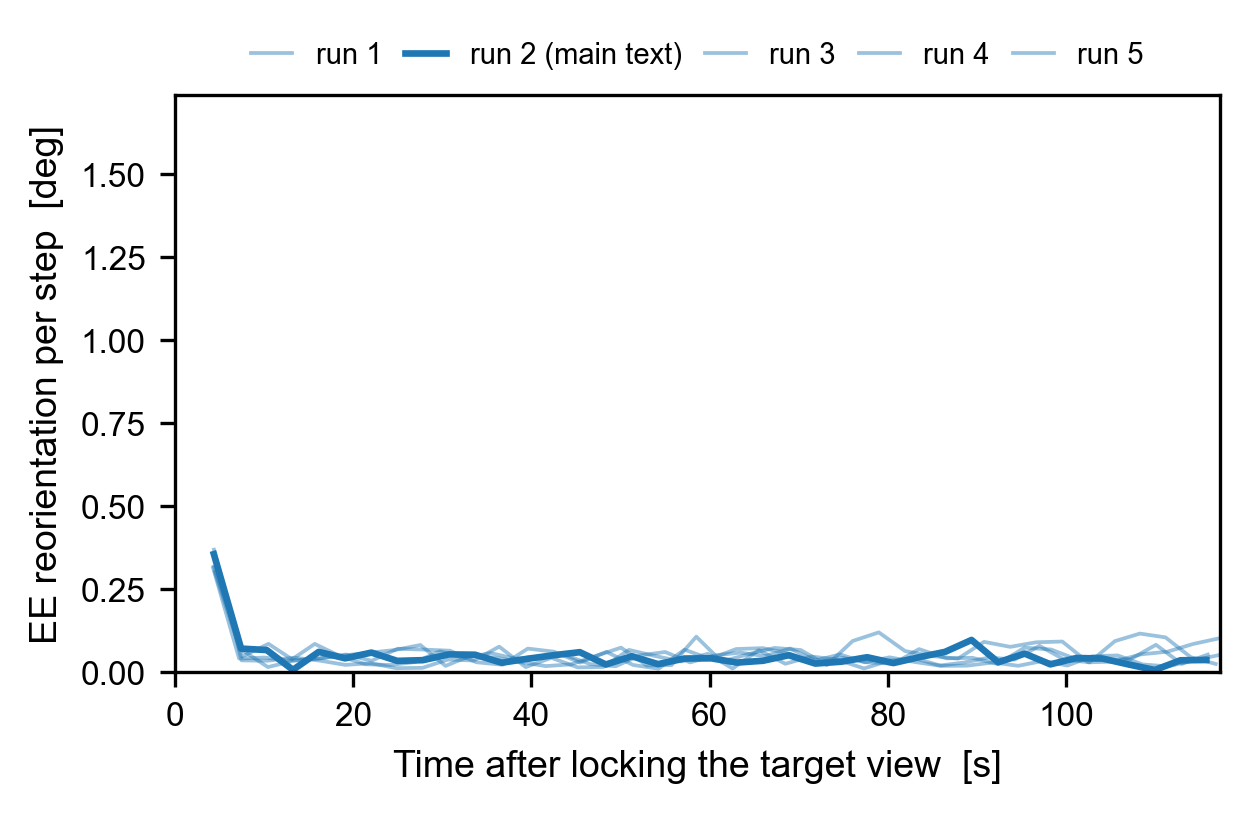}\\{\footnotesize (a) quARtet-D}\end{minipage}\hfill
\begin{minipage}[b]{0.48\linewidth}\centering\includegraphics[width=\linewidth]{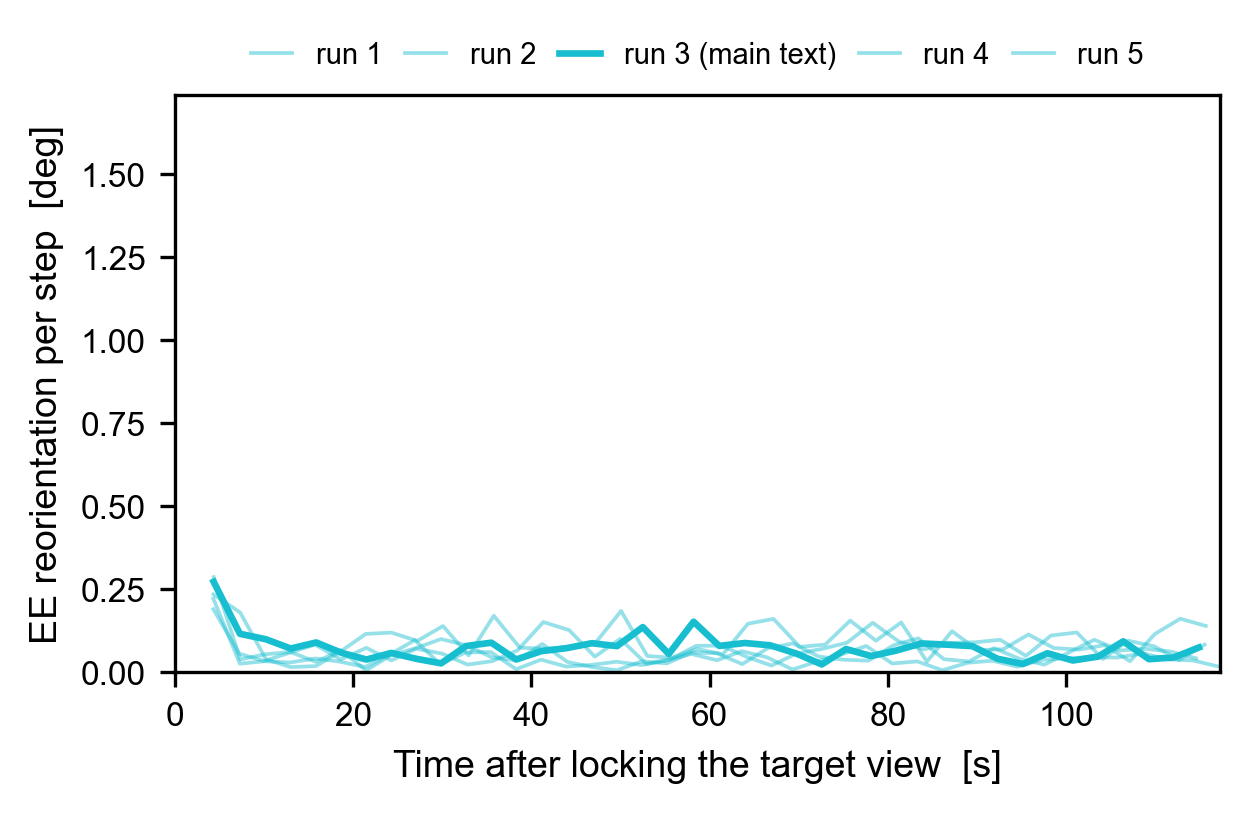}\\{\footnotesize (b) quARtet-P}\end{minipage}\par\vspace{1ex}
\begin{minipage}[b]{0.48\linewidth}\centering\includegraphics[width=\linewidth]{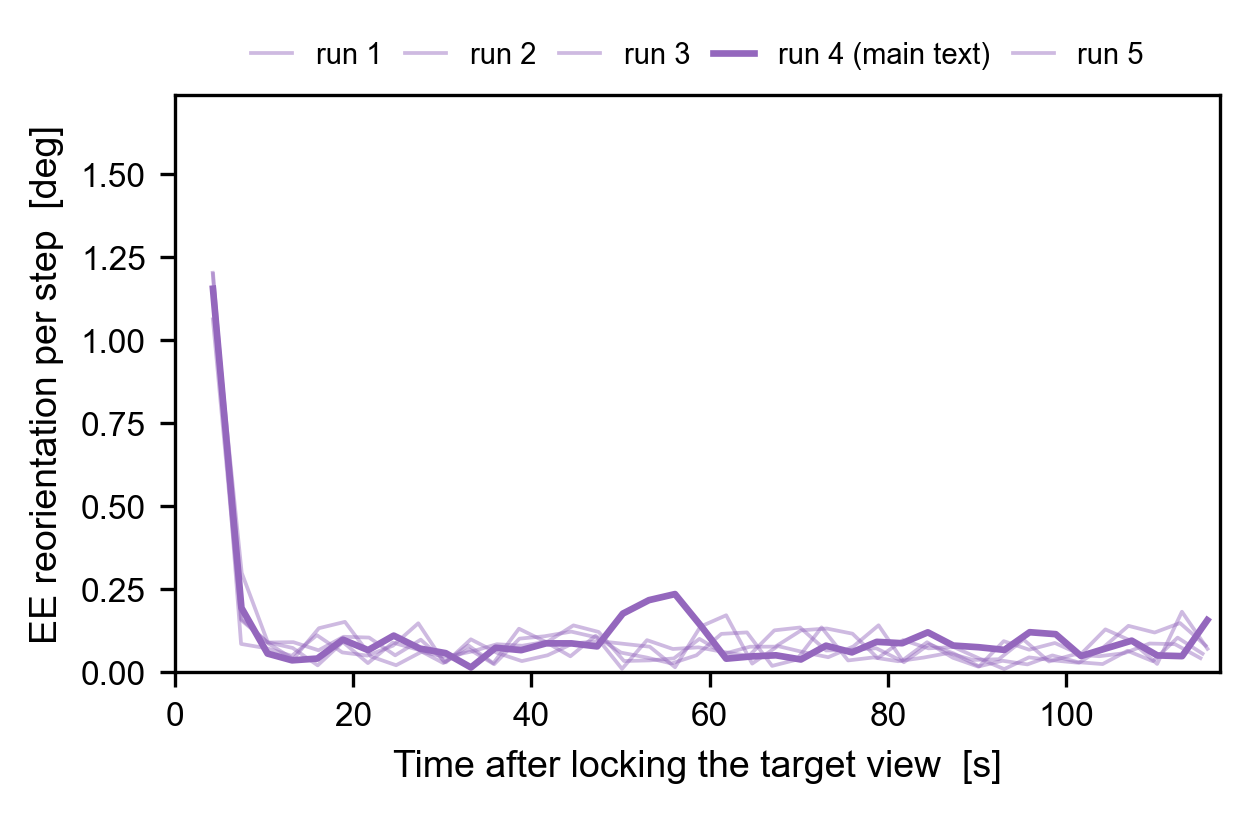}\\{\footnotesize (c) quARtet-rP}\end{minipage}\hfill
\begin{minipage}[b]{0.48\linewidth}\centering\includegraphics[width=\linewidth]{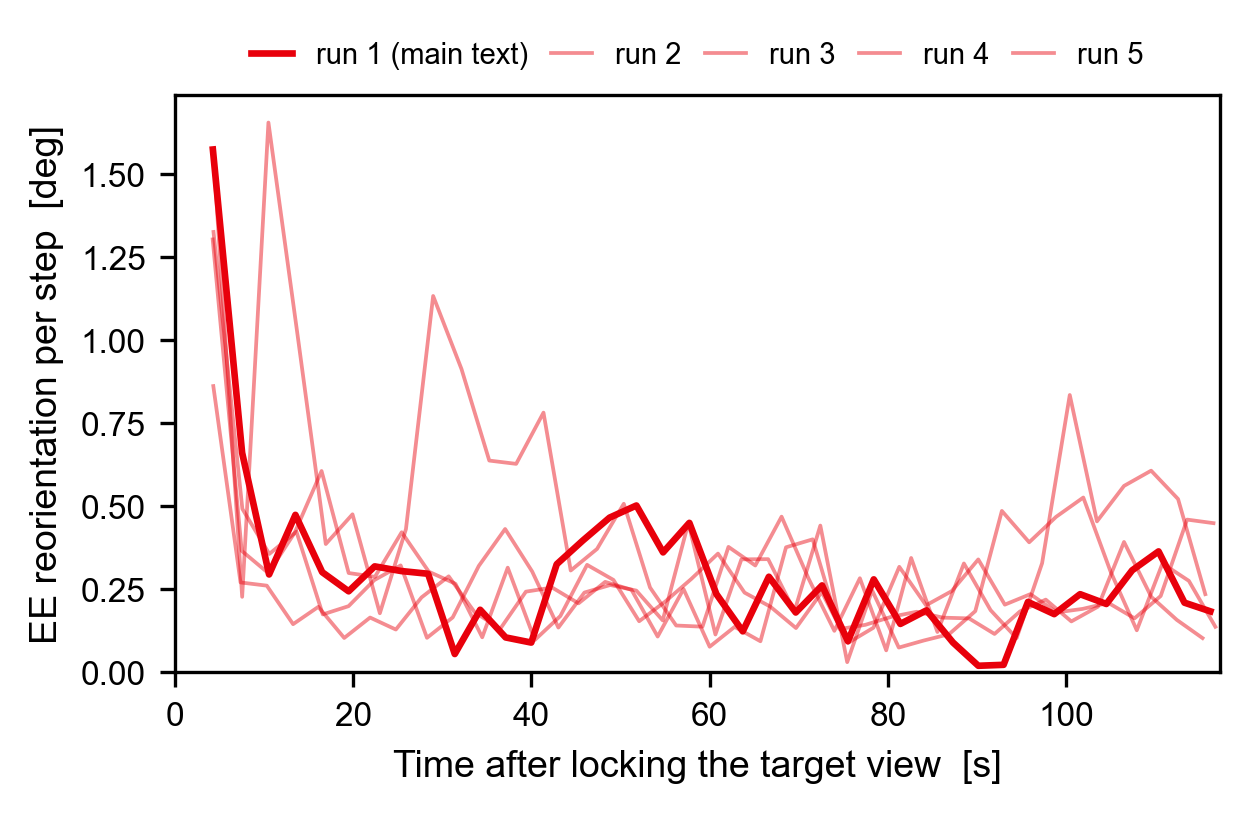}\\{\footnotesize (d) Single}\end{minipage}
\caption{Per-step end-effector reorientation of all five closed-loop runs per marker type over the common ${\sim}117$~s window, in the order quARtet-D, quARtet-P, quARtet-rP, and Single, with shared axis limits. The bold trace is the representative run shown in the main text.}\label{fig:closedloop_all_step}
\end{figure}

Figures~\ref{fig:closedloop_all_hold} and~\ref{fig:closedloop_all_step} overlay all five runs of each marker type, and Table~\ref{tab:closedloop_runs} lists the per-run means and medians. The statistics use the run as the unit of replication: each run contributes its within-run mean, and each quARtet layout is compared with Single by a two-sided Welch $t$-test on the five run means, Holm-corrected within each metric. All three layouts differ from Single in both metrics (adjusted $p<10^{-4}$ for the orientation hold error and adjusted $p=0.012$ for the per-step reorientation). The run-to-run SDs are $0.01$--$0.02\tcdegree$ for the quARtet layouts and $0.06\tcdegree$ for Single in hold error, and $0.01$--$0.02\tcdegree$ and $0.10\tcdegree$ in per-step reorientation. Because all runs were recorded in one session with the camera fixture unchanged, these results characterize repeatability across re-placed markers and re-locked targets, not variability across sessions or camera re-mounting. No settling criterion was pre-specified, so no settling time is claimed.

\begin{table}[!htbp]
\centering\small
\setlength{\tabcolsep}{5pt}
\caption{Per-run summary of the closed-loop pose-hold experiment (five runs per marker type, common ${\sim}117$~s window): within-run mean and median of the orientation hold error and of the per-step end-effector reorientation, in degrees. The run marked $\dagger$ is the representative run plotted in the main text (mean hold error closest to the median of the five runs). Group rows give the mean$\pm$SD of the five run means, the values reported in the closed-loop columns of main-text Table~2.}
\label{tab:closedloop_runs}
\begin{tabular}{@{}llcccc@{}}
\toprule
Marker & Run & \shortstack{Hold error\\mean [$\tcdegree$]} & \shortstack{Hold error\\median [$\tcdegree$]} & \shortstack{Reorientation\\mean [$\tcdegree$]} & \shortstack{Reorientation\\median [$\tcdegree$]} \\
\midrule
quARtet-D & 1 & 0.088 & 0.069 & 0.054 & 0.043 \\
quARtet-D & 2$^{\dagger}$ & 0.079 & 0.067 & 0.050 & 0.041 \\
quARtet-D & 3 & 0.079 & 0.065 & 0.064 & 0.054 \\
quARtet-D & 4 & 0.062 & 0.049 & 0.042 & 0.034 \\
quARtet-D & 5 & 0.072 & 0.059 & 0.054 & 0.044 \\
 & mean$\pm$SD & $0.076\pm0.010$ & & $0.053\pm0.008$ & \\
\addlinespace[2pt]
quARtet-P & 1 & 0.118 & 0.107 & 0.097 & 0.082 \\
quARtet-P & 2 & 0.074 & 0.062 & 0.058 & 0.052 \\
quARtet-P & 3$^{\dagger}$ & 0.089 & 0.065 & 0.073 & 0.070 \\
quARtet-P & 4 & 0.072 & 0.046 & 0.041 & 0.035 \\
quARtet-P & 5 & 0.089 & 0.065 & 0.074 & 0.070 \\
 & mean$\pm$SD & $0.088\pm0.018$ & & $0.069\pm0.021$ & \\
\addlinespace[2pt]
quARtet-rP & 1 & 0.115 & 0.085 & 0.090 & 0.055 \\
quARtet-rP & 2 & 0.111 & 0.079 & 0.110 & 0.077 \\
quARtet-rP & 3 & 0.101 & 0.064 & 0.096 & 0.067 \\
quARtet-rP & 4$^{\dagger}$ & 0.111 & 0.083 & 0.115 & 0.076 \\
quARtet-rP & 5 & 0.132 & 0.108 & 0.108 & 0.081 \\
 & mean$\pm$SD & $0.114\pm0.012$ & & $0.104\pm0.010$ & \\
\addlinespace[2pt]
Single & 1$^{\dagger}$ & 0.692 & 0.468 & 0.288 & 0.244 \\
Single & 2 & 0.762 & 0.640 & 0.297 & 0.275 \\
Single & 3 & 0.610 & 0.519 & 0.290 & 0.244 \\
Single & 4 & 0.651 & 0.448 & 0.213 & 0.182 \\
Single & 5 & 0.697 & 0.597 & 0.481 & 0.374 \\
 & mean$\pm$SD & $0.682\pm0.057$ & & $0.314\pm0.100$ & \\
\bottomrule
\end{tabular}
\end{table}

\section*{S6. Idealized planar-grasp model}
\begin{figure}[!htbp]
\centering
\begin{minipage}[b]{0.62\linewidth}\centering\includegraphics[width=0.95\linewidth]{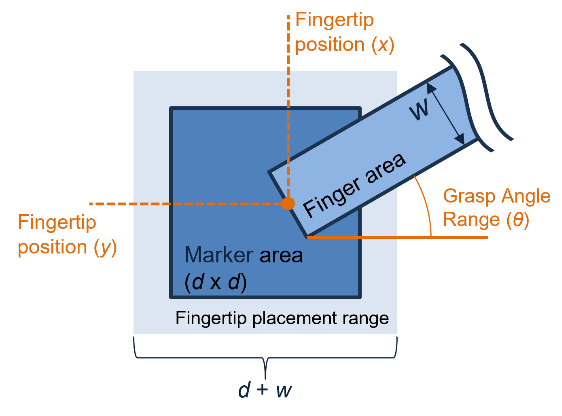}\\{\footnotesize (a) Evaluation model}\end{minipage}\par\vspace{1.5ex}
\begin{minipage}[b]{0.30\linewidth}\centering\includegraphics[width=\linewidth]{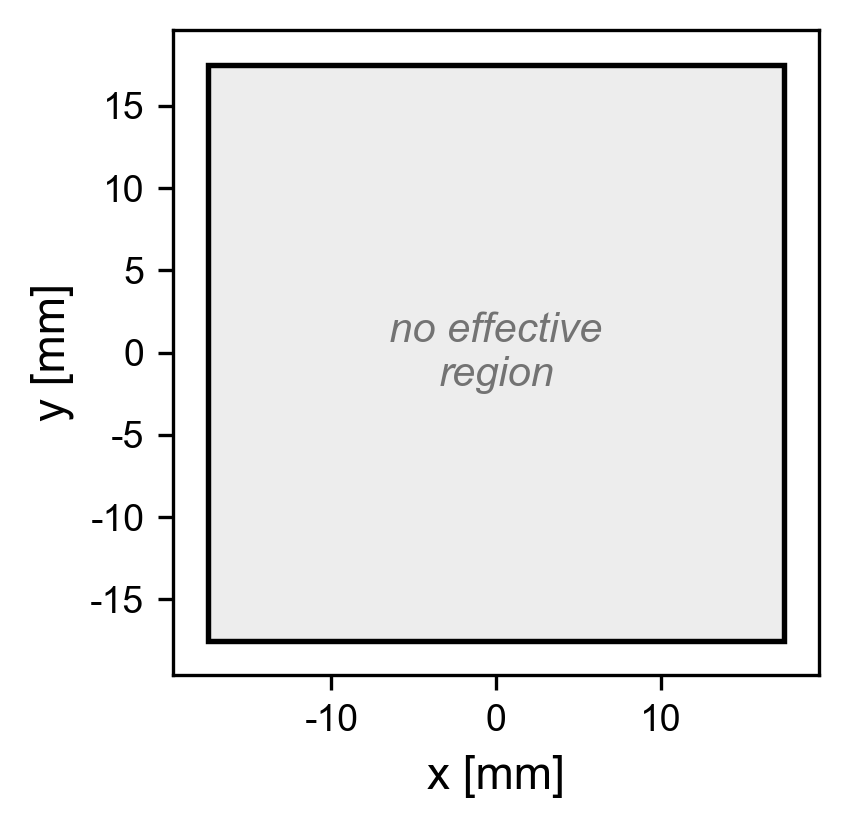}\\{\footnotesize (b) quARtet-D}\end{minipage}\hspace{0.02\linewidth}%
\begin{minipage}[b]{0.30\linewidth}\centering\includegraphics[width=\linewidth]{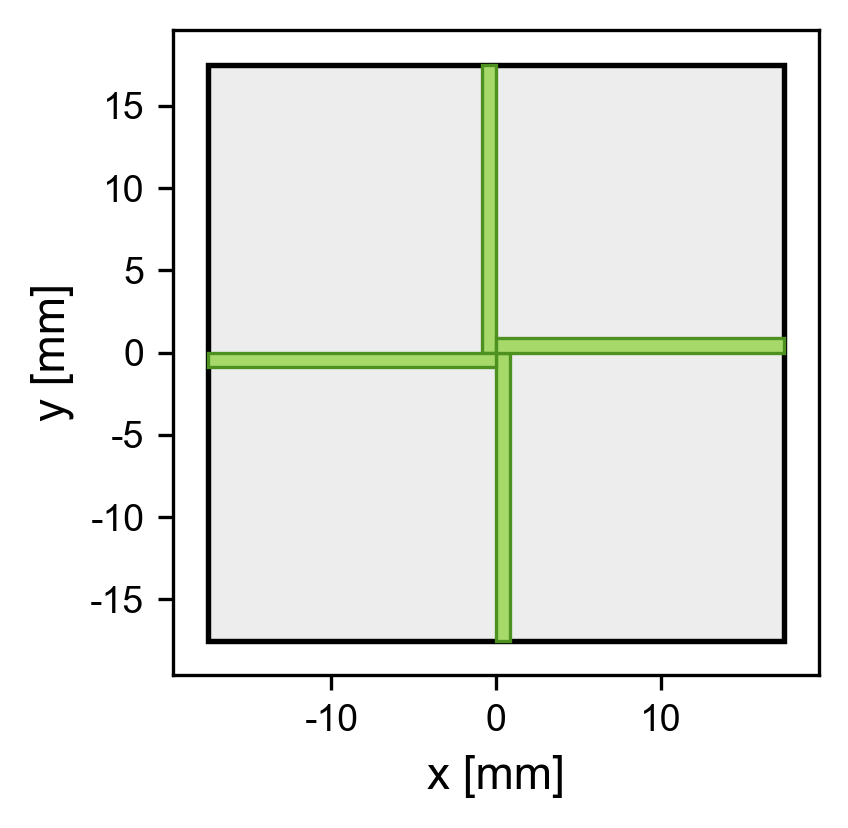}\\{\footnotesize (c) quARtet-P}\end{minipage}\hspace{0.02\linewidth}%
\begin{minipage}[b]{0.30\linewidth}\centering\includegraphics[width=\linewidth]{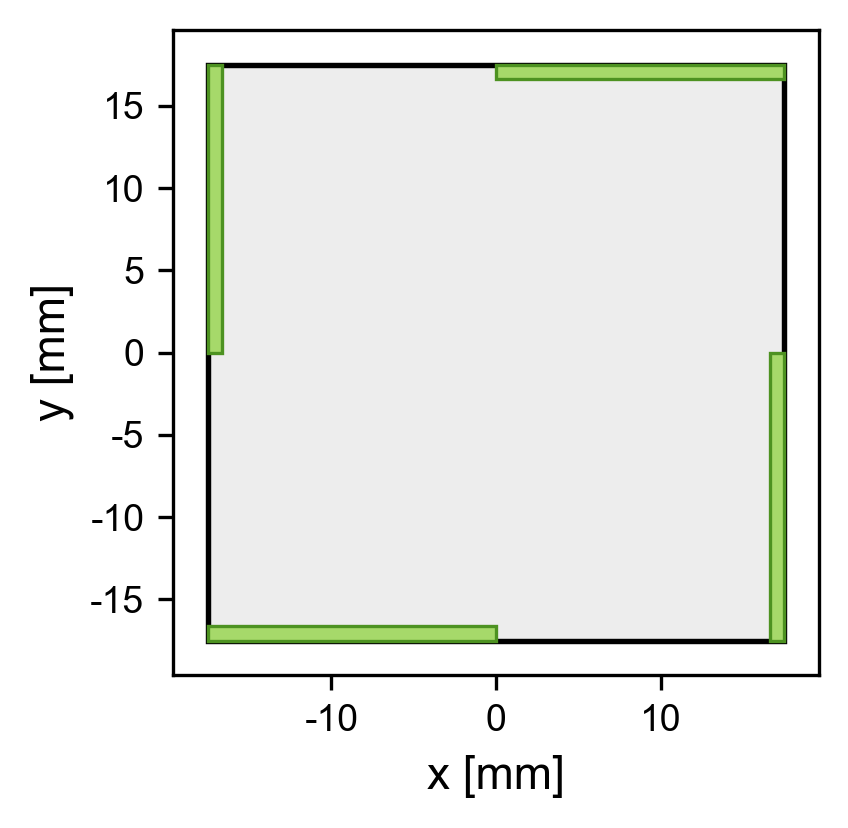}\\{\footnotesize (d) quARtet-rP}\end{minipage}
\caption{Idealized planar-grasp model and effective graspable regions on the 35~mm marker footprint. The angle labeled $\theta$ in panel~(a) is the contact angle $\alpha$ of Section~S6.1. quARtet-D exposes no effective region under this definition, whereas the pitch-based layouts retain strips. Single is the fully planar reference.}\label{fig:graspability}
\end{figure}

This section defines the idealized planar-grasp model (Fig.~\ref{fig:graspability}) used as a geometric aid for interpreting RQ2. The main-text swing-down experiment provides the primary physical evidence.

\subsection*{S6.1. Marker and finger domains}
Let the marker domain, with footprint side length $d=35$~mm, be
\begin{equation}D=\left[-\frac{d}{2},\frac{d}{2}\right]\times\left[-\frac{d}{2},\frac{d}{2}\right].\end{equation}
The effective graspable set is $E=\bigcup_{k=1}^{K}E_k$, where each $E_k$ is a rectangular strip along a flat raised edge. A candidate contact pose is $(\bm p,\alpha)$, with midpoint $\bm p$ and in-plane angle $\alpha$. The planar finger of width $w$ is modeled in local coordinates by
\begin{equation}F_0=\{(x,y)\in\mathbb R^2\mid x\ge0,\ |y|\le w/2\},\end{equation}
and in marker coordinates by $F(\bm p,\alpha)=\{\bm p+R(\alpha)\bm q\mid\bm q\in F_0\}$.

\subsection*{S6.2. Area-ratio metric}
For each sampled pose, the total overlap is $O=F\cap D$ and the effective overlap is $O_E=F\cap E$. On a regular grid $G$, the effective area is approximated by the convex-hull area
\begin{equation}A_E=\operatorname{area}\left(\operatorname{conv}(G\cap O_E)\right),\end{equation}
with $A_E=0$ for an empty or invalid hull. The total area is proportional to $|G\cap O|$, and the per-pose ratio is $r=A_E/A_{\rm all}$. The reported value is the mean $\bar r$ over valid sampled poses.

For each finger width, $\bar r$ is normalized (Table~\ref{tab:graspability}) using quARtet-D and Single as the reference values:
\begin{equation}\bar r_{\rm norm}=\frac{\bar r-\bar r_{\rm D}}{\bar r_{\rm Single}-\bar r_{\rm D}}.\end{equation}
Because quARtet-D's raw mean area-ratio is zero under this definition, the normalization reduces to $\bar r/\bar r_{\rm Single}$.

\begin{table}[!htbp]\centering
\caption{Normalized mean area-ratio for the idealized model.}
\label{tab:graspability}
\begin{tabular}{lccc}\toprule
Marker & $w=8.75$~mm & $w=17.5$~mm & $w=35$~mm \\
\midrule
quARtet-D  & 0 & 0 & 0 \\
quARtet-P  & 0.12 & 0.16 & 0.22 \\
quARtet-rP & 0.18 & 0.28 & 0.45 \\
Single     & 1 & 1 & 1 \\
\bottomrule\end{tabular}
\end{table}

The metric does not model friction, force closure, compliance, or collision with the full gripper body. It should be interpreted as a layout-level accessibility proxy, not as a predictor of physical grasp success or as a basis for ranking small differences between quARtet-P and quARtet-rP.

\section*{S7. Swing-down grasp-stability details}
Main-text Section~4.3 presents the setup, representative time series, and five-trial summary. This section provides the all-trial evidence: first the repeatability of the swing motion itself, then the per-trial behavior of the grasp. The object was re-grasped before each swing, and five trials were analyzed per marker type. The relative object pose was formed per frame as $R_g^{\mathsf T}(o-g)$ and $R_g^{\mathsf T}R_o$, averaged over the central 0.5~s of the static holds before and after the swing, and differenced to obtain position and orientation slip. Here $g$ and $o$ are the gripper and object rigid-body origins. Each coincides with the centroid of its six-sphere constellation (verified against the per-sphere motion-capture recordings), and the two centroids lie about $0.32$~m apart, consistently for all marker types. Translation slip is therefore evaluated at the object's sphere-constellation centroid and includes a lever-arm contribution when orientation slip is present. The same origins, and hence the same slip convention, are used for every marker type, so the lever-arm term enters the translation slip of every marker type in the same way.

\begin{figure}[!htbp]\centering
\includegraphics[width=.62\linewidth]{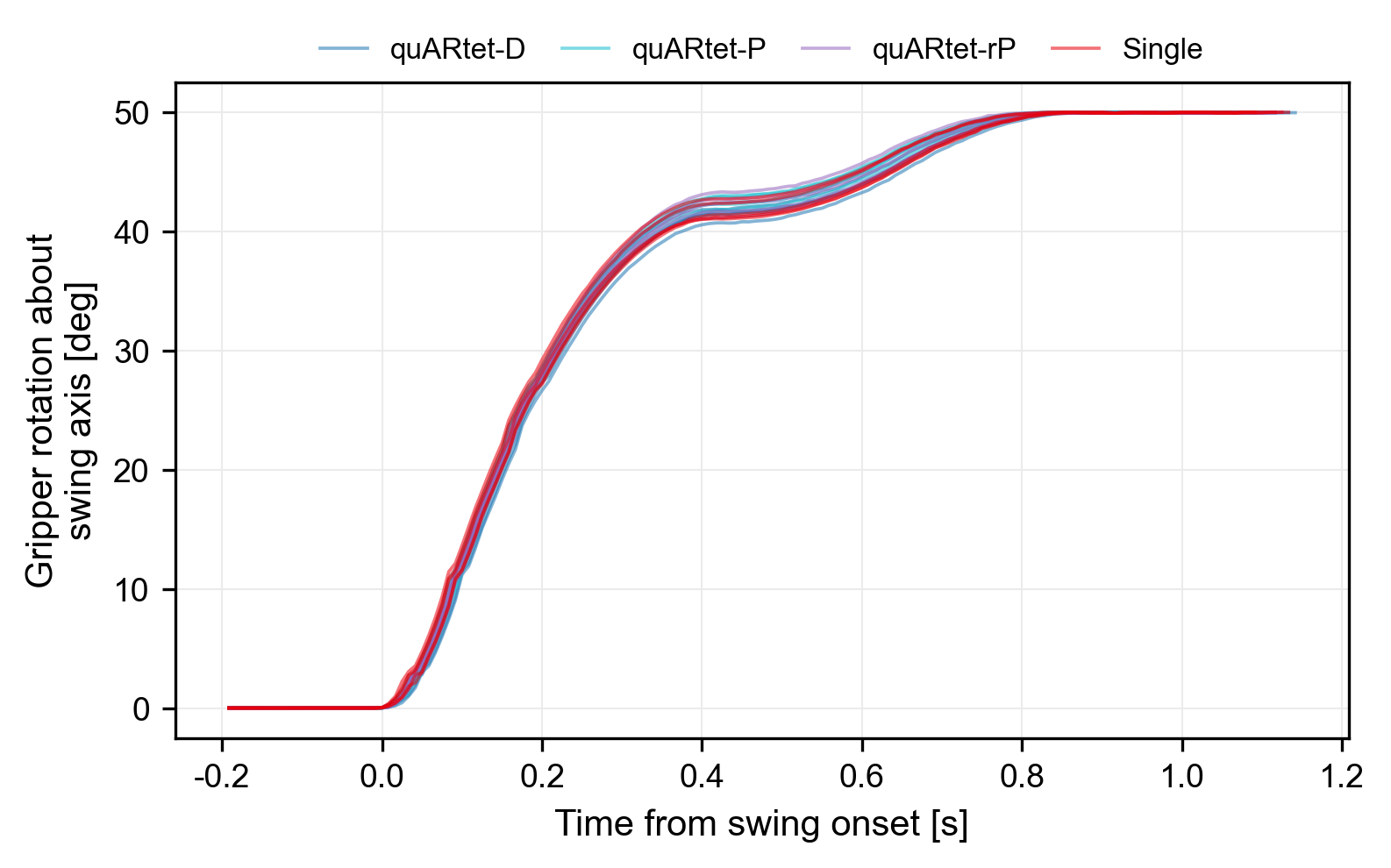}
\caption{Gripper tilt profiles of all 20 swings (4 marker types $\times$ 5 trials), aligned at swing onset with the swing shown positive. The trajectories coincide in shape across trials and marker types, complementing the scalar metrics of Table~\ref{tab:motion_rep}.}\label{fig:motion_overlay}\end{figure}

\textbf{Repeatability of the swing motion.} Table~\ref{tab:motion_rep} quantifies the realized swing of every trial from the motion-capture record, and Fig.~\ref{fig:motion_overlay} overlays the gripper tilt profiles of all 20 swings. Both show that the four marker types received the same repeatable input motion. The scalar metrics agree to within their small trial-to-trial SDs, and the trajectory shapes coincide.

\begin{table}[!htbp]
\centering
\caption{Repeatability of the realized swing-down motion, measured by motion capture (mean$\pm$SD over the $N=5$ swings of each marker type, and pooled over all 20 swings). Amplitude is the change in median gripper tilt between the pre- and post-swing holds. Duration is the time to traverse $5$--$95\%$ of that amplitude. Peak speed is the sustained peak of the swing-axis tilt rate ($5$-frame median filter at $120$~Hz). Brief unfiltered frame-to-frame peaks reach ${\sim}230\tcdegree$/s for at most $17$~ms. The near-identical values across trials and marker types show that every marker type experienced the same commanded motion and inertial load. The contact conditions at the fingers are layout-specific by design, and the grip force was not measured in situ.}
\label{tab:motion_rep}
\begin{tabular}{@{}lccc@{}}
\toprule
Marker & Amplitude [$\tcdegree$] & Duration [s] & Peak speed [$\tcdegree$/s] \\
\midrule
quARtet-D & $49.99\pm0.01$ & $0.65\pm0.01$ & $182\pm2$ \\
quARtet-P & $49.99\pm0.01$ & $0.63\pm0.01$ & $181\pm1$ \\
quARtet-rP & $49.96\pm0.00$ & $0.64\pm0.02$ & $181\pm3$ \\
Single & $50.01\pm0.00$ & $0.65\pm0.02$ & $181\pm3$ \\
\midrule
All 20 swings & $49.99\pm0.02$ & $0.64\pm0.02$ & $181\pm2$ \\
\bottomrule
\end{tabular}
\end{table}

\begin{figure}[!htbp]\centering
\begin{minipage}[b]{0.48\linewidth}\centering\includegraphics[width=.95\linewidth]{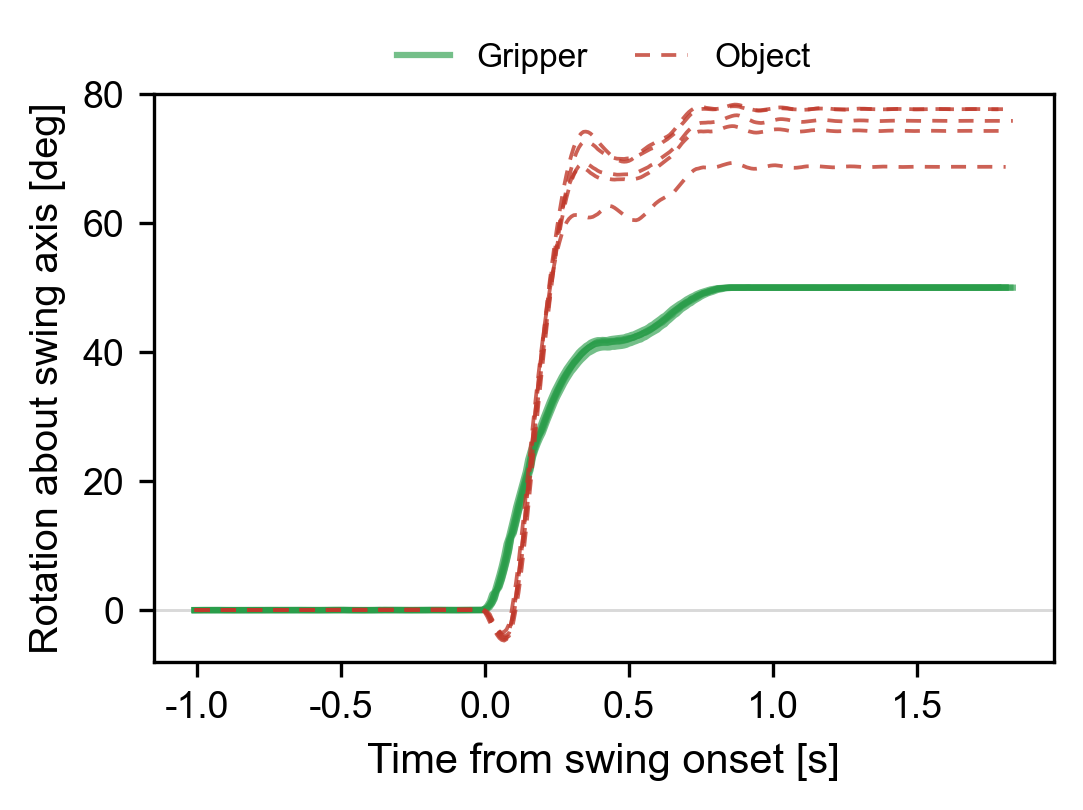}\\{\footnotesize (a) quARtet-D}\end{minipage}\hfill
\begin{minipage}[b]{0.48\linewidth}\centering\includegraphics[width=.95\linewidth]{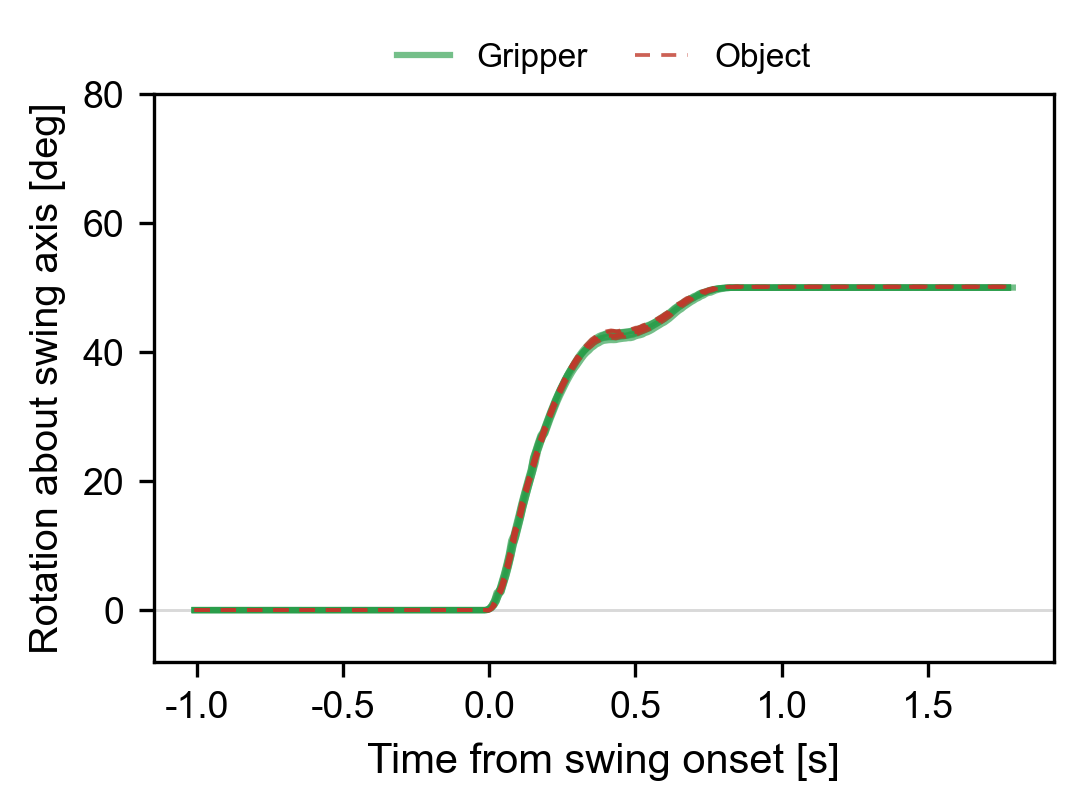}\\{\footnotesize (b) quARtet-P}\end{minipage}
\vspace{1ex}
\begin{minipage}[b]{0.48\linewidth}\centering\includegraphics[width=.95\linewidth]{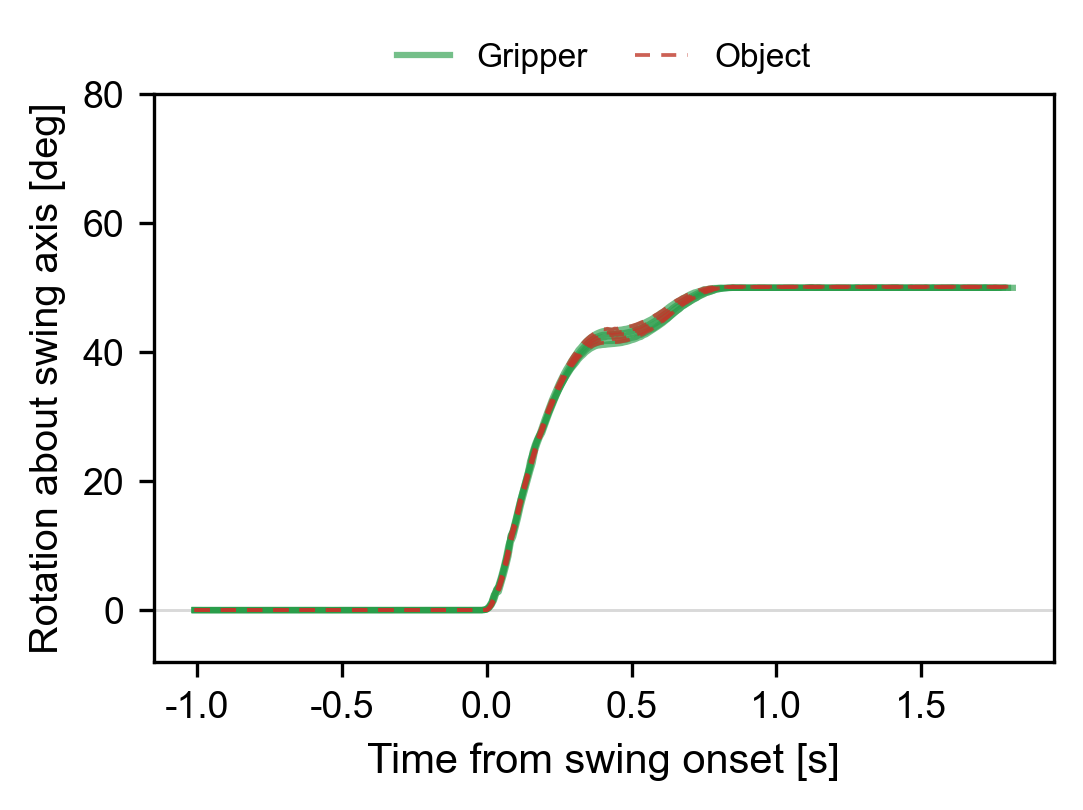}\\{\footnotesize (c) quARtet-rP}\end{minipage}\hfill
\begin{minipage}[b]{0.48\linewidth}\centering\includegraphics[width=.95\linewidth]{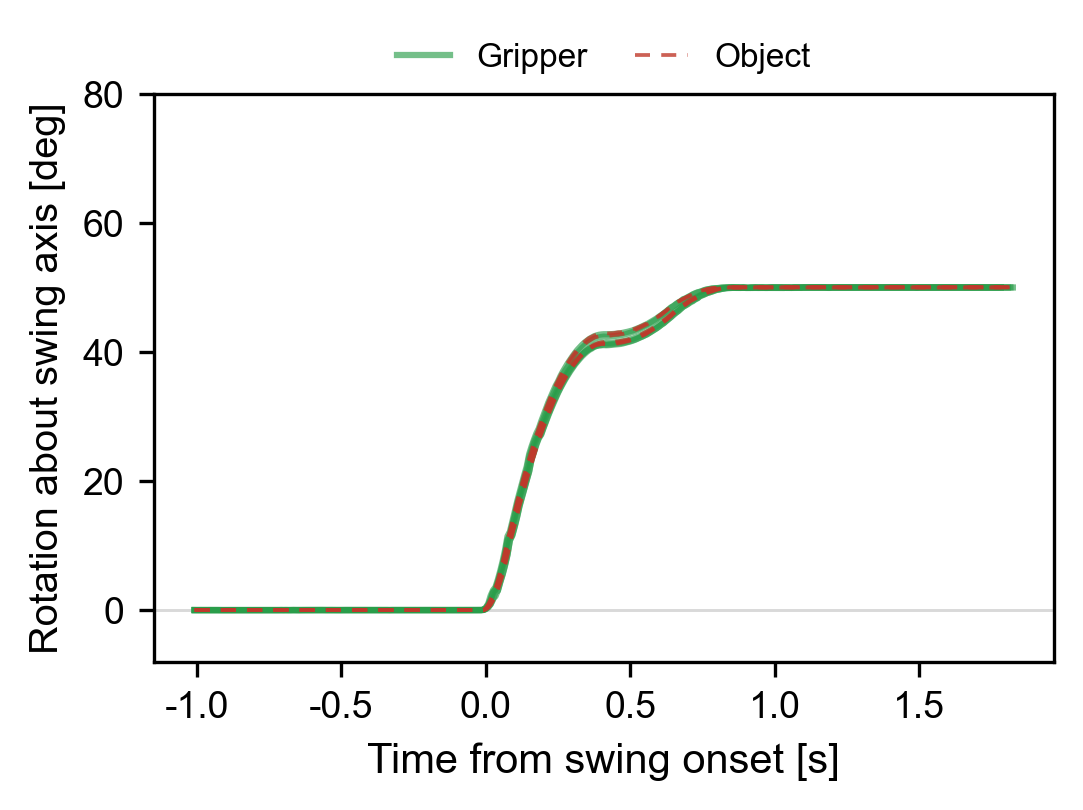}\\{\footnotesize (d) Single}\end{minipage}
\caption{All five swing-down time series per marker type (gripper solid green, object dashed red, with the dashed style keeping both visible where they coincide). Time is aligned at swing onset and each window spans the two bracketing holds. The five gripper curves of each panel coincide within $3\tcdegree$ at any instant and separate visibly only during the fast transient, so the trials repeat the same motion. The object curves separate persistently only for quARtet-D.}\label{fig:S12}\end{figure}

\begin{figure}[!htbp]\centering
\begin{minipage}[b]{0.48\linewidth}\centering\includegraphics[width=.95\linewidth]{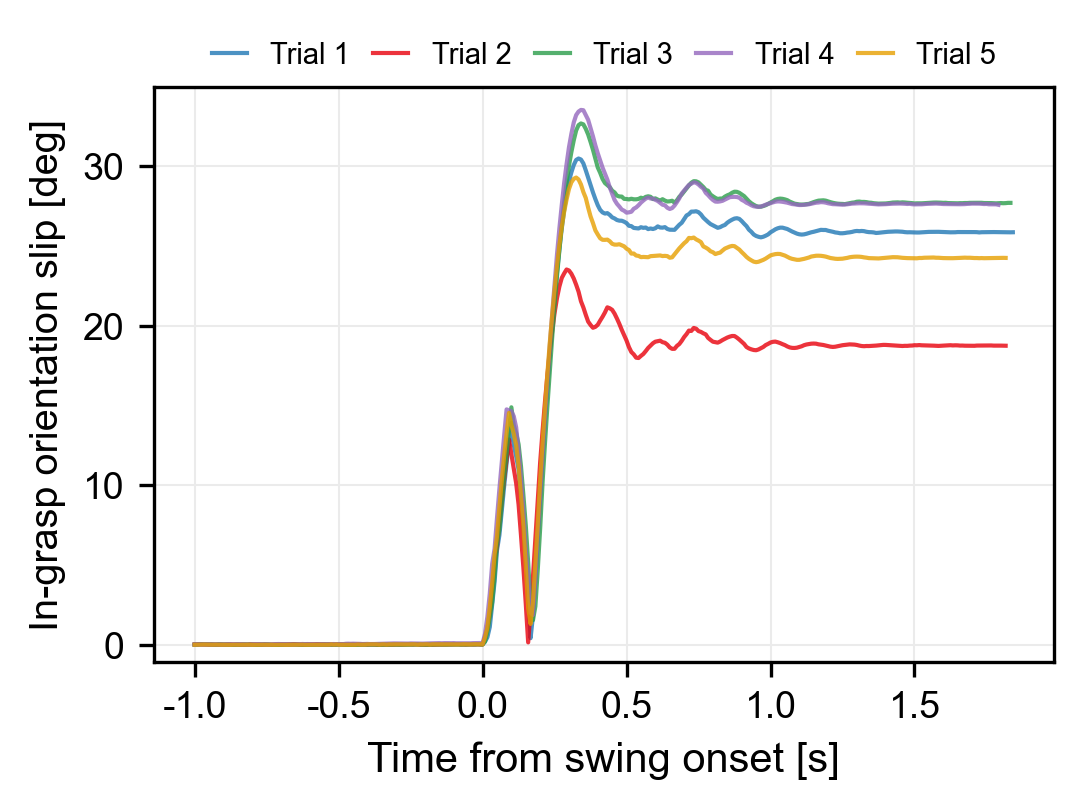}\\{\footnotesize (a) quARtet-D}\end{minipage}\hfill
\begin{minipage}[b]{0.48\linewidth}\centering\includegraphics[width=.95\linewidth]{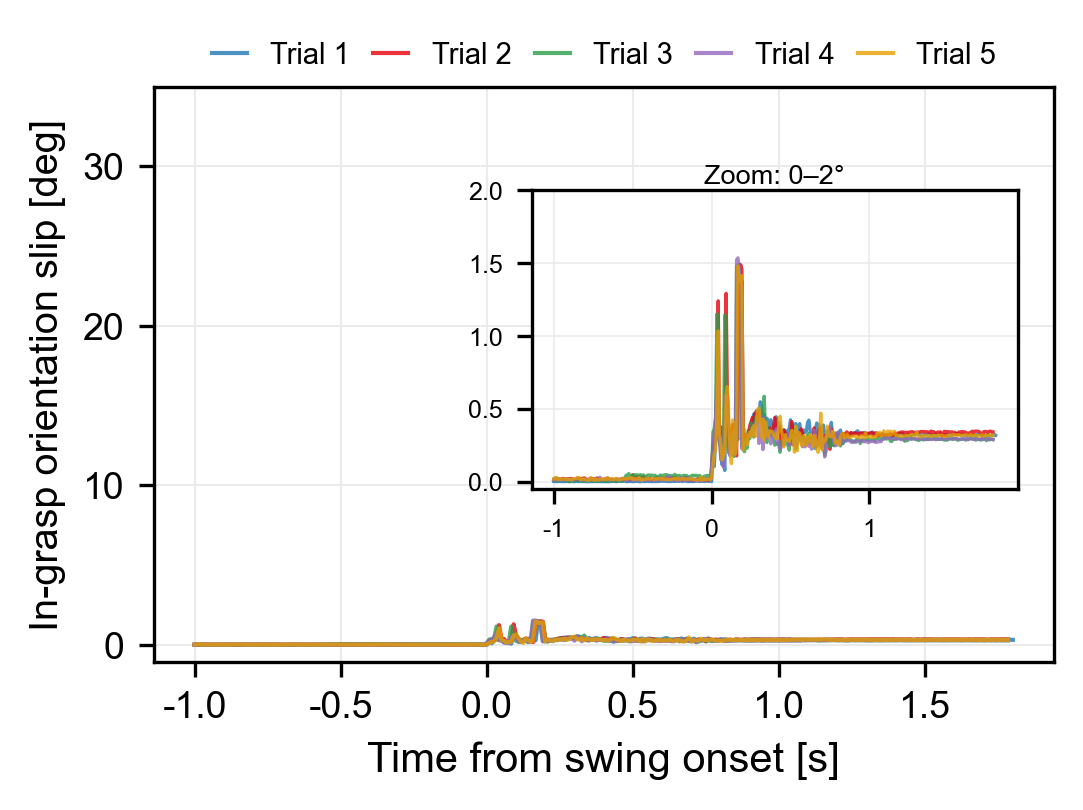}\\{\footnotesize (b) quARtet-P}\end{minipage}
\vspace{1ex}
\begin{minipage}[b]{0.48\linewidth}\centering\includegraphics[width=.95\linewidth]{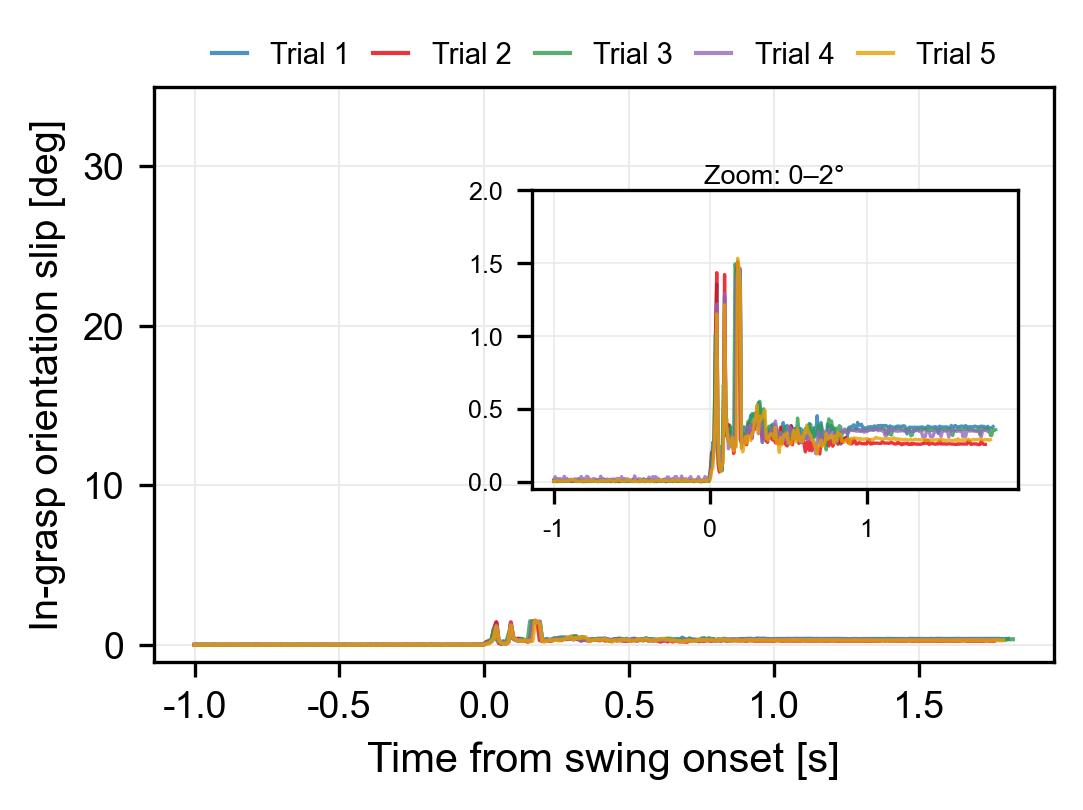}\\{\footnotesize (c) quARtet-rP}\end{minipage}\hfill
\begin{minipage}[b]{0.48\linewidth}\centering\includegraphics[width=.95\linewidth]{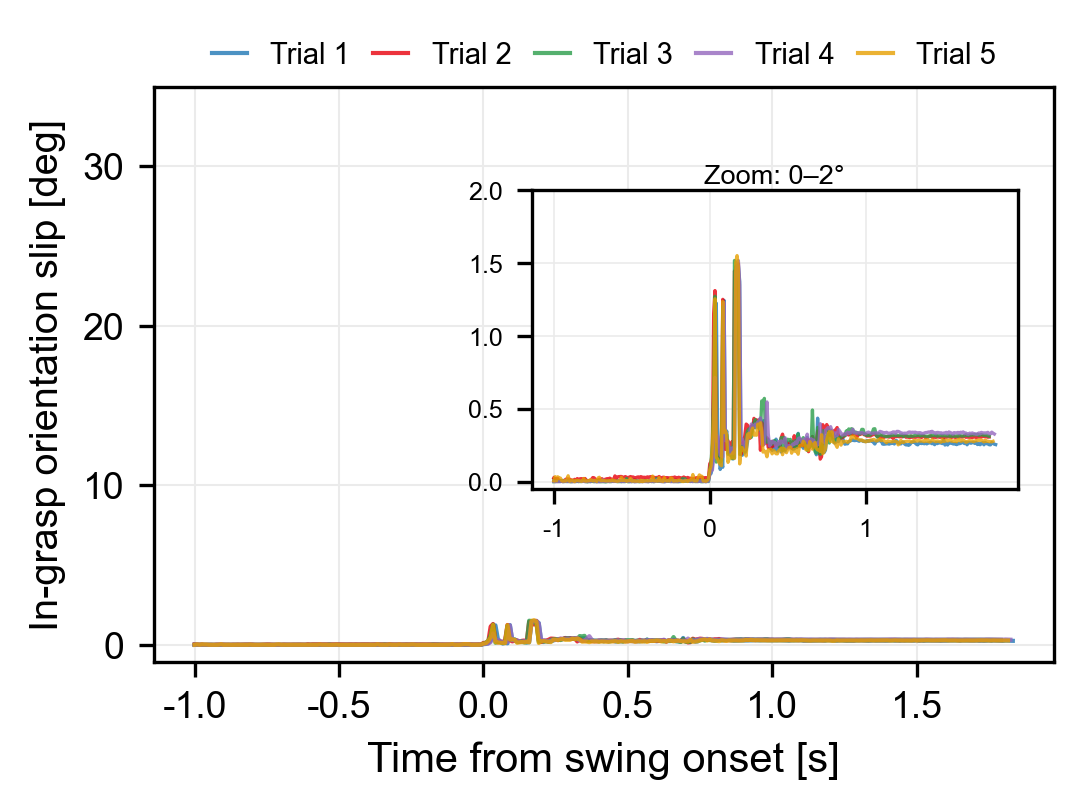}\\{\footnotesize (d) Single}\end{minipage}
\caption{Per-trial in-grasp orientation slip traces: for each trial, the geodesic angle between the current gripper-relative object rotation and its pre-swing-hold mean. The mean of each curve over the central 0.5~s of the post-swing hold is that trial's orientation slip, which enters the per-marker-type mean of the main-text swing-down table and is plotted in Fig.~\ref{fig:S13}. Insets magnify the $0$--$2\tcdegree$ range for the three retaining marker types. All five quARtet-D trials show the same failure sequence, a transient during the swing followed by a large permanent offset ($19$--$28\tcdegree$), whereas quARtet-P, quARtet-rP, and Single show a brief transient of at most ${\sim}1.5\tcdegree$ (elastic deformation of the grasp and tracking noise during the fast motion) that returns to ${\sim}0.3\tcdegree$.}\label{fig:reltrace}\end{figure}

\begin{figure}[!htbp]\centering
\includegraphics[width=.9\linewidth]{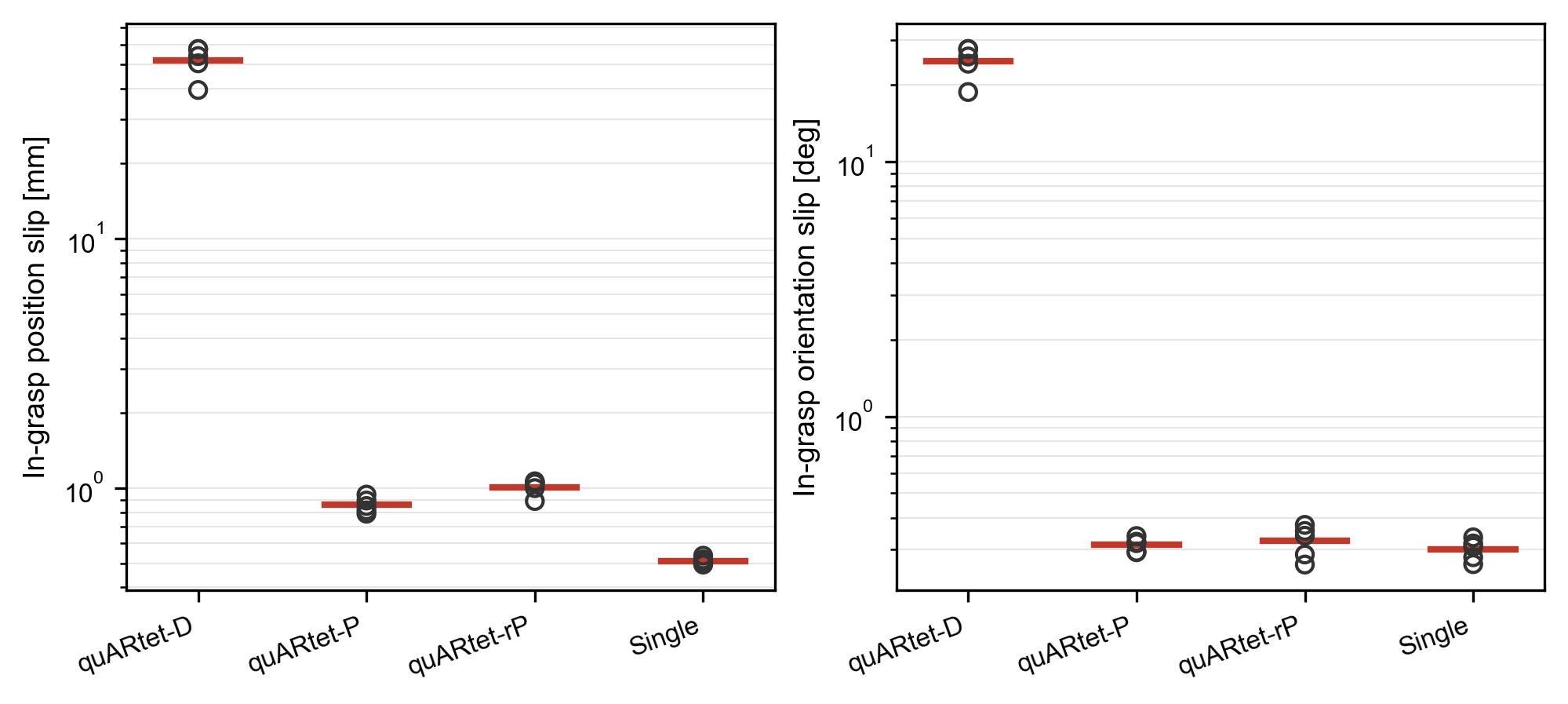}
\caption{Per-trial in-grasp position and orientation slip ($N=5$ independently re-grasped trials per marker type, logarithmic axes), in the order quARtet-D, quARtet-P, quARtet-rP, and Single. Red bars mark per-marker-type means.}\label{fig:S13}\end{figure}

The large quARtet-D slip appears in all five trials (Figs.~\ref{fig:S12} and~\ref{fig:reltrace}, with per-trial values in Fig.~\ref{fig:S13}). The object traces of the other three marker types remain close to the gripper traces.

\textbf{Statistical comparison.} Welch $t$-tests of each quARtet layout against Single on the five per-trial slips (two-sided, Holm-corrected within each metric, with Mann--Whitney tests giving the same conclusions at the $0.05$ level) yield the following differences. For position slip, quARtet-D $+51.3$~mm ($95\%$~CI $[42.0,60.5]$~mm, adjusted $p=2.1\times10^{-4}$), quARtet-P $+0.35$~mm ($[0.27,0.42]$~mm, $p=2.1\times10^{-4}$), and quARtet-rP $+0.50$~mm ($[0.41,0.59]$~mm, $p=1.8\times10^{-4}$). For orientation slip, quARtet-D $+24.6\tcdegree$ ($[20.0,29.1]\tcdegree$, $p=3.5\times10^{-4}$), while quARtet-P and quARtet-rP do not differ significantly from Single ($p=0.77$ after correction). The pitch-based layouts therefore incur a statistically detectable but small position-slip penalty of about half a millimeter and no detectable orientation-slip penalty, whereas quARtet-D slips about a hundred times more.

\textbf{Relation to the idealized metric.} The idealized model of Section~S6 favors quARtet-rP over quARtet-P (Table~\ref{tab:graspability}), whereas the physical test slightly favors quARtet-P. This is plausibly a consequence of the grasp pose used throughout the test: the gripper was centered on the marker with the distal fingertip edge aligned to the marker edge farthest from the palm, so the fingertips engaged mainly the central region of the face, where the pitch layout retains its accessible inner-edge strips, while the outer-edge strips that favor reversed-pitch in the idealized metric were only partially engaged.

\end{document}